\pdfoutput=1

\documentclass{article}
\usepackage[dvipsnames]{xcolor}
\usepackage[accepted]{icml2023}

\usepackage[utf8]{inputenc}
\usepackage[T1]{fontenc}
\usepackage{graphicx}
\usepackage{amsmath}
\usepackage{amssymb}
\usepackage{hyperref}
\usepackage{cancel}
\usepackage{subfiles}

\usepackage{microtype}
\usepackage{graphicx}
\usepackage{subfigure}
\usepackage{booktabs} 

\usepackage{amsmath}
\usepackage{amssymb}
\usepackage{mathtools}
\usepackage{amsthm}

\usepackage{amsmath,amssymb}
\usepackage{mathtools}
\usepackage{dsfont}
\newcommand{\x}{x}
\newcommand{\y}{y}
\newcommand{\xt}{\x_t}

\newcommand{\xstar}{\x^*}
\newcommand{\ystar}{\y^{*}}
\newcommand{\xbar}{\overline{\x}}
\newcommand{\ybar}{\overline{\y}}

\renewcommand*\hat\widehat
\renewcommand*\tilde\widetilde
\renewcommand*{\vec}[1]{\boldsymbol{#1}}

\newcommand{\opnorm}[1]{\left\|#1\right\|_{\text{op}}}

\newcommand{\Ceil}[1]{\left\lceil#1\right\rceil}

\newcommand{\elltilde}{\widetilde{\ell}}
\newcommand{\wtilde}{\widetilde{w}}
\newcommand{\cP}{\mathcal{P}}
\newcommand{\Ohat}{\widehat{O}}

\newcommand{\cmp}{u}

\newcommand{\w}{w}
\newcommand{\wt}{\w_t}
\newcommand{\wtpp}{\w_\tpp}

\usepackage{tikz}

\newcommand{\maxOp}{\vee}
\newcommand{\minOp}{\wedge}
\newcommand{\Max}[1]{\max\Set{#1}}
\newcommand{\Min}[1]{\min\Set{#1}}

\newcommand{\wrt}{w.r.t}

\newcommand{\sumtT}{\sum_{t=1}^T}

\newcommand{\R}{\mathbb{R}}

\newcommand{\Set}[1]{\left\{#1\right\}}
\newcommand{\sbrac}[1]{\left[#1\right]}
\newcommand{\brac}[1]{\left(#1\right)}
\newcommand{\cA}{\mathcal{A}}

\newcommand{\cM}{\mathcal{M}}

\newcommand{\cX}{\mathcal{X}}

\newcommand{\cC}{\mathcal{C}}
\newcommand{\cS}{\mathcal{S}}
\newcommand{\cY}{\mathcal{Y}}

\newcommand{\cL}{\mathcal{L}}

\newcommand{\argmin}{\operatorname{arg\,min}}

\newcommand{\norm}[1]{\left\|#1\right\|}

\newcommand{\grad}{\nabla}

\newcommand{\inner}[1]{\left\langle #1 \right\rangle}

\newcommand{\Log}[1]{\log\left(#1\right)}
\newcommand{\Exp}[1]{\exp\left(#1\right)}

\newcommand{\E}{\mathbb E}

\newcommand{\defeq}{\overset{\text{def}}{=}}

\newcommand{\eps}{\varepsilon}

\newcommand{\abs}[1]{\left|#1\right|}
\newcommand{\proj}{\Pi}

\newcommand{\zeros}{\mathbf{0}}

\newcommand{\half}{\frac{1}{2}}

\newcommand{\tpp}{{t+1}}
\newcommand{\tmm}{{t-1}}

\newcommand{\fG}{\mathfrak{G}}
\newcommand{\Otilde}{\widetilde{O}}

\newcommand{\gtilde}{\widetilde{g}}

\usepackage{thmtools,thm-restate}

\theoremstyle{plain}
\newtheorem{theorem}{Theorem}[section]

\newtheorem{manualtheoreminner}{Theorem}
\newenvironment{manualtheorem}[1]{%
  \renewcommand\themanualtheoreminner{#1}%
  \manualtheoreminner
}{\endmanualtheoreminner}

\theoremstyle{definition}
\newtheorem{definition}[theorem]{Definition}

\theoremstyle{remark}
\newtheorem{remark}[theorem]{Remark}

\newcommand{\For}[2]{
  \FOR{#1}
  #2
  \ENDFOR
}

\newcommand{\KwInput}[1]{\textbf{Input}: #1}
\newcommand{\KwInitialize}[1]{\textbf{Initialize}: #1}

\usepackage{tcolorbox}
\newcounter{BoxCounter}
\newcounter{PreBoxCounter}
\newenvironment{Boxed*}[2][\small]
{
  \begin{figure*}[!t]
    \begin{tcolorbox}[title=\textbf{Box \arabic{PreBoxCounter}:} #2, colback=white, colbacktitle=white, coltitle=black, arc=0pt,outer arc=0pt, fontupper=#1]
    }{
    \end{tcolorbox}
  \end{figure*}
}

\def\ie{\textit{i.e.}}
\def\eg{\textit{e.g.}}

\icmltitlerunning{Unconstrained Online Learning with Unbounded Losses}

\usepackage[capitalize,noabbrev]{cleveref}

\begin{document}

\twocolumn[
\icmltitle{Unconstrained Online Learning with Unbounded Losses}

\icmlsetsymbol{equal}{*}

\begin{icmlauthorlist}
\icmlauthor{Andrew Jacobsen}{ualberta,amii}
\icmlauthor{Ashok Cutkosky}{bu}
\end{icmlauthorlist}

\icmlaffiliation{ualberta}{Department of Computing Science, University of
  Alberta, Edmonton, Canada}
\icmlaffiliation{amii}{Alberta Machine Intelligence Institute (Amii), Edmonton, Canada}
\icmlaffiliation{bu}{Department of Electrical and Computer Engineering, Boston University, Boston, Massachussetts}

\icmlcorrespondingauthor{Andrew Jacobsen}{ajjacobs@ualberta.ca}

\icmlkeywords{Machine Learning, ICML}

\vskip 0.3in
]

\printAffiliationsAndNotice{}  

\begin{abstract}
  Algorithms for online learning typically require one or more boundedness assumptions: that the domain is bounded, that the losses are Lipschitz, or both. In this paper, we develop a new setting for online learning with unbounded domains and non-Lipschitz losses. For this setting we provide an algorithm which guarantees $R_{T}(u)\le \tilde O(G\|u\|\sqrt{T}+L\|u\|^{2}\sqrt{T})$ regret on any problem where the subgradients satisfy $\|g_{t}\|\le G+L\|w_{t}\|$, and show that this bound is unimprovable without further assumptions. We leverage this algorithm to develop new saddle-point optimization algorithms that converge in duality gap in unbounded domains, even in the absence of meaningful curvature. Finally, we provide the first algorithm achieving non-trivial dynamic regret in an unbounded domain for non-Lipschitz losses, as well as a matching lower bound. The regret of our dynamic regret algorithm automatically improves to a novel $L^{*}$ bound when the losses are smooth.
\end{abstract}

\newcommand{\SectionSmooth}{Online Learning with Smooth Losses}
\newcommand{\SectionQuadraticallyBounded}{Online Learning with Quadratically
  Bounded Losses}
\newcommand{\SectionDynamic}{Dynamic Regret}
\newcommand{\SectionStableExperts}{A Stable Experts Algorithm Without Domain Clipping}
\newcommand{\SectionAdaptive}{Adaptive Algorithms}
\newcommand{\SectionLowerbounds}{Lowerbounds}
\newcommand{\SecSaddlePoint}{Unconstrained Saddle-point Optimization}

\section{Online Learning}%
\label{sec:intro}

This paper introduces new techniques for online convex optimization (OCO), a
standard framework used to model learning from a stream of data
\citep{cesa2006prediction,shalev2011online, hazan2016introduction,
  orabona2019modern}. Formally, consider $T$ rounds of interaction between an
algorithm and an environment. In each round, the algorithm chooses a $w_t$ in
some convex subset $W$ of a Hilbert space, after which the environment chooses a
convex loss function $\ell_t:W\to \R$. The standard performance metric in
this setting is \emph{regret} $R_{T}(\cmp)$, the cumulative loss relative to an
unknown benchmark point $u\in W$:
\begin{align*}
    R_T(u) = \sum_{t=1}^T \ell_t(w_t) -\ell_t(u).
\end{align*}
In many applications of interest the appropriate
baseline is not any \textit{fixed} comparator, but rather
a \textit{trajectory} of points. This is often the case
in true streaming settings, wherein the losses are
generated from a data distribution that may be slowly shifting over time.
To better model settings such as these,
\textit{dynamic} regret measures the total loss
relative to that of a benchmark \textit{sequence}
of points $\vec{\cmp}=(\cmp_{1},\ldots,\cmp_{T})$
in $W$:
\begin{align*}
  R_{T}(\vec{\cmp})=\sumtT\ell_{t}(\wt)-\ell_{t}(\cmp_{t}).
\end{align*}
Our goal in this work is to develop algorithms that achieve favorable
regret and dynamic regret guarantees when \textit{both} the domain $W$ and range of $\ell_{t}$ may be
unbounded.

To illustrate the difficulty of our goal,
let us consider the special case where the loss functions
are linear functions, $\ell_{t}(\w)=\inner{g_{t},\w}$.
Clearly, if both $\norm{g_{t}}$ and $\norm{\w}$ are
allowed to be arbitrarily large then the adversary
can always ensure that the learner takes an
arbitrarily large loss on each round.
To alleviate this difficulty, prior works
assume
that one has access to a bound $D\ge \norm{\cmp}$
(usually by assuming that the domain is bounded with
$D\ge\sup_{x,y\in W}\norm{x-y}$),
that the subgradients are bounded $\fG_{T}\ge \max_{t}\norm{g_{t}}$,
that the losses map to a bounded range $\ell_{t}:W\to[a,b]$,
or some combination thereof.

In the simplest case, when one has access to both
a bound $D\ge\norm{\cmp}$ and a bound on the subgradients
$\fG_{T}\ge\norm{g_{t}}$ for all $t$,
classic methods based on Mirror Descent and Follow the
Regularized Leader achieve minimax optimal regret of $R_{T}(\cmp)\le O\big(D\fG_{T}\sqrt{T}\big)$
using a strongly convex regularizer
\citep{hazan2007adaptive,duchi10adagrad,mcmahan2010adaptive}.
When $D$ is available but not the Lipschitz bound
$\fG_{T}$, it is still possible to match
this guarantee up to constant factors,
in which case the algorithm is said to be \textit{Lipschitz adaptive}
\citep{orabona2018scale,mayo2022scalefree,cutkosky2019artificial}.
When the losses are $L$-smooth, these bounds can
be improved to
$R_{T}(\cmp)\le O\Big(LD^{2}+D\sqrt{L\sumtT\ell_{t}(\cmp)}\Big)$ --- referred to as
an $L^{*}$ bound ---
though prior works
still require one or more of the following assumptions: prior knowledge of
$\fG_{T}$, that $\ell_{t}$ has bounded range (known in advance),
prior knowledge of a lower bound $\ell_{t}^{*}\le \ell_{t}(\w)$ for all $w\in W$,
additional structural assumptions such as strong convexity or exp-concavity,
or by assuming the losses take some specific form such as the square loss
\citep{cesa1996worst,kivinen1997exponentiated,srebro2010smoothness,orabona2012beyond}.

If a bound $\fG_{T}\ge \max_{t}\norm{g_{t}}$ is known
but not the bound $D\ge \norm{\cmp}$, the situation gets significantly
trickier.
The essential difficulty is that without prior knowledge of how large the
comparator might be, the predictions $\wt$ could at any point be arbitrarily far away
from the benchmark, leading to high regret.
As such, the learner must take great care to control
$\norm{\wt}$ in such a way that it is
\emph{adaptive} to the unknown comparator norm $\norm{\cmp}$.
A standard result in this setting is
\begin{align}
    R_T(u)&\le O\left(\|u\|\fG_{T}\sqrt{T\log( T \|u\|+1)}\right)\label{eqn:paramfreebound},
\end{align}
which holds for all $u\in W$ and is known to be
optimal up to constants \citep{orabona2013dimensionfree}. Bounds of this form
are commonly referred to as ``comparator adaptive'' or
``parameter-free'' \citep{foster2015adaptive, orabona2016coin, van2019user, cutkosky2018black,mhammedi2020lipschitz,jacobsen2022parameter}.

The first results to avoid both the bounded domain and bounded gradient
assumptions have only been achieved in recent years.
\citet{cutkosky2019artificial} develops an algorithm
which achieves
$R_{T}(\cmp)\le O\brac{\norm{\cmp}\fG_{T}\sqrt{T\Log{\norm{\cmp}T+1}}+\fG_{T}\norm{\cmp}^{3}}$,
and \citet{mhammedi2020lipschitz} shows that the additional cubic penalty is
unavoidable while maintaining the $\tilde O\brac{\norm{\cmp}\fG_{T}\sqrt{T}}$
dependence.
Alternatively,
\citet{orabona2018scale} show that $R_{T}(\cmp)\le O(\norm{\cmp}^{2}\fG_{T}\sqrt{T})$ can
be attained without prior knowledge of $\fG_{T}$ in an unbounded domain,
avoiding the cubic penalty in exchange for a horizon-dependent
quadratic penalty.
Works such as \citet{mayo2022scalefree} and \citet{kempka2019adaptive}
show that \Cref{eqn:paramfreebound} can be achieved with
essentially no extra penalty in certain special cases such as
regression-type losses.

When it comes to dynamic regret, much less progress
has been made in alleviating
boundedness assumptions, with
nearly all existing results assuming both a bounded domain
and Lipschitz losses. Under both boundedness assumptions,
prior works have achieved minimax optimal dynamic regret of
$R_{T}(\vec{\cmp})\le O\brac{\fG_{T}\sqrt{(D^{2}+DP_{T})T}}$, where
$P_{T}=\sum_{t=2}^{T}\norm{\cmp_{t}-\cmp_{\tmm}}$ is the \textit{path-length} of the
comparator sequence \citep{zhang2018adaptive,jadbabaie2015online,cutkosky2020parameter}.
In an unbounded domain with Lipschitz losses, recent works
have achieved an analogous guarantee of
$R_{T}(\vec{\cmp})\le \tilde O(\fG_{T}\sqrt{(M^{2}+MP_{T})T})$, where
$M=\max_{t}\norm{\cmp_{t}}$
\cite{jacobsen2022parameter,luo2022corralling}.
We are unaware of any existing works that explicitly
investigate Lipschitz-adaptive dynamic regret, though
existing results can likely be made Lipschitz-adaptive
in a black-box manner
using the gradient clipping approach of
\citet{cutkosky2019artificial} in exchange for an appropriate
$\fG_{T}\max_{t}\norm{\cmp_{t}}^{3}$ penalty.

Importantly, note that \emph{in all of these prior works} there
is an implicit assumption that
\emph{there exists} a uniform bound such that
$G\ge \norm{\grad\ell_{t}(\w)}$ for any $\w\in W$ and
$\grad\ell_{t}(\w)\in\partial\ell_{t}(\w)$ --- even if it is not known
in advance. Otherwise, the
terms $\fG_{T}=\max_{t}\norm{g_{t}}$ can easily make
any of the aforementioned regret guarantees vacuous.

In this work, we study unconstrained online convex optimization
under a more general boundedness assumption on the gradients,
allowing the gradient norms to grow arbitrarily large away from
a given ``reference point'' $\w_{0}\in W$.
In \Cref{sec:quadratically-bounded} we provide an algorithm for this more general problem setting which achieves
a strict generalization of the usual comparator-adaptive bound in
\Cref{eqn:paramfreebound}, as well as a lower bound showing
that our result is unimprovable in general.
In
\Cref{sec:sp} we leverage this algorithm
to develop a new saddle-point optimization algorithm which
converges in duality gap in an unbounded domain without
requiring additional curvature assumptions such as strong-convexity/concavity.
In \Cref{sec:dynamic}, we turn to the problem of dynamic regret minimization
and develop an algorithm which
achieves dynamic regret
$R_{T}(\vec{\cmp})\le \tilde O\brac{M^{3/2}\sqrt{(M+P_{T})T}}$
and provide a matching lower bound.
This is the first algorithm to significantly alleviate
both the bounded domain and bounded subgradient assumptions
for dynamic regret. Moreover,
when the losses are $L_{t}$-smooth, the same algorithm automatically improves
to an $L^{*}$ bound of the form
$R_{T}(\vec{\cmp})\le \tilde O\brac{\sqrt{(M^{2}+MP_{T})\sumtT L_{t}\sbrac{\ell_{t}(\cmp_{t})-\ell_{t}^{*}}}}$.
To the best of our
knowledge, this is in fact the first $L^{*}$ bound
to be achieved for general smooth losses
without making either a uniformly-bounded subgradient or bounded range assumption
in an unbounded domain.

\textbf{Notations.} For brevity, we occasionally abuse notation
by letting $\grad f(x)$ denote an element of $\partial f(x)$. The Bregman
divergence \textit{w.r.t.} a differentiable function $\psi$
is
$D_{\psi}(x|y)=\psi(x)-\psi(y)-\inner{\grad\psi(y),x-y}$.
We use the compressed sum notation $g_{i:j}=\sum_{t=i}^{j}g_{t}$ and $\norm{g}_{a:b}^{2}=\sum_{t=a}^{b}\norm{g_{t}}^{2}$.
We denote $a\maxOp b = \Max{a,b}$ and $a\minOp b=\Min{a,b}$.
$\Delta_{N}$ denotes the $N$-dimensional simplex.
The notation $O(\cdot)$ hides constants,
$\Ohat(\cdot)$ hides constants and $\log(\log)$ terms, and $\Otilde(\cdot)$
hides up to and including $\log$ factors.

\section{\SectionQuadraticallyBounded}%
\label{sec:quadratically-bounded}

In an unbounded domain with unbounded losses,
it will generally be impossible to avoid linear regret without
\emph{some} additional assumptions. Intuitively, what's missing in this
problem is a frame-of-reference for the magnitude of a given loss.
In the Lipschitz or bounded-range settings, the learner always has
a frame-of-reference for the worst-case loss they might encounter.
In contrast, without these assumptions, hindsight becomes the only
frame-of-reference,
and the adversary can exploit this to ``trick'' the learner into
playing too greedily or too conservatively.

To make the problem tractable, yet still allowing the
losses to have unbounded range and subgradients,
we assume that the subgradients are bounded for all $t$
at \textit{some} reference point $\w_{0}$, but may become
arbitrarily large away from $\w_{0}$. This effectively gives the
learner access to an \textit{a priori} frame-of-reference for loss magnitudes,
yet still captures many problem settings where the losses can become arbitrarily
large in an unbounded domain.
\begin{definition}
  \label{def:quadratically-bounded}
  Let $(W,\norm{\cdot})$ be a normed space. A function $\ell:W\to \R$ is
  $(G,L)$-quadratically
  bounded \wrt{} $\norm{\cdot}$ at $w_{0}$ if
  for any $\w\in W$ and $\grad\ell(\w)\in\partial\ell(\w)$ it holds that
  \begin{align}
    \norm{\grad\ell(\w)}\le G+L\norm{\w-\w_{0}}.\label{eq:qb}
  \end{align}
\end{definition}
Note that \Cref{def:quadratically-bounded} is a strict generalization of the standard
Lipschitz condition: any $G$-Lipschitz
function is $(G,0)$-quadratically bounded. The definition also
captures $L$-smooth functions as a special case,
since any $L$-smooth function is
$(\norm{\partial\ell_{t}(\w_{0})},L)$-quadratically bounded at $\w_{0}$.
However, in general a function satisfying the quadratically bounded property
need not be smooth. \footnote{As a simple illustration, note that if $f(\w)$ is an
  $L$-smooth and
$(G,L)$-quadratically
bounded function, then $f(\w)+c\norm{\w}$ will be $(G+c,L)$ quadratically bounded but non-smooth.}
For the remainder of the paper we assume without loss of generality that
$\w_{0}=\zeros$ and $\norm{\cdot}$ is the Euclidean norm.

This assumption was initially studied in the context of
stochastic optimization by \citet{telgarsky2022stochastic},
where it was sufficient
to attain convergence in several settings of practical relevance.
In this work, we show that it
is also sufficient to achieve sublinear regret even in
\textit{adversarial} problem settings.
We will in fact take it one step further and consider
a stronger Online Linear Optimization (OLO)
version of the problem.
We say that
a \textit{sequence} $\Set{g_{t}}$ is $(G_{t},L_{t})$-quadratically bounded \wrt{}
$\Set{\wt}$ if for every $t$ we have $\norm{g_{t}}\le G_{t}+L_{t}\norm{\wt}$.
Then using the standard
reduction from OCO to OLO (see \eg{} \citet{zinkevich2003online}),
for any sequence of $(G_{t},L_{t})$-quadratically bounded convex functions we have
the following regret upper bound:
\begin{align*}
  R_{T}(\cmp) = \sumtT\ell_{t}(\wt)-\ell_{t}(\cmp)\le \sumtT\inner{g_{t},\wt-\cmp},
\end{align*}
where $g_{t}\in\partial\ell_{t}(\wt)$ and $\Set{g_{t}}$ is a $(G_{t},L_{t})$-quadratically bounded sequence \wrt{} $\Set{\wt}$.
Hence, one can solve OCO problems involving
quadratically bounded losses using any OLO algorithm that achieves sublinear
regret against sequences $\Set{g_{t}}$ that are quadratically bounded \wrt{} its
outputs $\Set{\wt}$.
Note that this is potentially a more difficult problem, as it gives the adversary freedom
to impose severe penalties whenever the learner plays large $\wt$,
yet this effect is experienced asymmetrically by the comparator:
the comparator can have large norm and not necessarily experience
large losses unless $\cmp$ is aligned with $g_{t}$ \emph{and}
the learner plays a point $\norm{\wt}\propto\norm{\cmp}$.
For brevity we refer to this harder problem setting as the
QB-OLO setting, and QB-OCO for the setting where
adversary must play $\ell_{t}$ satisfying
\Cref{def:quadratically-bounded}.

Surprisingly,  it turns out that
it is possible to achieve sublinear regret even in
the QB-OLO setting.
The following theorem
provides an algorithm which achieves sublinear regret
and requires no instance-specific hyperparameter tuning.
Proof can be found in \Cref{app:quadratically-bounded}.
\begin{restatable}{theorem}{QBAdaptiveRegret}\label{thm:qb-adaptive-regret}
  Let $\cA$ be an online learning algorithm and let $\wt\in W$
  its output on round $t$.
  Let $\Set{g_{t}}$ be a $(G_{t},L_{t})$-quadratically bounded sequence
  \wrt{} $\Set{\wt}$, where
  $G_{t}\in[0,G_{\max}]$ and $L_{t}\in [0,L_{\max}]$
  for all $t$.
  Let $\epsilon>0$, $V_{\tpp}=4G_{\max}^{2}+G_{1:t}^{2}$,
  $\rho_{\tpp}=\frac{1}{\sqrt{L_{\max}^{2}+L_{1:t}^{2}}}$,
  $\alpha_{\tpp}=\frac{\epsilon G_{\max}}{\sqrt{V_{\tpp}}\log^{2}\brac{V_{\tpp}/G_{\max}^{2}}}$.
  Denote $
    \Psi_{t}(\w)=3\int_{0}^{\norm{\w}}\min_{\eta\le \frac{1}{G_{\max}}}\sbrac{\frac{\Log{x/\alpha_{t}+1}}{\eta}+\eta V_{t}}dx$
  and set
  \begin{align*}
    \psi_{t}(\w)&=\Psi_{t}(\w)+\frac{2}{\rho_{t}}\norm{\w}^{2},\quad
    \varphi_{t}(\w)=\frac{L_{t}^{2}}{2\sqrt{L_{1:t}^{2}}}\norm{\w}^{2}.
  \end{align*}
  Then for any $\cmp\in W$, \Cref{alg:qb-adaptive-alg} guarantees
  \begin{align*}
      R_{T}(\cmp)\le O\brac{\epsilon +\norm{\cmp}\sqrt{G_{1:T}^{2}F_{T}(\norm{\cmp})}+\norm{\cmp}^{2}\sqrt{L_{1:T}^{2}}}
  \end{align*}
  where
  $F_{T}(\norm{\cmp})\le \Log{\frac{\norm{\cmp}\sqrt{T}\log^{2}(T)}{\epsilon}+1}$.
\end{restatable}
Let us briefly develop some intuition for how the above result is
constructed. \Cref{alg:qb-adaptive-alg} can be interpreted as an instance of the
Centered Mirror Descent algorithm recently developed by
\citet{jacobsen2022parameter}, which admits a generic regret guarantee of the form
$R_{T}(\cmp)\le \psi_{T}(\cmp)+\sumtT\varphi_{t}(\cmp)+\sumtT\delta_{t}$, where the
$\delta_{t}$ are similar to the ``stability'' terms encountered in vanilla
Mirror Descent, but with certain additional negative terms $\Delta_{t}$ and $\varphi_{t}$:
\begin{align*}
  \delta_{t}&\le O\Big(\inner{g_{t},\wt-\wtpp}-D_{\psi_{t}}(\wtpp|\wt)\\
  &\qquad\qquad-\Delta_{t}(\wtpp)-\varphi_{t}(\wt)\Big).
\end{align*}
It's easily verified that that $\psi_{T+1}(\cmp)+\sumtT\varphi_{t}(\cmp)$
match the terms in the upper bound, so the main difficulty
is making sure that the stability terms $\sumtT\delta_{t}$ disappear.
Crucially, because $\Set{g_{t}}$ is a $(G_{t},L_{t})$-quadratically bounded sequence
\wrt{} $\Set{\wt}$,
we have $\norm{g_{t}}\le G_{t}+L_{t}\norm{\wt}$. The utility of this is that
we can design \textit{separate regularizers} control the ``Lipschitz part''
$G_{t}$ and
the ``non-Lipschitz part'' $L_{t}\norm{\wt}$. In particular, using
a similar argument to \citet{jacobsen2022parameter},
by setting $\Psi_{t}(w)=O\bigg(G_{\max}\norm{\w}\sqrt{T\Log{\norm{\w}\sqrt{T}/\epsilon}}\bigg)$
we can ensure that the Lipschitz part of the bound is well-controlled:
\begin{align*}
  \sumtT\mspace{-2mu} G_{t}\norm{\wt-\wtpp}\mspace{-2mu}-\mspace{-2mu}D_{\Psi_{t}}(\wtpp|\wt)\mspace{-2mu}-\mspace{-2mu}\Delta_{t}(\wtpp)\mspace{-2mu}\le\mspace{-2mu} O(1).
\end{align*}
However, in general this $\Psi_{t}$ is not strong enough to control the
non-Lipschitz part $L_{t}\norm{\wt}$.
Instead,
for this term we use $\Phi_{t}(\w)=O\brac{L_{\max}\sqrt{T}\norm{\w}^{2}}$,
and then using
standard arguments for Mirror Descent with a strongly convex regularizer,
it can be shown that
\begin{align*}
  &L_{t}\norm{\wt}\norm{\wt-\wtpp}-D_{\Phi_{t}}(\wtpp|\wt)-\varphi_{t}(\wt)\\
  &\qquad\le
    O\brac{\frac{L_{t}\norm{\wt}^{2}}{\sqrt{T}}-\varphi_{t}(\wt)}\le 0
\end{align*}
by choosing $\varphi_{t}(\wt)=O\brac{\frac{L_{t}\norm{\wt}^{2}}{\sqrt{T}}}$.

\begin{algorithm}[t!]
  \caption{Algorithm for Quadratically Bounded Losses}%
  \label{alg:qb-adaptive-alg}
  \begin{algorithmic}
    \STATE\KwInput{$\psi_{1}:W\to \R_{\ge 0}$ with
      $\min_{\w\in W}\psi_{1}(\w)=0$, $G_{\max}$ and $L_{\max}$}
    \STATE\KwInitialize{$\w_{1}=\argmin_{w\in W}\psi_{1}(\w)$}
    \For{$t=1:T$}{
      \STATE Play $\wt$, observe $g_{t}\in\partial\ell_{t}(\wt)$
      \STATE Choose $G_{t}$ and $L_{t}$ satisfying $\norm{g_{t}}\le G_{t}+L_{t}\norm{\wt}$
      \STATE Choose functions $\psi_{\tpp}$, $\varphi_{t}$
      \STATE Set $\grad\varphi_{t}(\wt)\in\partial\varphi_{t}(\wt)$ and
      $\gtilde_{t}=g_{t}+\grad\varphi_{t}(\wt)$\\
      \STATE Set $\Delta_{t}(\w)=\psi_{\tpp}(\w)-\psi_{t}(\w)$
      \STATE Update
      \begin{align*}
        \wtpp &= \argmin_{\w\in W}\inner{\gtilde_{t},\w}+D_{\psi_{t}}(\w|\wt)+\Delta_{t}(\w)
      \end{align*}
    }
  \end{algorithmic}
\end{algorithm}

Note that in the setting of $G$-Lipschitz losses we have $L_{\max}=0$ and
hence set $G_{t}=\norm{g_{t}}$, so the bound reduces to the comparator-adaptive rate
of
$R_{T}(\cmp)\le  \hat O\brac{\norm{\cmp}\sqrt{\sumtT\norm{g_{t}}^{2}\Log{\frac{\norm{\cmp}\sqrt{T}}{\epsilon}+1}}}$,
which is known to be optimal up to constant and $\log(\log)$ terms \citep{mcmahan2012noregret,orabona2013dimensionfree}.
On the other hand, if $L_{\max}>0$ the algorithm can choose any
$G_{t}\le G_{\max}$ and $L_{t}\le L_{\max}$ such that $G_{t}+L_{t}\norm{\wt}\ge \norm{g_{t}}$.
Ideally these factors should be chosen to be tight, \ie{},
to minimize $G+L\norm{\wt}$ subject to the
constraints $\Set{G\le G_{\max}, L\le L_{\max}, G+L\norm{\wt}\ge \norm{g_{t}}}$.
However, there may be many such $(G,L)$ satisfying these conditions,
and in general it is unclear whether there exists a
general-purpose strategy to choose among them without further assumptions.
Indeed,
\Cref{thm:qb-adaptive-regret} suggests that
when $\norm{\cmp}$ is very large, we'd prefer to set
the $L_{t}$'s smaller at the expense of large $G_{t}$'s,
and vice-versa when $\norm{\cmp}$ is sufficiently small,
so optimally trading off $G_{t}$ and $L_{t}$ would require
some prior knowledge about $\norm{\cmp}$.

Nevertheless,
there are many situations in which one can
choose $(G_{t},L_{t})$ pairs along
some pareto-frontier.
As an illustrative example, consider
an online regression setting in which
$\ell_{t}(\w)=\half(y_{t}-\inner{\xt,\w})^{2}$ for some target variable $y_{t}\in\R$
and feature vector $x_{t}\in\R^{d}$.
Observe that $\grad\ell_{t}(\wt)=-(y_{t}-\inner{x_{t},\wt})x_{t}$,
so setting $G_{t}=\abs{y_{t}}\norm{x_{t}}$ and
$L_{t}=\abs{\inner{x_{t},\wt/\norm{\wt}}}\norm{x_{t}}$,
we have
\begin{align*}
  \norm{\grad\ell_{t}(\wt)}=\norm{(y_{t}-\inner{x_{t},\wt})x_{t}}\le G_{t}+L_{t}\norm{\wt},
\end{align*}
so $\Set{\grad\ell_{t}(\wt)}$ is a $(G_{t},L_{t})$-quadratically bounded sequence
\wrt{} $\Set{\wt}$,
and \Cref{thm:qb-adaptive-regret} quarantees regret scaling as
\begin{align*}
  \tilde O \brac{\mspace{-3mu}\norm{\cmp}\mspace{-3mu}\sqrt{\sumtT \mspace{-2mu}y_{t}^{2}\norm{x_{t}}^{2}}
    +\norm{\cmp}^{2}\mspace{-5mu}\sqrt{\sumtT\mspace{-3mu}\left\langle x_{t},\frac{\wt}{\norm{\wt}}\right\rangle^{2} \mspace{-5mu}\norm{x_{t}}^{2}}},
\end{align*}
which is more adaptive to sequence of observed feature vectors $x_{t}$ and targets
$y_{t}$ than the worst-case
bound
of $R_{T}(\cmp)\le \tilde O\brac{\norm{\cmp}\abs{y_{\max}}\norm{x_{\max}}\sqrt{T}+\norm{\cmp}^{2}\norm{x_{\max}}^{2}\sqrt{T}}$.

Finally, notice that for $L_{\max}>0$ \Cref{alg:qb-adaptive-alg} suffers
an additional $O(L_{\max}\norm{\cmp}^{2}\sqrt{T})$ penalty which
is not present in the Lipschitz losses setting.
The following theorem demonstrates that this penalty is in fact unavoidable
in our problem setting. Proof can be found in \Cref{app:qb-olo-static-lb}.

\begin{restatable}{theorem}{QBOLOStaticLB}\label{thm:qb-olo-static-lb}
Let $\cA$ be an algorithm defined over $\R^{2}$ and let $\wt$ denote the output
of $\cA$ on round $t$. Let $\epsilon>0$ and suppose $\cA$ guarantees $R_{T}(\zeros)\le \epsilon$
against any quadratically bounded sequence $\Set{g_{t}}$. Then for any $T\ge 1$, $G>0$ and $L\ge 0$ there exists a sequence $g_{1},\ldots,g_{T}$ satisfying
$\norm{g_{t}}\le G+L\norm{\wt}$
and a comparator $\cmp\in\R^{2}$
such that
\begin{align*}
  R_{T}(\cmp)\ge \Omega\brac{G\norm{\cmp}\sqrt{T\Log{\norm{\cmp}\sqrt{T}/\epsilon}}\maxOp L\norm{\cmp}^{2}\sqrt{T}}.
\end{align*}
\end{restatable}

\begin{remark}

  An alternative way to approach online learning in
  our problem setting would be
  to apply an algorithm which is both comparator-adaptive and
  Lipschitz-adaptive, since these algorithms do not require
  an \textit{a priori} upper bound on $\norm{\cmp}$ nor on $\norm{g_{t}}$.
  \Cref{thm:qb-olo-static-lb} demonstrates that this approach would be
  sub-optimal in our setting. Indeed, \citet{mhammedi2020lipschitz}
  show that without prior knowledge of a Lipschitz bound,
  there is an unavoidable
  $O(\norm{\cmp}^{3}\max_{t\le T}\norm{g_{t}})$
  penalty
  associated with comparator-norm adaptivity,
  which can lead to a sub-optimal
  $O(\norm{\cmp}^{3}L\max_{t}\norm{\w_{t}})\ge O(L\norm{\cmp}^{3}\sqrt{T})$
  dependence in our problem setting.

\end{remark}

\section{\SecSaddlePoint}
\label{sec:sp}

As a result of the algorithm in the previous section, we are immediately
able to produce a novel algorithm for saddle-point optimization
in unbounded domains. Consider the following convex-concave
saddle-point problem:
\begin{align}
  \inf_{x\in\cX}\sup_{y\in\cY}\cL(x,y)\label{eq:sp}
\end{align}
where $\cX$ and $\cY$ are convex sets, $\cL(\cdot,y)$ is convex for
all $y\in\cY$, and $\cL(x,\cdot)$ is concave for all $x\in\cX$.
Solutions to \Cref{eq:sp} are captured by the
notion of a \textit{saddle-point}:
a point $(\xstar,\ystar)\in\cX\times\cY$ is called saddle-point of $\cL$ if
for any $(x,y)\in\cX\times\cY$
it satisfies
\begin{align*}
  \cL(\xstar,y)\le \cL(\xstar,\ystar)\le\cL(x,\ystar)
  .
\end{align*}
When such a point exists,
it satisfies
$\cL(\xstar,\ystar)=\inf_{x\in\cX}\sup_{y\in\cY}\cL(x,y)=\sup_{y\in\cY}\inf_{x\in\cX}\cL(x,y)$.
Quality of a candidate solution $(x,y)\in\cX\times\cY$
is commonly measured in terms of the \textit{duality gap}:
\begin{align*}
  G(x,y)\defeq\sup_{\ystar\in\cY}\cL(x,\ystar)-\inf_{\xstar\in\cX}\cL(\xstar,y).
\end{align*}
It can be shown that the duality gap is non-negative, and that
any $(x,y)\in\cX\times\cY$ such that $G(x,y)=0$ must be a saddle-point \citep{boyd2004convex}.
Fortunately, any such gap is easily controlled
using an online learning algorithm via the well-known
reduction to Online Linear Optimization (OLO) shown in \Cref{alg:sp-redux}.
\begin{algorithm}[t!]
  \caption{Saddle-point Reduction}
  \label{alg:sp-redux}
  \begin{algorithmic}
    \STATE\textbf{Input} Domain $W=\cX\times\cY$, OLO Algorithm $\cA$
    \For{$t=1:T$}{
      \STATE Get $\wt=(x_{t},y_{t})\in W$ from $\cA$
      \STATE Receive $g_{t}^{x}\in \partial_{x}\cL(x_{t},y_{t})$ and
      $g_{t}^{y}\in\partial_{y}[-\cL(x_{t},y_{t})]$
      \STATE Send $g_{t}=(g_{t}^{x},g_{t}^{y})$ to $\cA$ as the $t^{\text{th}}$ subgradient
    }
    \STATE \textbf{Return} $\overline{w}_{T}=\brac{\frac{\sumtT x_{t}}{T},\frac{\sumtT y_{t}}{T}}$
  \end{algorithmic}
\end{algorithm}
\newcommand{\spcmp}{\mathring{\w}}
\newcommand{\spx}{\mathring{\x}}
\newcommand{\spy}{\mathring{\y}}
\begin{restatable}{lemma}{FolkTheorem}\label{lemma:folk-theorem}
  For any $\spcmp=(\spx,\spy)\in\cX\times\cY$, \Cref{alg:sp-redux} guarantees
  \begin{align*}
    \cL(\xbar_{T},\mathring{y})-\cL(\mathring{x},\ybar_{T})\le \frac{\sumtT\inner{g_{t},\wt-\spcmp}}{T}=\frac{R_{T}^{\cA}(\spcmp)}{T}.
  \end{align*}
\end{restatable}
\begin{proof}
To see why this is true, observe that by convexity of $x\mapsto\cL(x,y)$ and
$y\mapsto-\cL(x,y)$, we can apply Jensen's inequality in both arguments to get:
\begin{align*}
      &\cL(\xbar_{T},\spy)-\cL(\spx,\ybar_{T})\\
  &\quad\le
      \frac{1}{T}\sbrac{\sumtT \cL(x_{t},\spy)-\cL(\spx,y_{t})}\\
  \intertext{now add and subtract $\cL(x_{t},y_{t})$:}
  &\quad=
      \frac{\sumtT \cL(x_{t},y_{t})-\cL(\spx,y_{t})-\cL(x_{t},y_{t})+\cL(x_{t},\spy)}{T}\\
  \intertext{let $g_{t}^{x}\in\partial_{x} \cL(x_{t},y_{t})$ and
  $g_{t}^{y}\in\partial_{y}[-\cL(x_{t},y_{t})]$ and again use convexity to upper bound
  both difference terms:}
    &\quad\le
      \frac{\sumtT \inner{g_{t}^{x},x_{t}-\spx}+\inner{g_{t}^{y}, y_{t}-\spy}}{T}\\
  \intertext{and now define  $\wt=(x_{t},y_{t})$, $\spcmp=(\spx,\spy)$, and
  $g_{t}=(g_{t}^{x},g_{t}^{y})$ to complete the proof:}
  &\quad=
    \frac{\sumtT \inner{g_{t},\wt-\spcmp}}{T}
    =
      \frac{R_{T}^{\cA}(\spcmp)}{T}.
\end{align*}
\end{proof}
Thus in order to control the duality gap $G(x,y)$, it suffices to provide
any OLO algorithm that achieves sublinear regret under the given assumptions.

To the best of our knowledge,
the only existing work to
achieve a comparator-adaptive convergence guarantee for the duality gap
in general saddle-point problems is
\citet{liu2022initialization}.
Their approach does indeed guarantee a rate of
the form $G(\xbar_{T},\ybar_{T})\le \frac{R_{T}^{\cA}(\w^{*})}{T}\le\tilde O\brac{\frac{G\norm{\w^{*}}}{\sqrt{T}}}$
under the assumption that the $\cL(\cdot,\cdot)$ is $G$-Lipschitz
in both arguments,
which is justified by assuming that $\cX$ and $\cY$
are bounded domains. However, generally
saddle-point problems can have some
coupling between the $x\in \cX$ and $y\in \cY$, leading to factors of
$\norm{x}$ and $\norm{y}$ showing up in both $\norm{\grad_{x}\cL(x,y)}$ and $\norm{\grad_{y}\cL(x,y)}$.
Thus, even in a bounded domain a bound of the form
$R_{T}^{\cA}(\w^{*})\le \tilde O\brac{\norm{\w^{*}}G\sqrt{T}}$ actually still
falls short of being fully comparator-adaptive because the Lipschitz constant $G$ is
subtly hiding factors of $D_{\cX}=\max_{x,x'\in\cX}\norm{x-x'}$ and
$D_{\cY}=\max_{y,y'\in\cY}\norm{y-y'}$. See Section~\ref{sec:bilinear} for a more explicit example of this issue.

On the other hand, for many interesting problems
$\cL(\cdot,\cdot)$ is quadratically bounded in both arguments,
which will enable us to immediately apply \Cref{alg:qb-adaptive-alg}
to the linear losses $g_{t}=(g_{t}^{x},g_{t}^{y})$ as described above.
In particular, we have the following:

\begin{minipage}{\columnwidth}
\begin{restatable}{proposition}{QBSPGeneral}\label{prop:qbsp-general}
  Suppose that for all $\tilde y\in\cY$,
  the function $x\mapsto \cL(x,\tilde y)$ is
  $(G_{x}+L_{xy}\norm{\tilde y},L_{xx})$-quadratically
  bounded, and for all $\tilde x\in\cX$
  the function $y\mapsto -\cL(\tilde x,y)$ is
  $(G_{y}+L_{yx}\norm{\tilde x},L_{yy})$-quadratically bounded.
  Let $g_{t}^{x}\in\partial_{x}\cL(x_{t},y_{t})$ and
  $g_{t}^{y}\in\partial_{y}[-\cL(x_{t},y_{t})]$, and set $g_{t}=(g_{t}^{x},g_{t}^{y})$.
  Then $\Set{g_{t}}$ is a
  $(G_{w},L_{w})$-quadratically bounded sequence
  \wrt{} norm $\norm{(x,y)}=\sqrt{\norm{x}^{2}+\norm{y}^{2}}$,
  where $G_{w}\le O\brac{\sqrt{G_{x}^{2}+G_{y}^{2}}}$ and $L_{w}\le O\brac{\sqrt{L_{xx}^{2}+L_{yy}^{2}+L_{xy}^{2}+L_{yx}^{2}}}$.
\end{restatable}
\end{minipage}
\begin{proof}
  Let $(x,y)\in W$. For $g_{x}\in \partial_{x} \cL(x,y)$ observe that
  \begin{align*}
    \norm{g_{x}}^{2}
    &\le (G_{x}+L_{xy}\norm{y}+L_{xx}\norm{x})^{2}\\
    &\le 5\brac{G_{x}^{2}+L_{xy}^{2}\norm{y}^{2}+L_{xx}^{2}\norm{x}^{2}},
  \end{align*}
  where the first line uses the assumption that $x\mapsto \cL(x,y)$ is
  $(G_{x}+L_{xy}\norm{y},L_{xx})$ quadratically bounded for any $y\in \cY$ and
  the last line uses $(a+b+c)^{2}\le 5a^{2}+5b^{2}+5c^{2}$.
  Likewise,
  \begin{align*}
    \norm{g_{y}}^{2}
    &\le
      5\brac{G_{y}^{2}+L_{yx}^{2}\norm{x}^{2}+L_{yy}^{2}\norm{y}^{2}},
  \end{align*}
  and so overall, letting $g_{w}=(g_{x},g_{y})$ we have
  \begin{align*}
    \norm{g_{w}}
    &=
      \sqrt{\norm{g_{x}}^{2}+\norm{g_{y}}^{2}}\\
    &\overset{(\star)}{\le}
      \underbrace{\sqrt{5}\sqrt{G_{x}^{2}+G_{y}^{2}}}_{=:G_{w}}\\
    &\quad
      +\underbrace{\sqrt{5}\sqrt{L_{xx}^{2}+L_{yy}^{2}+L_{xy}^{2}+L_{yx}^{2}}}_{L_{w}}\sqrt{\norm{x}^{2}+\norm{y}^{2}}\\
    &=
      G_{w}+L_{w}\norm{w}
  \end{align*}
  where $(\star)$ uses
  $\sqrt{x+y}\le \sqrt{x}+\sqrt{y}$ for $x,y\ge 0$.
\end{proof}
Hence, with this in hand we can use \Cref{alg:qb-adaptive-alg}
to guarantee that for any $\spcmp=(\spx,\spy)\in W$,
\begin{align*}
  &\cL(\xbar_{T},\spy)-\cL(\spx,\ybar_{T})\\
  &\qquad\overset{\Cref{lemma:folk-theorem}}{\le }
    \frac{R_{T}^{\cA}(\spcmp)}{T}\\
  &\qquad\overset{~\Cref{thm:qb-adaptive-regret}}{\le}
    \tilde O\brac{\frac{G_{w}\norm{\spcmp}+L_{w}\norm{\spcmp}^{2}}{\sqrt{T}}},
\end{align*}
which is indeed fully adaptive to comparator $\spcmp$.

It is important
to note that our results in this section are made possible because our algorithm works even in the
more difficult QB-OLO setting.
It may be possible to get a similar result by using two QB-OCO algorithms
designed for quadratically bounded functions $\ell_{t}$, though it
it seems more challenging. In particular, letting $\ell_{t}^{x}(\cdot)=\cL(\cdot,y_{t})$ and
$\ell_{t}^{y}(\cdot)=-\cL(x_{t},\cdot)$, one might instead run separate algorithms
against the quadratically bounded loss sequences $\ell_{t}^{x}$ and $\ell_{t}^{y}$.
However, now both algorithms need to very carefully regularize their iterates
such that the gradients received by the other algorithm are never too large,
since $\norm{\grad\ell_{t}^{x}(x_{t})}$ may can contain factors of $\norm{y_{t}}$
and $\norm{\grad\ell_{t}^{y}(y_{t})}$ can contain factors of $\norm{x_{t}}$. Hence
careful coordination between the two algorithms will be required.
The upshot is that by using un-linearized losses $\ell_{t}^{x}$ and $\ell_{t}^{y}$
it may be possible to get faster rates in some settings by better accounting
for local curvature. We leave this as an exciting direction for future investigation.

\subsection{Example: Bilinearly-coupled saddle-points}\label{sec:bilinear}
Before moving on, let us make things less abstract with a simple example. Consider a \textit{bilinearly-coupled saddle-point}
problem of the form
\begin{align}
  \cL(x,y)= F_{x}(x)+H(x,y)-F_{y}(y)\label{eq:bilinearly-coupled}
\end{align}
where $F_{x}$ and $F_{y}$ are convex and $(\tilde G_{x},\tilde L_{x})$ and
$(\tilde G_{y},\tilde L_{y})$-quadratically bounded respectively, and
$H(x,y)=\inner{x,By}-\inner{u_{x},x}+\inner{u_{y},y}$
for some \textit{coupling matrix} $B$ and vectors $u_{x}$, $u_{y}$.
This problem captures several notable problem settings, such as
minimizing the mean-squared projected bellman error for off-policy policy
evaluation in reinforcement learning,
quadratic games, and regularized empirical risk minimization \citep{du2022optimal}.
The following proposition demonstrates that these problems do indeed
satisfy the conditions of \Cref{prop:qbsp-general}.
\begin{restatable}{proposition}{BilinearlyCoupled}\label{prop:bilinearly-coupled}
  \Cref{eq:bilinearly-coupled} satisfies the assumptions of
  \Cref{prop:qbsp-general}
  with
  $G_{x}=\tilde G_{x}+\norm{u_{x}}$, $L_{xx}=\tilde L_{x}$, $L_{xy}=\opnorm{B}$,
  $G_{y}=\tilde G_{y}+\norm{u_{y}}$, $L_{yy}=\tilde L_{y}$, and
  $L_{yx}=\opnorm{B^{\top}}$.
\end{restatable}
\begin{proof}
  Observe that for any $(x,y)\in\cX\times\cY$ and $g_{x}\in\partial_{x}\cL(x,y)$, we have
  \begin{align*}
    \norm{g_{x}}
    &=
      \norm{\grad F_{x}(x)+By-u_{x}}\\
    &\le
      \norm{\grad F_{x}(x)}+\opnorm{B}\norm{y}+\norm{u_{x}}\\
    &\le
      \tilde G_{x}+\norm{u_{x}}+\tilde L_{x}\norm{x}+\opnorm{B}\norm{y},
  \end{align*}
  where $\grad F_{x}(x)\in\partial F_{x}(x)$ and $\opnorm{B}$ denotes the operator
  norm $\opnorm{B}=\sup_{x:\norm{x}=1}\norm{Bx}$.
  Likewise,
  \begin{align*}
    \norm{g_{y}}
    &\le
      \tilde G_{y}+\norm{u_{y}}+\tilde L_{y}\norm{y}+\opnorm{B^{\top}}\norm{x}.
  \end{align*}
  Hence, $\cL(\cdot,\cdot)$ satisfies the assumptions of \Cref{prop:qbsp-general} with
  $G_{x}=\tilde G_{x}+\norm{u_{x}}$, $L_{xx}=\tilde L_{x}$, $L_{xy}=\opnorm{B}$,
  $G_{y}=\tilde G_{y}+\norm{u_{y}}$, $L_{yy}=\tilde L_{y}$, and
  $L_{yx}=\opnorm{B^{\top}}$.
\end{proof}

We note that this specific example is mainly for illustrative purposes ---
in many instances of \Cref{eq:bilinearly-coupled} the functions $F_{x}$
and $F_{y}$ satisfy stronger curvature assumptions than used here,
and our approach would be improved by more explicitly leveraging
these assumptions when they hold.
Nevertheless, our approach here does have a few key benefits: first, we naturally attain
convergence in duality gap with an explicit dependence on
the comparator,  whereas prior works generally
only attain a bound of this form making stronger
assumptions such as strong convexity
or one of the boundedness assumptions we're seeking to avoid
\citep{liu2022initialization,du2022optimal,ibrahim2020linear,azizian2020accelerating}.
Second, our approach can be applied under fairly weak assumptions: $\cL(\cdot,\cdot)$
need not be Lipschitz, strongly-convex, nor smooth in either argument,
and we do not require $\cX\times\cY$ to be a bounded domain.

\section{\SectionDynamic}
\label{sec:dynamic}

Next, we turn our attention to \emph{dynamic} regret.
In the static regret setting, we saw in \Cref{sec:quadratically-bounded} that to
control the stability of the algorithm it was necessary
to add an additional term $\Phi_{t}(\w)=O\brac{L_{\max}\sqrt{T}\norm{\w}^{2}}$
to the regularizer to help control the ``non-Lipschitz''
part of the loss.
We will likewise need a stronger regularizer
to control the gradients for dynamic regret, but now it will
lead to difficulties.
To see why, consider the dynamic regret of gradient descent with
a fixed step-size $\eta$. Using existing analyses one can arrive at
\begin{align}
  R_{T}(\vec{\cmp})
  &\le
    O\brac{\frac{\norm{\cmp_{T}}^{2}+\max_{t}\norm{\w_{t}}P_{T}}{2\eta} +\frac{\eta}{2}\sumtT\norm{g_{t}}^{2}},\label{eq:gd-dynamic}
\end{align}
where $g_{t}\in\partial\ell_{t}(\wt)$ and
$P_{T}=\sum_{t=2}^{T}\norm{\cmp_{t}-\cmp_{\tmm}}$. In a bounded domain of
diameter $D$, we can bound $\norm{\cmp_{T}}^{2}\le D^{2}$ and
$\max_{t}\norm{\wt}\le D$, and then by optimally tuning $\eta$
we get $R_{T}(\vec{\cmp})\le O\bigg(\sqrt{(D^{2}+DP_{T})\sumtT\norm{g_{t}}^{2}}\bigg)$ which
is optimal in the Lipschitz loss setting \citep{zhang2018adaptive}.
More generally, using Mirror Descent with regularizer $\psi(\w)$, one can
derive an analogous bound:
\begin{align*}
  R_{T}(\vec{\cmp})\le O\brac{\psi(\cmp_{T})+\max_{t}\norm{\grad \psi(w_{t})}P_{T}+\sumtT\delta_{t}},
\end{align*}
where $\delta_{t}$ are the stability terms discussed in \Cref{sec:quadratically-bounded}.
In an unbounded domain, \citet{jacobsen2022parameter} use
a regularizer of the form
$\psi(\w)=O\brac{\norm{\w}\Log{\norm{\w}T}/\eta}$, which enables
them  to bound $\sumtT\delta_{t}\le O(\eta)$, and moreover, they show that
$\max_{t}\norm{\grad \psi_{t}(\wt)}$ can be bound from above by $O\brac{\Log{MT/\epsilon}/\eta}$
after adding a composite penalty $\varphi_{t}(\w)=\eta\norm{g_{t}}^{2}\norm{\w}$ to the update.
Then optimally tuning $\eta$ leads to
regret scaling as
\begin{align}
R_{T}(\vec{\cmp})\le \tilde O\brac{\sqrt{(M^{2}+MP_{T})\sumtT\norm{g_{t}}^{2}}},\label{eq:unbounded-pf-dynamic}
\end{align}
where $M=\max_{t}\norm{\cmp_{t}}$,
which matches the bound from
the bounded-domain setting up to logarithmic terms.

In our setting, the situation gets significantly more challenging.
As in \Cref{sec:quadratically-bounded}, we will need to
include an $O(\norm{\w}^{2}/\eta)$ term in the regularizer $\psi_{t}$
in order to control the ``non-Lipschitz'' part of the loss function.
However, as above this leads to coupling
$\max_{t}\norm{\grad \psi_{t}(\wt)}P_{t}=\max_{t}\norm{\wt}P_{T}/\eta$ in the dynamic regret,
and the term
$\max_{t}\norm{\wt}$ is generally too large to cancel out with additional
regularization as done by \citet{jacobsen2022parameter}.
Even more troubling is that our lower bound in \Cref{thm:dynamic-lb}
suggests that the ideal dependence would be $O(MP_{T}/\eta)$,
which we can only hope to achieve by constraining $\norm{\wt}$
to a ball of diameter proportional to $M=\max_{t}\norm{\cmp_{t}}$.
Yet $M$ is unknown to the learner!

\begin{algorithm}[t!]
    \caption{Dynamic Regret Algorithm}
    \label{alg:dynamic-meta}
    \begin{algorithmic}
      \STATE\KwInput{$G_{\max}$, $L_{\max}$,
        weights $\beta_{1},\ldots,\beta_{T}$ in $[0,1]$,
        hyperparameter set
        $\cS=\Set{(\eta,D):\eta\le\frac{1}{8 L_{\max}}, D>0}$,
        $p_{1}\in\Delta_{\abs{\cS}}$.
      }
      \For{$\tau=(\eta,D)\in\cS$}{
        \STATE\KwInitialize{$\w_{1}^{(\tau)}=\zeros$, $q_{1}(\tau)=p_{1}(\tau)$}
        \STATE\textbf{Define}
        $\mu_{\tau}=\frac{1}{2D\brac{G_{\max}+D/\eta}}$
        \STATE\textbf{Define}
         $\psi_{\tau}(x)=\frac{9}{2 \mu_{\tau}}\int_{0}^{x}\Log{v}dv$
      }
      \For{$t=1:T$}{
        \STATE Play $\wt = \sum_{\tau\in\cS}p_{t}(\tau)\wt^{(\tau)}$
        \STATE Query $\ell_{t}(\wtilde_{t})$ for any $\wtilde_{t}\in W$ s.t.
        $\norm{\wtilde_{t}}\le D_{\min}$
        \For{$\tau=(\eta,D)\in\cS$}{
        \STATE Query $g_{t}^{(\tau)}\in\partial\ell_{t}(\wt^{(\tau)})$
        \STATE Set
        \(
          \wtpp^{(\tau)}=\mspace{-18mu}\operatornamewithlimits{\proj}\limits_{\Set{\w\in W:\norm{\w}\le D}}\brac{\wt^{(\tau)}-\eta(1+8\eta L_{t})g_{t}^{(\tau)}}
        \)
        \STATE Define $\elltilde_{t,\tau}=\ell_{t}(\wt^{(\tau)})-\ell_{t}(\wtilde_{t})$
        }
        \STATE Set
        \(
          q_{\tpp}=\operatornamewithlimits{argmin}\limits_{q\in\Delta_{\abs{\cS}}}\sum\limits_{\tau\in\cS}(\elltilde_{t\tau}+\mu_{\tau}\elltilde_{t\tau}^{2})q_{\tau}+D_{\psi_{\tau}}(q_{\tau}|p_{t\tau})
        \)
        \STATE Set $p_{\tpp}=(1-\beta_{t})q_{\tpp}+\beta_{t} p_{1}$.
      }
    \end{algorithmic}
\end{algorithm}

Luckily, hope is not all lost. Taking inspiration from \citet{luo2022corralling},
we can still attain a bound similar to \Cref{eq:unbounded-pf-dynamic} by tuning the diameter of
an artificial domain constraint. The approach is as follows: for each $(\eta,D)$ in some
set $\cS$, we run an instance of gradient descent $\cA(\eta,D)$ which uses step-size $\eta$ and
projects to the set $W_{D}=\Set{\w\in W:\norm{\w}\le D}$.
Then, using a carefully designed experts algorithm, it is possible
to ensure that the overall regret of the algorithm
scales roughly as
$R_{T}(\vec{\cmp})\le \tilde O\brac{R_{T}^{\cA(\eta,D)}(\vec{\cmp})}$ for
\textit{any} $(\eta,D)\in \cS$.
Thus if we can ensure that there is \emph{some} $(\eta,D)\in\cS$
for which $D\approx M$ and $\eta$ is near-optimal, then
we'll be able to achieve dynamic regret with the
desired $MP_{T}$ dependence.
The following theorem, proven in \Cref{app:dynamic}, characterizes an
algorithm which achieves dynamic regret analogous to
the above bounds,
and in \Cref{thm:dynamic-lb} we show that this is indeed unimprovable. Notably, our result
also \textit{automatically} improves to a novel $L^{*}$ bound when the losses
are smooth.

\begin{restatable}{theorem}{Dynamic}\label{thm:dynamic}
  For all $t$ let $\ell_{t}:W\to \R$ be a $(G_{t},L_{t})$-quadratically bounded
  convex function with $G_{t}\in[0,G_{\max}]$ and $L_{t}\in[0,L_{\max}]$.
  Let $\epsilon>0$, $\beta_{t}\le 1-\Exp{-1/T}$ for all $t$, and for any $i,j\ge 0$ let
  $D_{j}=\frac{\epsilon}{T} \sbrac{2^{j}\minOp 2^{T}}$ and
  $\eta_{i}=\sbrac{\frac{\epsilon 2^{i}}{8\brac{G_{\max}+\epsilon L_{\max}}T}\minOp\frac{1}{8L_{\max}}}$,
  and let $\cS=\Set{(\eta_{i},D_{j}):i,j\ge 0}$.
  For each $\tau=(\eta,D)\in\cS$ let
  $\mu_{\tau}=\frac{1}{2D\brac{G_{\max}+D/\eta}}$
  and set
  $p_{1}(\tau)=\frac{\mu_{\tau}^{2}}{\sum_{\tilde\tau\in\cS}\mu_{\tilde\tau}^{2}}$.
  Then for any
  $\vec{\cmp}=(\cmp_{1},\ldots,\cmp_{T})$ in $W$,
  \Cref{alg:dynamic-meta} guarantees
  \begin{align*}
    R_{T}(\vec{\cmp})
    \mspace{-3mu}
    &\le
      \mspace{-3mu}O\Bigg(
    G_{\max}(M+\epsilon)\Lambda_{T}^{*}+L_{\max}(M+\epsilon)^{2}\Lambda_{T}^{*}\\
    &\quad\qquad
      +G_{\max}P_{T}+L_{\max}(M+\epsilon)P_{T}\\
    &\quad\qquad
      +\sqrt{(M^{2}\Lambda_{T}^{*}+MP_{T})\Omega_{T}},
      \Bigg).
  \end{align*}
  where
  $\Lambda_{T}^{*}\le O\brac{\Log{\frac{MT\Log{T}}{\epsilon}}+\Log{\Log{\frac{G_{\max}}{\epsilon L_{\max}}}}}$,
  $\Omega_{T}\le \sumtT \sbrac{G_{t}^{2}+L_{t}^{2}M^{2}}$,
  $P_{T}=\sum_{t=2}^{T}\norm{\cmp_{t}-\cmp_{\tmm}}$, and
  $M=\max_{t}\norm{\cmp_{t}}$.
  Moreover,
  when the losses
  are $L_{t}$-smooth, $\Omega_{T}$ automatically improves to
  $\Omega_{T}\le \Min{\sumtT L_{t}\sbrac{\ell_{t}(\cmp_{t})-\ell_{t}^{*}}, \sumtT G_{t}^{2}+L_{t}^{2}M^{2}}$.
\end{restatable}

Hiding constants and logarithmic terms, the bound is effectively
\begin{align*}
  &R_{T}(\vec{\cmp})
    \le \tilde O\brac{\sqrt{\brac{M^{2}+MP_{T}}\sumtT G_{t}^{2}+L_{t}^{2}M^{2}}}.
\end{align*}
Notice that our result again generalizes the bounds established in prior works.
Unfortunately, the result is not a
strict generalization as \Cref{thm:dynamic} requires $L_{\max}>0$
for the hyperparameter set $\cS$ to be finite.
To achieve a strict generalization, one can simply
define a procedure which runs the algorithm of \citet{jacobsen2022parameter}
when $L_{\max}=0$ and \Cref{alg:dynamic-meta} otherwise; this
is possible because $L_{\max}$ must be provided as input to the algorithm. Notably, the algorithm of
\citet{jacobsen2022parameter} does not use the aforementioned domain tuning
trick and requires significantly less per-round computation as a result
($O(d\Log{T})$ vs. $O(dT\Log{T})$). We leave open the question of whether the exists a unifying analysis for $L_{\max}=0$ and $L_{\max}>0$, and whether the per-round computation can be improved.

As in \Cref{sec:quadratically-bounded}, we
again observe an additional penalty associated with non-Lipschitzness,
this time on the
order of
$\tilde O\brac{M^{3/2}\sqrt{(M+P_{T})\sumtT L_{t}^{2}}}$. The following
theorem shows that these penalties are unavoidable in general (proof in \Cref{app:dynamic-lb}).

\begin{restatable}{theorem}{DynamicLB}\label{thm:dynamic-lb}
  For any $M>0$ there is a sequence
  of $(G,L)$-quadratically bounded functions with $\frac{G}{L}\le M$
  such that for any $\gamma\in[0,\half]$,
  \begin{align*}
      R_{T}(\vec{\cmp})\ge \Omega\brac{GM^{1-\gamma}\sbrac{P_{T}T}^{\gamma}+LM^{2-\gamma}\sbrac{P_{T}T}^{\gamma}}.
  \end{align*}
  where $P_{T}=\sum_{t=2}^{T}\norm{\cmp_{t}-\cmp_{\tmm}}$
  and $M\ge\max_{t}\norm{\cmp_{t}}$.
\end{restatable}

Notice that with $\gamma=\half$, we have
$R_{T}(\vec{\cmp})\ge \Omega\brac{G\sqrt{MP_{T}T}+LM^{3/2}\sqrt{P_{T}T}}$,
matching our upper bound in \Cref{thm:dynamic} up to logarithmic terms.
On the otherhand, for
$\gamma=0$ we have
$R_{T}(\vec{\cmp})\ge GM+LM^{2}$,
suggesting that the lower-order leading terms of our upper bound are also necessary.
We also note that the assumption $G/L \le M$ is without loss of generality: when
$G/L\ge M$ one can construct a sequence of $(G+LM)$-Lipschitz losses
according to existing lower bounds to show that
\begin{align*}
  R_{T}(\vec{\cmp})&\ge \Omega\brac{(G+LM)\sqrt{MP_{T}T}}\\
  &=\Omega\brac{G\sqrt{MP_{T}T}+LM^{3/2}\sqrt{P_{T}T}}.
\end{align*}

Interestingly, when the losses are smooth, the bound in \Cref{thm:dynamic} has the appealing
property that it automatically improves to an $L^{*}$
bound of the form
\begin{align*}
  R_{T}(\vec{\cmp})&\le\tilde O\brac{\sqrt{(M^{2}+MP_{T})\sumtT L_{t}\sbrac{\ell_{t}(\cmp_{t})-\ell_{t}^{*}}}},
\end{align*}
which matches bounds established in the Lipschitz and bounded domain setting
up to logarithmic penalties
\citep{zhao2020dynamic}.
This is the first $L^{*}$ bound that we are aware of to be achieved
in an unbounded domain for general smooth losses without a Lipschitz or bounded-range assumption.
Moreover, our bound features improved adaptivity to the \textit{individial} $L_{t}$'s,
scaling as
$\sumtT L_{t}\sbrac{\ell_{t}(\cmp_{t})-\ell_{t}^{*}}$
instead of the usual $L_{\max}\sumtT\ell_{t}(\cmp_{t})-\ell_{t}^{*}$ achieved by
prior works \citep{srebro2010smoothness,orabona2012beyond,zhao2020dynamic}.

On the other hand, our
upper bound bound contains terms of the form $\frac{G_{\max}}{L_{\max}\epsilon}$.
Such ratios are unappealing in general because $G_{\max}$ and $L_{\max}$
are not under our control --- it's possible for this ratio
to be arbitrarily large. Fortunately, this ratio only shows
up only in doubly-logarithmic terms, and hence these penalties
can be regarded as effectively constant as far as the regret
bound is concerned.

A more pressing issue is that the ratio $\frac{G_{\max}}{\epsilon L_{\max}}$
shows up in the number of experts. That is, setting
$\cS$ as in \Cref{thm:dynamic} requires a collection of
$O(T\log_{2}(\sqrt{T})+T\log_{2}\brac{G_{\max}/L_{\max}\epsilon})$
experts, so in practice we can only tolerate
$G_{\max}/L_{\max}\epsilon\le \text{poly}(T)$ without increasing the
(already quite high!) order of computation.
We note that any algorithm that guarantees
$R_{T}(\zeros)\le G_{\max}\epsilon$ can't hope to
ensure non-vacuous regret when $G_{\max}/L_{\max}\epsilon > T$ anyways,
so this seems to be a fundamental restriction in this setting.
Nevertheless, the following result shows that if we know
\textit{a priori} that the losses will be smooth, then
we can avoid this $\Log{\Log{\frac{G_{\max}}{L_{\max}\epsilon}}}$ penalty
entirely and reduce the number of experts to $T\log_{2}(\sqrt{T})$
by instead setting $\eta_{\min}\propto\frac{1}{L_{\max}\sqrt{T}}$. Proof can be
found in \Cref{app:smooth-dynamic}.

\begin{restatable}{theorem}{SmoothDynamic}\label{thm:smooth-dynamic}
  For all $t$ let $\ell_{t}:W\to \R$ be  $(G_{t},L_{t})$-quadratically bounded
  and $L_{t}$-smooth convex function with $G_{t}\in[0,G_{\max}]$ and $L_{t}\in[0,L_{\max}]$.
  Let $\epsilon>0$ and for any $i,j\ge 0$ let
  $D_{j}=\frac{\epsilon }{\sqrt{T}}\sbrac{2^{j}\minOp 2^{T}}$ and
  $\eta_{i}=\frac{1}{8L_{\max}\sqrt{T}}\sbrac{2^{i}\minOp \sqrt{T}}$,
  and let $\cS=\Set{(\eta_{i},D_{j}):i,j\ge 0}$.
  Then for any
  $\vec{\cmp}=(\cmp_{1},\ldots,\cmp_{T})$ in $W$,
  \Cref{alg:dynamic-meta} guarantees
  \begin{align*}
    R_{T}(\vec{\cmp})
    \mspace{-3mu}&\le\mspace{-3mu}
      O\Bigg(
       G_{\max}(M+\epsilon) \Lambda_{T}^{*}+L_{\max}(M+\epsilon)^{2}\Lambda_{T}^{*}\\
    &\quad
      +\mspace{-3mu}L_{\max}(M+\epsilon)P_{T}+\sqrt{\sumtT \sbrac{\ell_{t}(\cmp_{t})-\ell_{t}^{*}}^{2}}\\
    &\quad
      +\mspace{-3mu}\sqrt{(M^{2}\Lambda_{T}^{*}\mspace{-3mu}+\mspace{-3mu}MP_{T})\sumtT L_{t}\sbrac{\ell_{t}(\cmp_{t})-\ell_{t}^{*}}}\Bigg),
  \end{align*}
  where
  $\Lambda_{T}^{*}\le O\brac{\Log{\frac{M\sqrt{T}\Log{\sqrt{T}}}{\epsilon}}}$,
  $M=\max_{t}\norm{\cmp_{t}}$, and
  $P_{T}=\sum_{t=2}^{T}\norm{\cmp_{t}-\cmp_{\tmm}}$.
\end{restatable}

\section{Conclusion}%
\label{sec:conclusion}

In this paper, we developed new algorithms for online learning
in unbounded domains with potentially unbounded losses.
We achieve several regret guarantees that have previously
only been attained by assuming Lipschitz losses, losses with bounded range,
a bounded domain, or a combination thereof.
We provide
algorithms for both static and dynamic regret,
as well as an application in saddle-point optimization
leading to new results for unbounded decision sets.
Our lower bounds
show that our results are optimal, and moreover, our algorithms
achieve these results without appealing to any instance-specific
hyperparameter tuning.

There are a few natural directions for future work.
It is still unclear whether
the dynamic regret achieved in \Cref{thm:dynamic} can be achieved
in the more general QB-OLO setting. Moreover, while our dynamic regret algorithm
attains the optimal bound, it requires $O(dT\log(\sqrt{T}))$ computation per round,
whereas the optimal bounds in the Lipschitz loss setting are attained
using $O(d\Log{T})$ per-round computation.
Yet it is unclear
how to achieve the lower bound in \Cref{thm:dynamic-lb} without the artificial
domain trick discussed in \Cref{sec:dynamic}.
We leave these questions as exciting directions for
future work.

\section*{Acknowledgements}

AJ is supported by an NSERC CGS-D
Scholarship. AC is supported by NSF grant CCF-2211718, a Google gift and an Amazon Research Award.

\bibliography{refs}
\bibliographystyle{icml2023}

\newpage

\appendix

\onecolumn

\section{Centered Mirror Descent with Adjustment}

\begin{algorithm}
  \caption{Centered Mirror Descent with Adjustment}
  \label{alg:centered-md-ii}
  \begin{algorithmic}
    \STATE \KwInput{$\w_{1}\in W$, $\psi_{1}:W\to\R_{\ge 0}$}
    \For{$t=1:T$}{
      \STATE Play $\wt\in W$, observe $g_{t}$
      \STATE Choose functions $\psi_{\tpp}$, $\phi_{t}$
      \STATE Define $\Delta_{t}(\w)=D_{\psi_{\tpp}}(\w|\w_{1})-D_{\psi_{t}}(\w|\w_{1})$
      \STATE Update
      $\wtilde_{\tpp} = \argmin_{\wtilde\in W}\inner{g_{t},\tilde \w}+D_{\psi_{t}}(\wtilde|\wt)+\Delta_{t}(\wtilde)+\phi_{t}(\wtilde)$

      \STATE Choose mapping $\cM_{\tpp}:W\to W$
      \STATE Update
      $\wtpp=\cM_{\tpp}(\wtilde_{\tpp})$
    }
  \end{algorithmic}
\end{algorithm}

The key tool we'll use to build our algorithms is a slight generalization of the
Centered Mirror Descent algorithm of \citet{jacobsen2022parameter},
which accounts for an additional post-hoc ``adjustment'' of $\wt$
through the use of an arbitrary mapping $\cM_{t}:W\to W$.
Algorithms of this form have been studied in prior works
such as \citet{gyorgy2016shifting,hall2016online}, wherein
$\cM_{t}$ is interpreted as a dynamical model.
In contrast, we will use $\cM_{t}$ as a convenient way
to formulate a multi-scale version of the fixed-share
update, similar to the generalized share algorithm of \citet{cesa2012mirror}.
Note that when $\cM_{t}$ is the identity mapping, \Cref{alg:centered-md-ii}
is identical to the Centered Mirror Descent algorithm of \citet{jacobsen2022parameter}.

The following lemma provides a regret template for \Cref{alg:centered-md-ii}.
Observe that several of the key terms related to the algorithm's stability
replace the mirror descent iterates $\wtilde_{t}$ with the adjusted
iterates $\wt=\cM_{t}(\wtilde_{t})$. The trade-off is that we must
be careful to ensure that the new penalty terms
$\xi_{t}=D_{\psi_{\tpp}}(\cmp_{t}|\wtpp)-D_{\psi_{\tpp}}(\cmp_{t}|\wtilde_{\tpp})$
are not too large, which places an implicit restriction on how much we can
adjust the iterates via $\cM_{t}$.
\begin{restatable}{lemma}{CenteredMDII}\label{lemma:centered-md-ii}
  For all $t$ let
  $\psi_{t}:W\to\R_{\ge 0}$ be differentiable convex functions,
  $\phi_{t}:W\to \R_{\ge0}$ be subdifferentiable
  convex functions, and let
  $\cM_{t}:W\to W$ be arbitrary mappings.
  Then for any sequence $\vec{\cmp}=(\cmp_{1},\ldots,\cmp_{T})$ in $W$, \Cref{alg:centered-md-ii} guarantees
  \begin{align*}
    R_{T}(\vec{\cmp})
    &\le
      D_{\psi_{T+1}}(\cmp_{T}|\w_{1})-D_{\psi_{T+1}}(\cmp_{T}|\w_{T+1})
      +\sumtT\phi_{t}(\cmp_{t})\\
    &
    +\sum_{t=2}^{T}\underbrace{\inner{\grad\psi_{t}(\wt)-\grad\psi_{t}(\w_{1}),\cmp_{\tmm}-\cmp_{t}}}_{=:\cP_{t}}
      +\sumtT \underbrace{D_{\psi_{\tpp}}(\cmp_{t}|\wtpp)-D_{\psi_{\tpp}}(\cmp_{t}|\wtilde_{\tpp})}_{\xi_{t}}\\
    &
      +\sumtT\underbrace{\inner{g_{t},\wt-\wtilde_{\tpp}}-D_{\psi_{t}}(\wtilde_{\tpp}|\wt)-\Delta_{t}(\wtilde_{\tpp})-\phi_{t}(\wtilde_{\tpp})}_{=:\delta_{t}}.
  \end{align*}
\end{restatable}
\begin{proof}
  Letting $g_{t}\in\partial\ell_{t}(\wt)$, we have
  \begin{align*}
    R_{T}(\vec{\cmp})
    &\le
      \sumtT\inner{g_{t},\wt-\cmp_{t}}
    =
      \sumtT\inner{g_{t},\wtilde_{\tpp}-\cmp_{t}}+\sumtT\inner{g_{t},\wt-\wtilde_{\tpp}}.
  \end{align*}
  From the first-order optimality condition
  $\wtilde_{\tpp}=\argmin_{\wtilde\in W}\inner{g_{t},\wtilde}+D_{\psi_{t}}(\wtilde|\wt)+\Delta_{t}(\wtilde)+\phi_{t}(\wtilde)$,
  for any $\cmp_{t}\in W$ we have
  \begin{align*}
    0&\ge
       \inner{g_{t}+\grad\psi_{t}(\wtilde_{\tpp})-\grad\psi_{t}(\wt),\wtilde_{\tpp}-\cmp_{t}}
       +\inner{\grad\Delta_{t}(\wtilde_{\tpp})+\grad\phi_{t}(\wtilde_{\tpp}),\wtilde_{\tpp}-\cmp_{t}}
  \end{align*}
  so re-arranging,
  \begin{align*}
       \inner{g_{t},\wtilde_{\tpp}-\cmp_{t}}
    &\le
      \inner{\grad\psi_{t}(\wt)-\grad\psi_{t}(\wtilde_{\tpp}),\wtilde_{\tpp}-\cmp_{t}}\\
    &\qquad
      -\inner{\grad\Delta_{t}(\wtilde_{\tpp}),\wtilde_{\tpp}-\cmp_{t}}
      -\inner{\grad\phi_{t}(\wtilde_{\tpp}),\wtilde_{\tpp}-\cmp_{t}}\\
     &\overset{(*)}{\le}
       \inner{\grad\psi_{t}(\wt)-\grad\psi_{t}(\wtilde_{\tpp}),\wtilde_{\tpp}-\cmp_{t}}\\
    &\qquad
      +\phi_{t}(\cmp_{t})-\phi_{t}(\wtilde_{\tpp})\\
    &\qquad
       +\Delta_{t}(\cmp_{t})-\Delta_{t}(\wtilde_{\tpp})-D_{\Delta_{t}}(\cmp_{t}|\wtilde_{\tpp})
  \end{align*}
  where $(*)$ uses the definition of Bregman divergence to write
  $\inner{-\grad f(x),x-y}=f(y)-f(x)-D_{f}(y|x)$ and bounds
  $\inner{\grad\phi_{t}(\wtilde_{\tpp}),\cmp_{t}-\wtilde_{\tpp}}\le \phi_{t}(\cmp_{t})-\phi_{t}(\wtilde_{\tpp})$
  by convexity of $\phi_{t}$.
  Plugging this back into the previous display, we have
  \begin{align*}
    R_{T}(\vec{\cmp})
    &\le
      \sumtT\inner{\grad\psi_{t}(\wt)-\grad\psi_{t}(\wtilde_{\tpp}),\wtilde_{\tpp}-\cmp_{t}}\\
    &\qquad
      +\sumtT\sbrac{\Delta_{t}(\cmp_{t})+\phi_{t}(\cmp_{t})}
      +\sumtT -D_{\Delta_{t}}(\cmp_{t}|\wtilde_{\tpp})\\
    &\qquad
      +\sumtT\inner{g_{t},\wt-\wtilde_{\tpp}}-\Delta_{t}(\wtilde_{\tpp})-\phi_{t}(\wtilde_{\tpp}),
    \intertext{and using the three-point relation for Bregman divergences
    $\inner{\grad f(y)-\grad f(x),x-z}=D_{f}(z|y)-D_{f}(z|x)-D_{f}(x|y)$:}
    &=
        \sumtT D_{\psi_{t}}(\cmp_{t}|\wt)-D_{\psi_{t}}(\cmp_{t}|\wtilde_{\tpp})
      +\sumtT\sbrac{\Delta_{t}(\cmp_{t})+\phi_{t}(\cmp_{t})}
        +\sumtT -D_{\Delta_{t}}(\cmp_{t}|\wtilde_{\tpp})\\
    &\qquad
      +\sumtT\underbrace{\inner{g_{t},\wt-\wtilde_{\tpp}}-D_{\psi_{t}}(\wtilde_{\tpp}|\wt)
          -\Delta_{t}(\wtilde_{\tpp})-\phi_{t}(\wtilde_{\tpp})}_{=:\delta_{t}}\\
    &\overset{(a)}{=}
        \sumtT D_{\psi_{t}}(\cmp_{t}|\wt)-D_{\psi_{\tpp}}(\cmp_{t}|\wtilde_{\tpp})
        +\sumtT\sbrac{\Delta_{t}(\cmp_{t})+\phi_{t}(\cmp_{t})}+\delta_{1:T}\\
    &=
        \sumtT D_{\psi_{t}}(\cmp_{t}|\wt)-D_{\psi_{\tpp}}(\cmp_{t}|\wtpp)
        +\sumtT \underbrace{D_{\psi_{\tpp}}(\cmp_{t}|\wtpp)-D_{\psi_{\tpp}}(\cmp_{t}|\wtilde_{\tpp})}_{=:\xi_{t}}
        +\sumtT\sbrac{\Delta_{t}(\cmp_{t})+\phi_{t}(\cmp_{t})}+\delta_{1:T}\\
    &=
      D_{\psi_{1}}(\cmp_{1}|\w_{1})-D_{\psi_{T+1}}(\cmp_{T}|\w_{T+1})+
      \underbrace{\sum_{t=2}^{T}\sbrac{D_{\psi_{t}}(\cmp_{t}|\wt)-D_{\psi_{t}}(\cmp_{\tmm}|\wt)}+\sumtT\Delta_{t}(\cmp_{t})}_{=:(\star)}\\
      &\qquad
        +\sumtT\phi_{t}(\cmp_{t})+\xi_{1:T}+\delta_{1:T},
  \end{align*}
  where $(a)$ observes $D_{\Delta_{t}}(\cmp_{t}|\wtilde_{\tpp})=D_{\psi_{\tpp}}(\cmp_{t}|\wtilde_{\tpp})-D_{\psi_{t}}(\cmp_{t}|\wtilde_{\tpp})$.
  Observe that the term $(\star)$ simplifies as
  \begin{align*}
    (\star)
    &= \sum_{t=1}^{T}\Delta_{t}(\cmp_{t})+\sum_{t=2}^{T} D_{\psi_{t}}(\cmp_{t}|\wt)-D_{\psi_{t}}(\cmp_{\tmm}|\wt)\\
    &=
      \sum_{t=1}^{T}D_{\psi_{\tpp}}(\cmp_{t}|\w_{1})-D_{\psi_{t}}(\cmp_{t}|\w_{1})
      +\sum_{t=2}^{T} D_{\psi_{t}}(\cmp_{t}|\wt)-D_{\psi_{t}}(\cmp_{\tmm}|\wt)\\
    &=
      D_{\psi_{T+1}}(\cmp_{T}|\w_{1})-D_{\psi_{1}}(\cmp_{1}|\w_{1})\\
    &\qquad
      +\sum_{t=1}^{T-1}D_{\psi_{\tpp}}(\cmp_{t}|\w_{1})-D_{\psi_{\tpp}}(\cmp_{\tpp}|\w_{1})\\
    &\qquad
      +\sum_{t=2}^{T} D_{\psi_{t}}(\cmp_{t}|\wt)-D_{\psi_{t}}(\cmp_{\tmm}|\wt)\\
    &=
      D_{\psi_{T+1}}(\cmp_{T}|\w_{1})-D_{\psi_{1}}(\cmp_{1}|\w_{1})\\
    &\qquad
      +\sum_{t=2}^{T}D_{\psi_{t}}(\cmp_{\tmm}|\w_{1})-D_{\psi_{t}}(\cmp_{t}|\w_{1})\\
    &\qquad
      +\sum_{t=2}^{T} D_{\psi_{t}}(\cmp_{t}|\wt)-D_{\psi_{t}}(\cmp_{\tmm}|\wt)\\
    &=
      D_{\psi_{T+1}}(\cmp_{T}|\w_{1})-D_{\psi_{1}}(\cmp_{1}|\w_{1})\\
    &\qquad
      +\sum_{t=2}^{T}\psi_{t}(\cmp_{\tmm})-\psi_{t}(\cmp_{t})-\inner{\grad\psi_{t}(\w_{1}),\cmp_{\tmm}-\cmp_{t}}\\
    &\qquad
      +\sum_{t=2}^{T}\psi_{t}(\cmp_{t})-\psi_{t}(\cmp_{\tmm})-\inner{\grad\psi_{t}(\w_{t}),\cmp_{t}-\cmp_{\tmm}}\\
    &=
      D_{\psi_{T+1}}(\cmp_{T}|\w_{1})-D_{\psi_{1}}(\cmp_{1}|\w_{1})
      +\sum_{t=2}^{T}\underbrace{\inner{\grad\psi_{t}(\wt)-\grad\psi_{t}(\w_{1}),\cmp_{\tmm}-\cmp_{t}}}_{=:\cP_{t}}
  \end{align*}
  so plugging this back in above yields
  \begin{align*}
    R_{T}(\vec{\cmp})
    &\le
      D_{\psi_{1}}(\cmp_{1}|\w_{1})-D_{\psi_{T+1}}(\cmp_{T}|\w_{T+1})
      +(\star)
      +\sumtT\phi_{t}(\cmp_{t})
      +\xi_{1:T}+\delta_{1:T}\\
    &\le
      D_{\psi_{1}}(\cmp_{1}|\w_{1})-D_{\psi_{T+1}}(\cmp_{T}|\w_{T+1})
      +D_{\psi_{T+1}}(\cmp_{T}|\w_{1})-D_{\psi_{1}}(\cmp_{1}|\w_{1})+\cP_{2:T}\\
    &\qquad
      +\sumtT\phi_{t}(\cmp_{t})
      +\xi_{1:T}+\delta_{1:T}\\
    &=
      D_{\psi_{T+1}}(\cmp_{T}|\w_{1})-D_{\psi_{T+1}}(\cmp_{T}|w_{T+1})
      +\sumtT\phi_{t}(\cmp_{t})+\cP_{2:T}+\xi_{1:T}+\delta_{1:T}.
  \end{align*}
\end{proof}
In this paper, we will frequently use composite penalties $\phi_{t}$ which
are a linearization of some other function $\varphi_{t}$. The
next lemma shows how this changes the bound.
\newpage
\begin{restatable}{lemma}{CenteredMDLinear}\label{lemma:centered-md-linear}
  Under the same conditions as \Cref{lemma:centered-md-ii},
  let $\varphi_{t}:W\to \R_{+}$ be subdifferentiable convex functions
  and suppose we set $\phi_{t}(\w)=\inner{\grad\varphi_{t}(\wt),\w}$
  for some $\grad\varphi_{t}(\wt)\in \partial\varphi_{t}(\wt)$.
  Then \Cref{alg:centered-md-ii} guarantees
  \begin{align*}
    R_{T}(\vec{\cmp})&\le
    D_{\psi_{T+1}}(\cmp_{T}|\w_{1})-D_{\psi_{T+1}}(\cmp_{T}|\w_{T+1})
      +\sumtT\varphi_{t}(\cmp_{t})+\cP_{2:T}+\xi_{1:T}\\
    &\qquad
      +\sumtT\underbrace{\inner{\gtilde_{t},\wt-\wtilde_{\tpp}}-D_{\psi_{t}}(\wtilde_{\tpp}|\wt)-\Delta_{t}(\wtilde_{\tpp})-\varphi_{t}(\wt)}_{=:\delta_{t}},
  \end{align*}
  where $\gtilde_{t}=g_{t}+\grad\varphi_{t}(\wt)$.
\end{restatable}
\begin{proof}
  The proof is immediate by convexity of $\varphi_{t}$.
  From \Cref{lemma:centered-md-ii} we have
  \begin{align*}
    R_{T}(\vec{\cmp})
    &\le
      D_{\psi_{T+1}}(\cmp_{T}|\w_{1})-D_{\psi_{T+1}}(\cmp_{T}|w_{T+1})
      +\sumtT\phi_{t}(\cmp_{t})+\cP_{2:T}+\xi_{1:T}\\
    &\qquad
      +\sumtT\inner{g_{t},\wt-\wtilde_{\tpp}}-D_{\psi_{t}}(\wtilde_{\tpp}|\wt)-\Delta_{t}(\wtilde_{\tpp})-\phi_{t}(\wtilde_{\tpp}).
  \end{align*}
  Observe that we can write
  \begin{align*}
    \sumtT\phi_{t}(\cmp_{t})-\phi_{t}(\wtilde_{\tpp})
    &=
      \sumtT\inner{\grad\varphi_{t}(\wt),\cmp_{t}-\wtilde_{\tpp}}\\
    &=
      \sumtT\inner{\grad\varphi_{t}(\wt),\cmp_{t}-\wt}+\inner{\grad\varphi_{t}(\wt),\wt-\wtilde_{\tpp}}\\
    &\le
      \sumtT\varphi_{t}(\cmp_{t})-\varphi_{t}(\wt)+\inner{\grad\varphi_{t}(\wt),\wt-\wtilde_{\tpp}},
  \end{align*}
  so plugging this back in above gives the stated result:
  \begin{align*}
    R_{T}(\vec{\cmp})
    &\le
      D_{\psi_{T+1}}(\cmp_{T}|\w_{1})-D_{\psi_{T+1}}(\cmp_{T}|w_{T+1})
      +\sumtT\varphi_{t}(\cmp_{t})+\cP_{2:T}+\xi_{1:T}\\
    &\qquad
      +\sumtT\inner{\gtilde_{t},\wt-\wtilde_{\tpp}}-D_{\psi_{t}}(\wtilde_{\tpp}|\wt)-\Delta_{t}(\wtilde_{\tpp})-\varphi_{t}(w_{t}),
  \end{align*}
  where $\gtilde_{t}=g_{t}+\grad\varphi_{t}(\wt)$.
\end{proof}

\subsection{Multi-scale Experts Algorithm}
\label{app:stable-experts}

\begin{algorithm}
  \caption{Multi-scale Fixed-share}
  \label{alg:multi-scale-fixed-share}
  \begin{algorithmic}
    \STATE \KwInput{$p_{1}\in\Delta_{N}\cap(0,1]^{N}$, $\mu_{1},\ldots, \mu_{N}$ in $\R_{>0}$, $k>0$,
      weights $\beta_{1},\ldots,\beta_{T}$ in $[0,1]$}
    \STATE \KwInitialize{$q_{1}=p_{1}$}
    \STATE \textbf{Define} $\psi_{i}(x)=\frac{k}{\mu_{i}}\int_{0}^{x}\Log{v}dv$ for
    $i\in[N]$
    \For{$t=1:T$}{
      \STATE Play $p_{t}\in\Delta_{N}$, receive loss $\tilde\ell_{t}\in\R^{N}$
      \STATE Update
      $q_{\tpp}=\argmin_{q\in\Delta_{N}}\sum_{i=1}^{N}(\elltilde_{ti}+\mu_{i}\elltilde_{ti}^{2})q_{i} +  D_{\psi_{i}}(q_{i}|p_{ti})$
      \STATE Set $p_{\tpp}=(1-\beta_{t})q_{\tpp}+\beta_{t}p_{1}$
    }
  \end{algorithmic}
\end{algorithm}

For completeness, in this section we provide a multi-scale experts algorithm
which achieves the bound required for our dynamic regret algorithm in \Cref{sec:dynamic}. Our approach is
inspired by the Multi-scale Multiplicative-weight with Correction (MsMwC)
algorithm of \citet{chen2021impossible}, but formulated as a fixed-share
update instead of an update on a ``clipped'' simplex
$\tilde\Delta_{N}=\Delta_{N}\cap[\beta,1]^{N}$. The MsMwC algorithm provides a
guarantee analogous to the following theorem,
but formulating it as a fixed-share update
will allow us a bit more modularity when constructing our dynamic regret
algorithm in \Cref{app:dynamic}, which requires several
rather delicate conditions to come together in the right way.

\begin{restatable}{theorem}{MultiScaleFixedShareTemplate}\label{thm:multi-scale-fixed-share-template}
  Let
  $k\ge \frac{9}{2}$ and assume $\mu_{1},\ldots,\mu_{N}$ satisfy
  $\mu_{i}\elltilde_{ti}\le 1$ for all $t\in[T]$ and $i\in[N]$.
  Then for any $\cmp\in\Delta_{N}$, \Cref{alg:multi-scale-fixed-share}
  guarantees
  \begin{align*}
    \sumtT\inner{\elltilde_{t},p_{t}-\cmp}&\le
      \sum_{i=1}^{N}\cmp_{i}\sbrac{\frac{k\sbrac{\Log{\cmp_{i}/p_{1i}}+\sumtT\Log{\frac{1}{1-\beta_{t}}}}}{\mu_{i}}+\mu_{i}\sumtT\elltilde_{ti}^{2}}
      +k(1+\beta_{1:T})\sum_{i=1}^{N}\frac{p_{1i}}{\mu_{i}}.
  \end{align*}
  Moreover, for
  $\beta_{t}\le 1-\Exp{-\frac{1}{T}}$,
  \begin{align*}
    \sumtT\inner{\elltilde_{t},p_{t}-\cmp}
    &\le
      \sum_{i=1}^{N}\cmp_{i}\sbrac{\frac{k\sbrac{\Log{\cmp_{i}/p_{1i}}+1}}{\mu_{i}}+\mu_{i}\sumtT\elltilde_{ti}^{2}}+2k\sum_{i=1}^{N}\frac{p_{1i}}{\mu_{i}}
  \end{align*}
\end{restatable}
\begin{proof}
  The described algorithm is an instance of \Cref{alg:centered-md-ii}
  applied to the simplex $\Delta_{N}$ with
  $\varphi_{t}(p)=\sum_{i=1}^{N}\mu_{i}\elltilde_{ti}^{2}p_{i}$,
  and $\cM_{\tpp}(p)=(1-\beta_{t})p+\beta_{t}p_{1}$.
  Applying \Cref{lemma:centered-md-linear}:
  \begin{align*}
    \sumtT\inner{\elltilde_{t},p_{t}-\cmp}
    &\le
      D_{\psi}(\cmp|p_{1})-D_{\psi}(\cmp|p_{T+1})
      +\varphi_{1:T}(\cmp)
      +\xi_{1:T}+\delta_{1:T},
  \end{align*}
  where
  \begin{align*}
    \xi_{t}&=D_{\psi}(\cmp|p_{\tpp})-D_{\psi}(\cmp|q_{\tpp})\\
    \delta_{t}&=\inner{\elltilde_{t}+\grad\varphi_{t}(p_{t}),p_{t}-q_{\tpp}}-D_{\psi}(q_{\tpp}|p_{t})-\varphi_{t}(p_{t}).
  \end{align*}
  Observe that for any $\cmp$, $p$, and $q$ in $\Delta_{N}$ we can write
  \begin{align*}
    D_{\psi}(\cmp|p)-D_{\psi}(\cmp|q)
    &=
      \sum_{i=1}^{N}\frac{k}{\mu_{i}}\sbrac{\cmp_{i}\Log{\cmp_{i}/p_{i}}-\cmp_{i}+p_{i}}
      -\sum_{i=1}^{N}\frac{k}{\mu_{i}}\sbrac{\cmp_{i}\Log{\cmp_{i}/q_{i}}-\cmp_{i}+q_{i}}\\
    &=
      \sum_{i=1}^{N}\frac{k}{\mu_{i}}\sbrac{\cmp_{i}\Log{q_{i}/p_{i}}+p_{i}-q_{i}},
  \end{align*}
  so we have
  \begin{align*}
    D_{\psi}(\cmp|p_{1})-D_{\psi}(\cmp|p_{T+1})
    &=
      k\sum_{i=1}^{N}\frac{\cmp_{i}\Log{p_{T+1,i}/p_{1i}}+p_{1i}-p_{T+1,i}}{\mu_{i}}\\
    &\le
      k\sum_{i=1}^{N}\sup_{p\ge 0}\frac{\cmp_{i}\Log{p/p_{1i}}+p_{1i}-p}{\mu_{i}}\\
    &=
      k\sum_{i=1}^{N}\frac{\cmp_{i}\Log{\cmp_{i}/p_{1i}}+p_{1i}-\cmp_{i}}{\mu_{i}}\\
    &\le
      k\sum_{i=1}^{N}\frac{\cmp_{i}\Log{\cmp_{i}/p_{1i}}}{\mu_{i}}+k\sum_{i=1}^{N}\frac{p_{1i}}{\mu_{i}}
  \end{align*}
  and
  \begin{align*}
    \sumtT\xi_{t}
    &=
      \sumtT D_{\psi}(\cmp|p_{\tpp})-D_{\psi}(\cmp|q_{\tpp})\\
    &=
      k\sumtT\sum_{i=1}^{N}\frac{\cmp_{i}\Log{q_{\tpp,i}/p_{\tpp,i}}+p_{\tpp,i}-q_{\tpp,i}}{\mu_{i}}\\
    &=
      k\sumtT\sum_{i=1}^{N}\frac{\cmp_{i}\Log{\frac{q_{\tpp,i}}{(1-\beta_{t})q_{\tpp,i}+\beta_{t}q_{1,i}}}}{\mu_{i}}\\
    &\qquad
      +k\sumtT\sum_{i=1}^{N}\frac{(1-\beta_{t})q_{\tpp,i}+\beta_{t}q_{1,i}-q_{\tpp,i}}{\mu_{i}}\\
    &=
      k\sumtT\sum_{i=1}^{N}\frac{\cmp_{i}}{\mu_{i}}\Log{\frac{q_{\tpp,i}}{(1-\beta_{t})q_{\tpp,i}+\beta_{t}q_{1,i}}}\\
    &\qquad
      +k\sumtT\sum_{i=1}^{N}\frac{\beta_{t}(q_{1,i}-q_{\tpp,i})}{\mu_{i}}\\
    &\le
      k\sumtT\sum_{i=1}^{N}\frac{\cmp_{i}}{\mu_{i}}\Log{\frac{1}{1-\beta_{t}}}+\frac{\beta_{t}q_{1i}}{\mu_{i}}\\
    &=
      k\sum_{i=1}^{N}\frac{\cmp_{i}}{\mu_{i}}\sumtT\Log{\frac{1}{1-\beta_{t}}}+k\beta_{1:T}\sum_{i=1}^{N}\frac{p_{1i}}{\mu_{i}},
  \end{align*}
  where the last line recalls $p_{1}=q_{1}$.
  Plugging these bounds back into the above regret bound yields
  \begin{align}
    \sumtT\inner{\elltilde_{t},p_{t}-\cmp}\nonumber
    &\le
      k\sum_{i=1}^{N}\Bigg[\frac{\cmp_{i}\Log{\cmp_{i}/p_{1i}}}{\mu_{i}}+(1+\beta_{1:T})\frac{p_{1i}}{\mu_{i}}
      +\frac{\cmp_{i}}{\mu_{i}}\sumtT\Log{\frac{1}{1-\beta_{t}}}\Bigg]
      +\varphi_{1:T}(\cmp)+\delta_{1:T}\nonumber\\
    &=
      \sum_{i=1}^{N}\cmp_{i}\Bigg[\frac{k\sbrac{\Log{\cmp_{i}/p_{1i}}+\sumtT\Log{\frac{1}{1-\beta_{t}}}}}{\mu_{i}}
      +\mu_{i}\sumtT\elltilde_{ti}^{2}\Bigg]
      + k(1+\beta_{1:T})\sum_{i=1}^{N}\frac{p_{1i}}{\mu_{i}}\nonumber\\
    &\qquad
      +\sum_{i=1}^{N}\sumtT\underbrace{(\elltilde_{ti}+\mu_{i}\elltilde_{ti}^{2})(p_{ti}-q_{\tpp,i})
          -D_{\psi_{i}}(q_{\tpp,i}|p_{t,i})-\mu_{i}\elltilde_{ti}^{2}p_{ti}}_{=:\delta_{ti}},\label{eq:multi-scale-template-1}
  \end{align}
  where the last line recalls
  $\delta_{t}=\inner{\elltilde_{t}+\grad\varphi_{t}(p_{t}),p_{t}-q_{t}}-D_{\psi}(q_{\tpp}|p_{t})-\varphi_{t}(p_{t})$,
  $\varphi_{t}(p)=\sum_{i=1}^{N}\mu_{i}\elltilde_{ti}^{2}p_{i}$, and denotes
  $\psi_{i}(p)=\frac{k}{\mu_{i}}\int_{0}^{p}\Log{x}dx$ so that $\psi(p)=\sum_{i=1}^{N}\psi_{i}(p_{i})$.
  We next focus our attention on the terms in the last line, $\delta_{ti}$.

  Note that by construction, we have $p_{ti}=(1-\beta_{t})q_{ti}+\beta_{t}q_{1i}\ge \beta_{t}q_{1i}>0$
  for all $i$. Thus, $\psi_{i}(p)=\frac{k}{\mu_{i}}\int_{0}^{p}\Log{v}dv$ is twice differentiable everywhere on the
  line connecting $p_{ti}$ and $q_{\tpp,i}$ for any $i$ with $q_{\tpp,i}>0$.
  For any such $i$, we have via Taylor's theorem that there exists a $\tilde p_{i}$
  on the line connecting $p_{ti}$ and $q_{\tpp,i}$ such that
  \begin{align*}
    D_{\psi_{i}}(q_{\tpp,i}|p_{ti})
    &\ge
      \half(p_{ti}-q_{\tpp,i})^{2}\psi_{i}''(\tilde p_{i})
    =
      \half\frac{(p_{ti}-q_{\tpp,i})^{2}k}{\mu_{i}\tilde p_{i}}
  \end{align*}
  so using this with the assumption that $\mu_{i}\abs{\elltilde_{ti}}\le 1$, we have
  \begin{align*}
    \delta_{ti}
    &\le
      \abs{\elltilde_{ti}+\mu_{i}\elltilde_{ti}^{2}}\abs{p_{ti}-q_{\tpp,i}}-\half\frac{(p_{ti}-q_{\tpp,i})^{2}k}{\mu_{i}\tilde p_{i}}
      -\mu_{i}\elltilde_{ti}^{2}p_{ti}\\
    &\le
      2\abs{\elltilde_{ti}}\abs{p_{ti}-q_{\tpp,i}}-\half\frac{(p_{ti}-q_{\tpp,i})^{2}k}{\mu_{i}\tilde p_{i}}
      -\mu_{i}\elltilde_{ti}^{2}\tilde p_{i}+\mu_{i}\elltilde_{ti}^{2}\abs{p_{ti}-\tilde p_{i}}\\
    &\overset{(a)}{\le}
      3\abs{\elltilde_{ti}}\abs{p_{ti}-q_{\tpp,i}}-\half\frac{(p_{ti}-q_{\tpp,i})^{2}k}{\mu_{i}\tilde p_{i}}-\mu_{i}\elltilde_{ti}^{2}\tilde p_{i}\\
    &\le
      \frac{9}{2k}\mu_{i}\abs{\elltilde_{ti}}^{2}\tilde p_{i}-\mu_{i}\elltilde_{ti}^{2}\tilde p_{i}\\
    &\overset{(b)}{\le}
      0,
  \end{align*}
  where $(a)$ uses $\abs{\tilde p_{i} -p_{ti}}\le \abs{q_{\tpp,i}-p_{ti}}$ for any $\tilde p_{i}$ on the
  line connecting $q_{\tpp,i}$ and $p_{ti}$ and $(b)$ chooses $k\ge\frac{9}{2}$.
  Similarly, for any $i$ for which $q_{\tpp,i}=0$ we have
  \begin{align*}
    \delta_{ti}
    &=
      (\elltilde_{ti}+\mu_{i}\elltilde_{ti}^{2})p_{ti}-D_{\psi_{i}}(0|p_{ti})-\mu_{i}\elltilde_{ti}^{2}p_{ti}\\
    &\le
      \elltilde_{ti}p_{ti}-\frac{p_{ti}}{\mu_{i}}\\
    &\le
      \frac{p_{ti}}{\mu_{i}}-\frac{p_{ti}}{\mu_{i}}\le 0,
  \end{align*}
  where the last line again uses $\mu_{i}\abs{\elltilde_{ti}}\le1$. Thus, in either case we have
  $\delta_{ti}\le 0$. Plugging this into \Cref{eq:multi-scale-template-1}
  reveals the first statement of the theorem:
  \begin{align*}
    \sumtT\inner{\elltilde_{t},p_{t}-\cmp}
    &\le
    \sum_{i=1}^{N}\cmp_{i}\sbrac{\frac{k\sbrac{\Log{\cmp_{i}/p_{1i}}+\sumtT\Log{\frac{1}{1-\beta_{t}}}}}{\mu_{i}}+\mu_{i}\sumtT\elltilde_{ti}^{2}}
      + k(1+\beta_{1:T})\sum_{i=1}^{N}\frac{p_{1i}}{\mu_{i}}.
  \end{align*}

  For the second statement of the theorem,
  observe that
  $\beta_{t}\le 1-\Exp{-1/T}\le \frac{1}{T}$, so $\beta_{1:T}\le 1$,  and
  likewise $\Log{\frac{1}{1-\beta_{t}}}= \Log{\Exp{1/T}}=\frac{1}{T}$, so
  $\sumtT\Log{\frac{1}{1-\beta_{t}}}\le 1$. Hence, the previous display is
  bounded as
  \begin{align*}
    \sumtT\inner{\elltilde_{t},p_{t}-\cmp}
    &\le
      \sum_{i=1}^{N}\cmp_{i}\sbrac{\frac{k\sbrac{\Log{\cmp_{i}/p_{1i}}+1}}{\mu_{i}}+\mu_{i}\sumtT\elltilde_{ti}^{2}}+2k\sum_{i=1}^{N}\frac{p_{1i}}{\mu_{i}}
  \end{align*}

\end{proof}

\section{Proofs for Section~\ref{sec:quadratically-bounded} (\SectionQuadraticallyBounded)}
\label{app:quadratically-bounded}

\subsection{Proof of Theorem~\ref{thm:qb-adaptive-regret}}
\label{app:qb-adaptive-regret}
\begin{manualtheorem}{\ref{thm:qb-adaptive-regret}}
  Let $\cA$ be an online learning algorithm and let $\wt\in W$ be
  its output on round $t$.
  Let $\Set{g_{t}}$ be a $(G_{t},L_{t})$-quadratically bounded sequence
  \wrt{} $\Set{\wt}$, where
  $G_{t}\in[0,G_{\max}]$ and $L_{t}\in [0,L_{\max}]$
  for all $t$.
  Let $\epsilon>0$, $k\ge3$, $\kappa\ge 4$, $c\ge 4$, $V_{\tpp}=cG_{\max}^{2}+G_{1:t}^{2}$,
  $\rho_{\tpp}=\frac{1}{\sqrt{L_{\max}^{2}+L_{1:t}^{2}}}$,
  $\alpha_{\tpp}=\frac{\sqrt{V_{\tpp}}\log^{2}\brac{V_{\tpp}/G_{\max}^{2}}}{\epsilon G_{\max}}$,
  and set
  \begin{align*}
    \psi_{t}(\w)&=k\int_{0}^{\norm{\w}}\min_{\eta\le \frac{1}{G_{\max}}}\sbrac{\frac{\Log{x/\alpha_{t}+1}}{\eta}+\eta V_{t}}dx+\frac{\kappa \norm{\w}^{2}}{2\rho_{t}}\qquad\text{and}\qquad
    \varphi_{t}(\w)=\frac{L_{t}^{2}}{2\sqrt{L_{1:t}^{2}}}\norm{\w}^{2}.
  \end{align*}
  Then for any $\cmp\in W$, \Cref{alg:qb-adaptive-alg} guarantees
  \begin{align*}
    R_{T}(\cmp)
    &\le
      2\epsilon G_{\max}
      +\kappa\norm{\cmp}^{2}\sqrt{L_{\max}^{2}+L_{1:T}^{2}}
      +2k\norm{\cmp}\Max{\sqrt{V_{T+1}F_{T+1}(\norm{\cmp})}, G_{\max}F_{T+1}(\norm{\cmp})}
  \end{align*}
  where $F_{T+1}(\norm{\cmp})=\Log{\norm{\cmp}/\alpha_{T+1}+1}$.
\end{manualtheorem}
\begin{proof}
  We can assume without loss of generality that $\zeros\in W$, since we could
  otherwise just perform a coordinate
  translation. Hence, we have $\w_{1}=\argmin_{\w\in W}\psi_{1}(\w)=\zeros$, and it
  is easily seen that for any $\w\in W$ we'll have
  $D_{\psi_{t}}(\w|\w_{1})=D_{\psi_{t}}(\w|\zeros)=\psi_{t}(\w)$.

  First apply \Cref{lemma:centered-md-linear} with
  $\cM_{t}(\w)=\w$ and $\varphi_{t}(\w)=\frac{L_{t}^{2}}{2\sqrt{L_{1:t}^{2}}}\norm{\w}^{2}$
  to get
  \begin{align*}
    \sumtT\inner{g_{t},\wt-\cmp}
    &\le
      D_{\psi_{T+1}}(\cmp|\w_{1})+\varphi_{1:T}(\cmp)
      +\sumtT\underbrace{\inner{g_{t}+\grad\varphi_{t}(\wt),\wt-\wtpp}-D_{\psi_{t}}(\wtpp|\wt)
          -\Delta_{t}(\wtpp)-\varphi_{t}(\wt)}_{=:\delta_{t}}\\
    &\le \psi_{T+1}(\cmp)+\varphi_{1:T}(\cmp)+\delta_{1:T}.
  \end{align*}
  Let us first bound the leading term $\psi_{T+1}(\cmp)$.
  For brevity, denote $F_{t}(x)=\Log{x/\alpha_{t}+1}$ and let
  $\Psi_{t}(\norm{\w})=\int_{0}^{\norm{\w}}\min_{\eta\le 1/G_{\max}}\sbrac{\frac{F_{t}(x)}{\eta}+\eta V_{t}}dx$
  and $\Phi_{t}(\norm{\w})=\frac{\kappa}{2\rho_{t}}\norm{\w}^{2}$, so
  that $\psi_{t}(\w)=\Psi_{t}(\norm{\w})+\Phi_{t}(\norm{\w})$.
  Then
  \begin{align*}
    \psi_{T+1}(\cmp)
    &=
      k\int_{0}^{\norm{\cmp}}\Psi_{T+1}'(x)dx +\frac{\kappa}{2\rho_{t}}\norm{\cmp}^{2}\\
    &\le
      k\norm{\cmp}\Psi_{T+1}'(\norm{\cmp})+\frac{\kappa}{2}\norm{\cmp}^{2}\sqrt{L_{\max}^{2}+L_{1:T}^{2}}.
  \end{align*}
  Moreover,
  \begin{align*}
    \Psi_{t}'(\norm{\cmp})
    &=
      k\min_{\eta\le 1/G_{\max}}\sbrac{\frac{F_{t}(\norm{\cmp})}{\eta}+\eta V_{t}}\\
    &=
      \begin{cases}
        2k\sqrt{V_{t}F_{t}(\norm{\cmp})}&\text{if }G_{\max}\sqrt{F_{t}(\norm{\cmp})}\le\sqrt{V_{t}}\\
        kG_{\max}F_{t}(\norm{\cmp})+k\frac{V_{t}}{G_{\max}}&\text{otherwise}
      \end{cases}\\
    &\overset{(*)}{\le}
      \begin{cases}
        2k\sqrt{V_{t}F_{t}(\norm{\cmp})}&\text{if }G_{\max}\sqrt{F_{t}(\norm{\cmp})}\le\sqrt{V_{t}}\\
        2kG_{\max}F_{t}(\norm{\cmp})&\text{otherwise}
      \end{cases}\\
    &=
      2k\Max{\sqrt{V_{t}F_{t}(\norm{\cmp})},G_{\max}F_{t}(\norm{\cmp})}.
  \end{align*}
  where $(*)$ observes that $V_{t}/G_{\max}\le G_{\max}F_{t}(x)$ whenever
  $\Psi_{t}'(x)=kG_{\max}F_{t}(x)+kV_{t}/G_{\max}$.
  Next, using \Cref{lemma:sqrt-bounds} we have
  \begin{align*}
    \varphi_{1:T}(\cmp)
    &=
      \half\norm{\cmp}^{2}\sumtT\frac{L_{t}^{2}}{\sqrt{L_{1:t}^{2}}}\le \norm{\cmp}^{2}\sqrt{L_{1:T}^{2}},
  \end{align*}
  so overall we have
  \begin{align}
    \sumtT\inner{g_{t},\wt-\cmp}
    &\le
        2k\norm{\cmp}\Max{\sqrt{V_{T+1}F_{T+1}(\norm{\cmp})}, G_{\max}F_{T+1}(\norm{\cmp})}
      +\frac{\kappa}{2}\norm{\cmp}^{2}\sqrt{L_{\max}^{2}+L_{1:T}^{2}}
      +\norm{\cmp}^{2}\sqrt{L_{1:T}^{2}}+\delta_{1:T}\label{eq:adaptive-1}
  \end{align}
  We conclude by bounding the stability terms $\delta_{1:T}$.
  Recall that
  \begin{align*}
    \delta_{t}&=\inner{g_{t}+\grad\varphi_{t}(\wt),\wt-\wtpp}
          -D_{\psi_{t}}(\wtpp|\wt)-\Delta_{t}(\wtpp)-\varphi_{t}(\wt),
  \end{align*}
  where $\Delta_{t}(\w)=\psi_{\tpp}(\w)-\psi_{t}(\w)$.
  We first separate into terms related to the $G_{t}$'s and terms related to the
  $L_{t}$'s:
  \begin{align*}
    \delta_{t}
    &\le
      \brac{\norm{g_{t}}+\norm{\grad\varphi_{t}(\wt)}}\norm{\wt-\wtpp}\\
    &\quad
      -D_{\psi_{t}}(\wtpp|\wt)-\Delta_{t}(\wtpp)-\varphi_{t}(\wt)\\
    &\le
      G_{t}\norm{\wt-\wtpp}-D_{\Psi_{t}}(\wtpp|\wt)-\Delta_{t}(\wtpp)\\
    &\quad
      +2L_{t}\norm{\wt}\norm{\wt-\wtpp}-D_{\Phi_{t}}(\wtpp|\wt)-\varphi_{t}(\wt),
  \end{align*}
  where we slightly abuse notations $D_{\Psi_{t}}$ and $D_{\Phi_{t}}$ to denote the
  Bregman divergences \wrt{} the function $w\mapsto\Psi_{t}(\norm{\w})$ and $\w\mapsto\Phi_{t}(\norm{\w})$.
  In the second line, observe that
  $\Phi_{t}(\norm{\w})=\frac{\kappa}{2\rho_{t}}\norm{\w}^{2}$ is
  $\frac{\kappa}{\rho_{t}}$ strongly convex, so
  $D_{\Phi_{t}}(\wtpp|\wt)\ge \frac{\kappa}{2\rho_{t}}\norm{\wtpp-\wt}^{2}$
  and an application of Fenchel-Young inequality yields
  \begin{align*}
    2L_{t}\norm{\wt}\norm{\wt-\wtpp}-D_{\Phi_{t}}(\wtpp|\wt)-\varphi_{t}(\wt)
    &\le
      2L_{t}\norm{\wt}\norm{\wt-\wtpp}-\frac{\kappa}{2\rho_{t}}\norm{\wtpp-\wt}^{2}-\varphi_{t}(\wt)\\
    &\le
      \frac{4\rho_{t}L_{t}^{2}\norm{\wt}^{2}}{2\kappa}-\varphi_{t}(\wt)\\
    &=
      \frac{2 L_{t}^{2}\norm{\wt}^{2}}{\kappa\sqrt{L_{\max}+L_{1:\tmm}^{2}}}-\frac{L_{t}^{2}}{2\sqrt{L_{1:t}^{2}}}\norm{\wt}^{2}\\
    &\le
      \frac{2L_{t}^{2}\norm{\wt}^{2}}{\kappa\sqrt{L_{1:t}^{2}}}-\frac{L_{t}^{2}}{2\sqrt{L_{1:t}^{2}}}\norm{\wt}^{2}\\
    &\le0
  \end{align*}
  for $\kappa\ge 4$. Hence,
  \begin{align*}
    \delta_{t}\le G_{t}\norm{\wt-\wtpp}-D_{\Psi_{t}}(\wtpp|\wt)-\Delta_{t}(\wtpp),
  \end{align*}
  which we will bound by showing that $\Delta_{t}(\w)\ge \eta_{t}(\w)G_{t}^{2}$ for
  some suitable $G_{t}$-Lipschitz convex function $\eta_{t}$ and then invoking \Cref{lemma:lipschitz-stability}.
  To this end, observe that
  \begin{align*}
    \Delta_{t}(\w)&=\psi_{\tpp}(\w)-\psi_{t}(\w)\\
            &=
              \underbrace{\Psi_{\tpp}(\norm{\w})-\Psi_{t}(\norm{\w})}_{=:\Delta_{t}^{\Psi}(\w)}+\underbrace{\Phi_{\tpp}(\norm{\w})-\Phi_{t}(\norm{\w})}_{=:\Delta_{t}^{\Phi}(\w)}\\
            &\ge
              \Delta_{t}^{\Psi}(\w).
  \end{align*}
  Moreover, writing
  $\Delta_{t}^{\Psi}(\w)=\Psi_{\tpp}(\norm{\w})-\Psi_{t}(\norm{\w})=\int_{0}^{\norm{\w}}\Psi_{\tpp}'(x)-\Psi_{t}'(x)dx$,
  we have
  \begin{align*}
    \Psi_{\tpp}'(x)-\Psi_{t}'(x)
    &=
      k\min_{\eta\le 1/G_{\max}}\sbrac{\frac{F_{\tpp}(x)}{\eta}+\eta V_{\tpp}}
      -k\min_{\eta\le 1/G_{\max}}\sbrac{\frac{F_{t}(x)}{\eta}+\eta V_{t}}\\
    &\ge
      k\min_{\eta\le 1/G_{\max}}\sbrac{\frac{F_{t}(x)}{\eta}+\eta V_{\tpp}}
      -k\min_{\eta\le 1/G_{\max}}\sbrac{\frac{F_{t}(x)}{\eta}+\eta V_{t}}\\
    \intertext{and using the fact that for any $\eta\le 1/G_{\max}$, we can
    bound
    $\frac{F_{t}(x)}{\eta}+\eta V_{t}+\eta G_{t}^{2}\ge \min_{\eta^{*}\le 1/G}\sbrac{\frac{F_{t}(x)}{\eta^{*}}+\eta^{*} V_{t}}+\eta G_{t}^{2}$,
    we have}
    &\ge
      k\min_{\eta\le 1/G_{\max}}\sbrac{\frac{F_{t}(x)}{\eta}+\eta V_{t}}
      -k\min_{\eta\le 1/G_{\max}}\sbrac{\frac{F_{t}(x)}{\eta}+\eta V_{t}}
      +kG_{t}^{2}\Min{\sqrt{\frac{F_{t}(x)}{V_{\tpp}}},\frac{1}{G_{\max}}}\\
    &\ge
      kG_{t}^{2}\Min{\sqrt{\frac{F_{t}(x)}{2V_{t}}},\frac{1}{G_{\max}}}
    \ge
      G_{t}^{2}\Min{\sqrt{\frac{F_{t}(x)}{V_{t}}},\frac{1}{G_{\max}}},
  \end{align*}
  where the last line observes that
  $\frac{1}{V_{t}}=\frac{1}{V_{\tpp}}\frac{V_{\tpp}}{V_{t}}=\frac{1}{V_{\tpp}}\brac{1+\frac{G_{t}^{2}}{V_{t}}}\le \frac{2}{V_{\tpp}}$
  for $V_{t}\ge G_{t}^{2}$
  and recalls $k\ge 3$.
  Defining
  $\eta_{t}(\norm{\w})=\int_{0}^{\norm{\w}}\Min{\sqrt{\frac{F_{t}(x)}{V_{t}}},\frac{1}{G_{\max}}}dx$,
  we then immediately have:
  \begin{align*}
    \Delta_{t}^{\Psi}(\norm{\w})&\ge G_{t}^{2}\int_{0}^{\norm{\w}}\Min{\sqrt{\frac{F_{t}(x)}{V_{t}}},\frac{1}{G_{\max}}}dx
    =\eta_{t}(\norm{\w})G_{t}^{2}.
  \end{align*}
  Hence:
  \begin{align}
    \delta_{t}
    &\le
      G_{t}\norm{\wt-\wtpp}-D_{\psi_{t}}(\wtpp|\wt)-\Delta_{t}(\wtpp)\nonumber\\
    &\le
      G_{t}\norm{\wt-\wtpp}-D_{\psi_{t}}(\wt|\wtpp)-\eta_{t}(\norm{\wtpp})G_{t}^{2}\label{eq:adaptive-2}
  \end{align}
  Finally, we conclude by showing that $\psi_{t}$ satisfies the assumptions of
  \Cref{lemma:lipschitz-stability} \wrt{} this function $\eta_{t}$.

  We can write
  \begin{align*}
    \Psi_{t}(x)&=k\int_{0}^{x}\min_{\eta\le 1/G_{\max}}\sbrac{\frac{F_{t}(v)}{\eta}+\eta V_{t}}dv\\
           &=k\int_{0}^{x}\Max{2\sqrt{V_{t}F_{t}(v)},G_{\max}F_{t}(v)+\frac{V_{t}}{G_{\max}}}dv
  \end{align*}
  and so for any $x>0$ we have
  \begin{align*}
    \Psi_{t}'(x)
    &=
      \begin{cases}
        2k\sqrt{V_{t}F_{t}(x)}&\text{if }G_{\max}\sqrt{F_{t}(x)}\le\sqrt{V_{t}}\\
        kG_{\max}F_{t}(x)+\frac{kV_{t}}{G_{\max}}&\text{otherwise}
      \end{cases}\\
    \Psi_{t}''(x)
    &=
      \begin{cases}
        \frac{k\sqrt{V_{t}}}{(x+\alpha_{t})\sqrt{F_{t}(x)}}&\text{if }G_{\max}\sqrt{F_{t}(x)}\le\sqrt{V_{t}}\\
        \frac{kG_{\max}}{x+\alpha_{t}}&\text{otherwise}
      \end{cases}\\
    \Psi_{t}'''(x)
    &=
      \begin{cases}
        \frac{-k\sqrt{V_{t}}(1+2F_{t}(x))}{2(x+\alpha_{t})^{2}F_{t}(x)^{3/2}}&\text{if }G_{\max}\sqrt{F_{t}(x)}\le\sqrt{V_{t}}\\
        \frac{-kG_{\max}}{(x+\alpha_{t})^{2}}&\text{otherwise}
      \end{cases}.
  \end{align*}
  Clearly, we have $\Psi_{t}(x)\ge 0$, $\Psi_{t}'(x)\ge 0$,
  $\Psi_{t}''(x)\ge 0$, and $\Psi_{t}'''(x)\le 0$ for all $x> 0$.
  Moreover, for any $x\ge\alpha_{t}(e-1)=:\mathring{x}_{t}$, we have
  \begin{align*}
    \frac{\abs{\Psi_{t}'''(x)}}{\Psi_{t}''(x)^{2}}
    &=
      \begin{cases}
        \frac{k\sqrt{V_{t}}(1+2F_{t}(x))}{2(x+\alpha_{t})^{2}F_{t}(x)^{3/2}}\frac{(x+\alpha_{t})^{2}F_{t}(x)}{k^{2}V_{t}}&\text{if
        }G_{\max}\sqrt{F_{t}(x)}\le\sqrt{V_{t}}\\
        \frac{kG_{\max}}{(x+\alpha_{t})^{2}}\frac{(x+\alpha_{t})^{2}}{k^{2}G_{\max}^{2}}&\text{otherwise}
      \end{cases}\\
    &=
      \begin{cases}
        \frac{1}{2k\sqrt{V_{t}}}\brac{\frac{1}{\sqrt{F_{t}(x)}}+2\sqrt{F_{t}(x)}}&\text{if
        }G_{\max}\sqrt{F_{t}(x)}\le\sqrt{V_{t}}\\
        \frac{1}{k G_{\max}}&\text{otherwise}
      \end{cases}\\
    \intertext{and since $x> \alpha_{t}(e-1)$, we have $F_{t}(x)>1$ and hence
    $\frac{1}{\sqrt{F_{t}(x)}}\le\sqrt{F_{t}(x)}$:}
    &\le
      \begin{cases}
        \frac{3\sqrt{F_{t}(x)}}{2k\sqrt{V_{t}}}&\text{if
        }G_{\max}\sqrt{F_{t}(x)}\le\sqrt{V_{t}}\\
        \frac{1}{kG_{\max}}&\text{otherwise}
        \end{cases}\\
    &\le
      \half\Min{\sqrt{\frac{F_{t}(x)}{V_{t}}},\frac{1}{G_{\max}}}=\half\eta_{t}'(x),
  \end{align*}
  where the last line recalls
  $\eta_{t}(x)=\int_{0}^{x}\Min{\sqrt{\frac{F_{t}(v)}{V_{t}}},\frac{1}{G_{\max}}}dv$
  and chooses $k\ge 3$. Further, observe that $\eta_{t}(x)$ is convex and
  $\eta_{t}'(x)\le \frac{1}{G_{\max}}$, hence $\frac{1}{G_{\max}}$-Lipschitz. Thus, $\Psi_{t}$ satisfies the conditions
  of \Cref{lemma:lipschitz-stability} with
  $\eta_{t}(x)=\int_{0}^{x}\Min{\sqrt{\frac{F_{t}(x)}{V_{t}}},\frac{1}{G_{\max}}}dx$
  and $\mathring{x}_{t}=\alpha_{t}(e-1)$,
  so summing \Cref{eq:adaptive-2} over all $t$, we have
  \begin{align*}
    \sumtT\delta_{t}
    &\le
      \sumtT G_{t}\norm{\wt-\wtpp}-D_{\Psi_{t}}(\wtpp|\wt)-\eta_{t}(\norm{\wtpp})G_{t}^{2}\\
    &\le
      \sumtT\frac{2 G_{t}^{2}}{\Psi_{t}''(\mathring{x}_{t})}
    =
      \sumtT\frac{2 G_{t}^{2}}{k\sqrt{V_{t}}}(\norm{\mathring{x}_{t}}+\alpha_{t})\\
    &\le
      \sumtT\frac{2e\alpha_{t} G_{t}^{2}}{k\sqrt{V_{t}}}
    \le
      \sumtT2\frac{\alpha_{t} G_{t}^{2}}{\sqrt{V_{t}}}
    \end{align*}
    where the last line bounds $e/k\le 3/k\le 1$ for $k\ge 3$.
    Next, substitute
    $\alpha_{t}=\frac{\epsilon G_{\max}}{\sqrt{V_{t}}\log^{2}(V_{t}/G_{\max})}$ to bound
    \begin{align*}
      \sumtT\delta_{t}
      &\le
      2\epsilon G_{\max}\sumtT\frac{G_{t}^{2}}{V_{t}\log^{2}\brac{V_{t}/G_{\max}^{2}}}\\
    &\le
      2\epsilon G_{\max}\sumtT\frac{G_{t}^{2}}{((c-1)G_{\max}^{2}+G_{1:t}^{2})\log^{2}\brac{\frac{(c-1)G_{\max}^{2}+G_{1:t}^{2}}{G_{\max}^{2}}}}\\
    &\le
      2\epsilon G_{\max}\int_{(c-1)G_{\max}^{2}}^{(c-1)G_{\max}^{2}+G_{1:T}^{2}}\frac{1}{x\log^{2}(x/G_{\max}^{2})}dx\\
    &=
      2\epsilon G_{\max}\frac{1}{\log(x/G_{\max}^{2})}\Bigg|_{(c-1)G_{\max}^{2}}^{(c-1)G_{\max}^{2}+G_{1:T}^{2}}\\
    &\le
      \frac{2\epsilon G_{\max}}{\Log{c-1}}\le 2\epsilon G_{\max},
  \end{align*}
  for $c\ge 4$.
  Finally, plugging this back into \Cref{eq:adaptive-1} yields
  \begin{align*}
    \sumtT\inner{g_{t},\wt-\cmp}
    &\le
      2k\norm{\cmp}\Max{\sqrt{V_{T+1}F_{T+1}(\norm{\cmp})}, G_{\max}F_{T+1}(\norm{\cmp})}\\
    &\qquad
      +\frac{\kappa}{2}\norm{\cmp}^{2}\sqrt{L_{\max}^{2}+L_{1:T}^{2}}
      +\norm{\cmp}^{2}\sqrt{L_{1:T}^{2}}+\delta_{1:T}\\
    &\le
      2k\norm{\cmp}\Max{\sqrt{V_{T+1}F_{T+1}(\norm{\cmp})}, G_{\max}F_{T+1}(\norm{\cmp})}\\
    &\qquad
      +\frac{\kappa}{2}\norm{\cmp}^{2}\sqrt{L_{\max}^{2}+L_{1:T}^{2}}
      +\norm{\cmp}^{2}\sqrt{L_{1:T}^{2}}+2\epsilon G_{\max}\\
    &\le
      2\epsilon G_{\max}+\kappa\norm{\cmp}^{2}\sqrt{L_{\max}^{2}+L_{1:T}^{2}}\\
    &\qquad
      2k\norm{\cmp}\Max{\sqrt{V_{T+1}F_{T+1}(\norm{\cmp})}, G_{\max}F_{T+1}(\norm{\cmp})}\\
  \end{align*}

\end{proof}

\subsection{Proof of Theorem~\ref{thm:qb-olo-static-lb}}
\label{app:qb-olo-static-lb}
\QBOLOStaticLB*
\begin{proof}
  Let $\wt\in \R^{2}$ be the output of algorithm $\cA$ at time $t$.
  Consider sequences $g_{1},\ldots,g_{T}$ where
  $g_{t}\in\Set{\begin{pmatrix}-G\\L\norm{\wt}\end{pmatrix},\begin{pmatrix}-G\\-L\norm{\wt}\end{pmatrix}}$,
  and define the randomized sequence $\gtilde_{t}=\begin{pmatrix}-G\\-\eps_{t}L\norm{\wt}\end{pmatrix}$
  where $\eps_{t}$ are independent random signs.
  Consider the worst-case regret
  against a comparator constrained to an $\ell_{\infty}$ ball of
  radius $U$:
  \begin{align*}
    \sup_{g_{1},\ldots,g_{T}}R_{T}
    &=
      \sup_{g_{1},\ldots,g_{T}}\sumtT\inner{g_{t},\wt}-\min_{u:\norm{\cmp}_{\infty}\le U}\sumtT\inner{g_{t},\cmp}\\
    &\ge
      \E_{\eps_{1},\ldots,\eps_{T}}\sbrac{\sumtT\inner{\gtilde_{t},\wt}-\min_{\cmp:\norm{\cmp}_{\infty}\le U}\sumtT\inner{\gtilde_{t},\cmp}}\\
    &\ge
      \E_{\eps_{1},\ldots,\eps_{T}}\sbrac{-\sumtT G\norm{\wt}-\min_{\cmp:\norm{\cmp}_{\infty}\le U}\sumtT-G\cmp_{1}-\cmp_{2}\eps_{t}L\norm{\wt}}\\
    &=
      \E_{\eps_{1},\ldots,\eps_{T}}\sbrac{-G\sumtT\norm{\wt}+GTU+\max_{\abs{u_{2}}\le U}u_{2}L\sumtT\eps_{t}\norm{\wt}}\\
    &=
      \E_{\eps_{1},\ldots,\eps_{T}}\sbrac{GTU+UL\abs{\sumtT\eps_{t}\norm{\wt}}-G\sumtT\norm{\wt}}\\
    &\overset{(a)}{\ge}
      \E_{\eps_{1},\ldots,\eps_{T}}\sbrac{GTU+\frac{UL}{\sqrt{2}}\sqrt{\sumtT\norm{\wt}^{2}}-G\sumtT\norm{\wt}}\\
    &\overset{(b)}{\ge}
      \E_{\eps_{1},\ldots,\eps_{T}}\sbrac{GTU+\frac{UL}{\sqrt{2}}\sqrt{\sumtT\norm{\wt}^{2}}-G\sqrt{T\sumtT\norm{\wt}^{2}}}\\
    \intertext{where $(a)$ applies Khintchine inequality, $(b)$ applies
    Cauchy-Schwarz inequality, and choosing $U=\frac{G}{L}\sqrt{2T}$ we have}
    &=
      GTU
      =
      \frac{L}{\sqrt{2}}U^{2}\sqrt{T}=\frac{L\norm{\cmp}^{2}\sqrt{T}}{2\sqrt{2}},
  \end{align*}
  where the final equality bounds $\norm{\cmp}^{2}=\cmp_{1}^{2}+\cmp_{2}^{2}\le 2U^{2}$.
  Hence, there exists a sequence of $g_{t}$  which incurs at least
  $\Omega(L\norm{\cmp}^{2}\sqrt{T})$ regret. Moreover,
  for any algorithm which guarantees $R_{T}(\zeros)\le \epsilon$,
  there exists
  a sequence $g_{1},\ldots,g_{T}$ with $\norm{g_{t}}\le G$ for all $t$ such that for any $T$ and $\cmp$,
  $R_{T}(\cmp)\ge \frac{G}{3\sqrt{2}}\norm{\cmp}\sqrt{T\Log{\norm{\cmp}\sqrt{T}/\sqrt{2}\epsilon}}$
  \citep[Theorem
  8]{mcmahan2012noregret}.
  Thus, taking the worst of these two sequences
  yields
  \begin{align*}
    \sup_{g_{1},\ldots,g_{T}}R_{T}\ge \Max{\frac{G}{3\sqrt{2}}\norm{\cmp}\sqrt{T\Log{\norm{\cmp}\sqrt{T}/\sqrt{2}\epsilon}}, \frac{L\norm{\cmp}^{2}\sqrt{T}}{2\sqrt{2}}}
  \end{align*}

\end{proof}

\section{Proofs for Section~\ref{sec:dynamic} (\SectionDynamic)}
\label{app:dynamic}

The main objective of this section is to prove
\Cref{thm:dynamic,thm:smooth-dynamic}.
At a high level, the strategy is simple: we run several
instances of projected gradient descent, each with
a different restricted domain $W_{D}=\Set{\w\in W:\norm{\w}\le D}$
and stepsize $\eta$, and then use a particular
experts algorithm to combine them. We first assemble a collection
of core lemmas that provide the regret of the
base algorithm (\Cref{lemma:ogd-dynamic}),
the regret of \Cref{alg:dynamic-meta}
in terms of the regret of any of the base algorithms
(\Cref{lemma:dynamic-untuned}), as well as
some utility lemmas
(\Cref{lemma:mu-bounds,lemma:omega-bound,lemma:large-M,lemma:small-M})
to help tame some unwieldy algebraic expressions and
case work.
We then prove the main results \Cref{thm:dynamic,thm:smooth-dynamic}
in \Cref{app:dynamic-thm,app:smooth-dynamic} respectively.
Finally, we prove our lowerbound \Cref{thm:dynamic-lb}
in \Cref{app:dynamic-lb}.

The base algorithms that we combine are instances of
(projected) online gradient descent
with an additional bias term 
added to the update.
The following lemma provides the regret template
for this algorithm.
\begin{minipage}{\columnwidth}
\begin{restatable}{lemma}{OGDDynamic}\label{lemma:ogd-dynamic}
  For all $t$ let $\ell_{t}: W\to \R$ be convex.
  Let
  $K\ge 1$,   $L_{t}\ge 0$, and $K\eta L_{t}\le 1$ for
  all $t$.
  Let $W_{D}=\Set{\w\in W:\norm{\w}\le D}$, $\w_{1}=\zeros$,
  and on each round update
  $\wtpp=\proj_{\w\in W_{D}}\brac{\wt - \eta(1+K \eta L_{t})g_{t}}$,
  where $g_{t}\in\partial\ell_{t}(\wt)$.
  Then for any $\vec{\cmp}=(\cmp_{1},\ldots,\cmp_{T})$ in $W_{D}$,
  \begin{align*}
    R_{T}(\vec{\cmp})&\le
                      \frac{\norm{\cmp_{T}}^{2}+2DP_{T}}{2\eta} +K\eta \sumtT L_{t}\sbrac{\ell_{t}(\cmp_{t})-\ell_{t}(\wt)}+2\eta\sumtT\norm{g_{t}}^{2}
  \end{align*}
  where
  $P_{T}=\sum_{t=2}^{T}\norm{\cmp_{t}-\cmp_{\tmm}}$.
\end{restatable}
\end{minipage}

\begin{proof}
  The result follows easily using existing analyses. For instance,
  the update can be seen as an instance of \Cref{alg:centered-md-ii}
  with $\psi_{t}(\w)=\frac{1}{2\eta}\norm{\w}^{2}$,
  $\phi_{t}(\w)=K\eta L_{t}\inner{g_{t},\w}$ for $g_{t}\in\partial\ell_{t}(\wt)$,
  domain $W_{D}=\Set{\w\in W:\norm{\w}\le D}$, and $\cM_{t}(\w)=w$ for all $t$. Letting $\w_{1}=\zeros$
  and applying \Cref{lemma:centered-md-linear}, we have:
  \begin{align*}
    R_{T}(\vec{\cmp})\nonumber
    &\le
      \psi_{T+1}(\cmp_{T})+\sum_{t=2}^{T}\inner{\grad\psi_{t}(\w_{t}),\cmp_{\tmm}-\cmp_{t}}
    +K\eta\sumtT L_{t}\ell_{t}(\cmp_{t})\nonumber\\
    &\qquad+\sumtT\inner{g_{t}+K\eta L_{t}g_{t},\wt-\wtpp}-D_{\psi_{t}}(\wtpp|\wt)-K\eta L_{t}\ell_{t}(\wt)\nonumber\\
    &\le
      \frac{\norm{\cmp_{T}}^{2}}{2\eta}+\sum_{t=2}\frac{D}{\eta}\norm{\cmp_{t}-\cmp_{\tmm}}
      +K\eta \sumtT L_{t}\sbrac{\ell_{t}(\cmp_{t})-\ell_{t}(\wt)}\nonumber\\
    &\qquad
      +\sumtT(1+K\eta L_{t})\inner{g_{t},\wt-\wtpp}-D_{\psi_{t}}(\wtpp|\wt)\nonumber\\
    &\overset{(a)}{\le}
      \frac{\norm{\cmp_{T}}^{2}+2DP_{T}}{2\eta}
      +K\eta \sumtT L_{t}\sbrac{\ell_{t}(\cmp_{t})-\ell_{t}(\wt)}\nonumber\\
    &\qquad
      +\sumtT(1+K\eta L_{t})\inner{g_{t},\wt-\wtpp}-\frac{\norm{\wtpp-\wt}^{2}}{2\eta}\nonumber\\
    &\overset{(b)}{\le}
      \frac{\norm{\cmp_{T}}^{2}+2DP_{T}}{2\eta}
      +K\eta \sumtT L_{t}\sbrac{\ell_{t}(\cmp_{t})-\ell_{t}(\wt)}
      +\frac{\eta}{2}\sumtT(1+K\eta L_{t})^{2}\norm{g_{t}}^{2}\nonumber\\
    &\overset{(c)}{\le}
      \frac{\norm{\cmp_{T}}^{2}+2DP_{T}}{2\eta}
      +K\eta \sumtT L_{t}\sbrac{\ell_{t}(\cmp_{t})-\ell_{t}(\wt)}
      +2\eta\sumtT\norm{g_{t}}^{2}
  \end{align*}
  the $(a)$ observes that
  $D_{\psi_{t}}(\wtpp|\wt)\ge \frac{\norm{\wtpp-\wt}^{2}}{2\eta}$ by
  $\frac{1}{\eta}$-strong convexity of $\psi$, $(b)$ is
  Fenchel-Young inequality, and $(c)$ uses $K\eta L_{t}\le 1$.

\end{proof}

The following lemma provides a generic regret bound
for \Cref{alg:dynamic-meta}. The take-away is that
the regret will scale with the regret of any of
the experts up to two extra terms $C_{\cS}$ and $\Lambda_{T}(\eta,D)$,
which we will later ensure are small.
\begin{minipage}{\columnwidth}
\begin{restatable}{lemma}{DynamicUntuned}\label{lemma:dynamic-untuned}
  For any $\tau=(\eta,D)\in\cS$ with $\eta\le\frac{1}{KL_{\max}}$ and sequence
  $\vec{\cmp}=(\cmp_{1},\ldots,\cmp_{T})$ in $W$ satisfying $\norm{\cmp_{t}}\le D$
  for all $t$,
  \Cref{alg:dynamic-meta} guarantees
  \begin{align*}
    R_{T}(\vec{\cmp})
    &\le
      2k C_{\cS}+2kDG_{\max}\Lambda_{T}(\tau)
      +\frac{\norm{\cmp_{T}}^{2}+2DP_{T}+4kD^{2}\Lambda_{T}(\tau)}{2\eta}\\
    &\quad
      +K\eta\sumtT L_{t}\sbrac{\ell_{t}(\cmp_{t})-\ell_{t}(\wt^{(\tau)})} +4\eta\sumtT\norm{g_{t}^{(\tau)}}^{2}\\
  \end{align*}
  where $k\ge 9/2$ and
  \begin{align*}
    C_{\cS}&\defeq\frac{\sum_{\tilde\tau\in\cS}\mu_{\tilde\tau}}{\sum_{\tilde\tau\in\cS}\mu_{\tilde\tau}^{2}},\qquad
    \Lambda_{T}(\tau)\defeq \Log{\frac{\sum_{\tilde\tau\in\cS}\mu_{\tilde \tau}^{2}}{\mu_{\tau}^{2}}}+1.
  \end{align*}
\end{restatable}
\end{minipage}
\begin{proof}
  Let $\tau=(\eta, D)\in\cS$ and let $\cA_{\tau}$ denote an algorithm
  playing
  $\wtpp^{(\tau)}=\proj_{\w\in W:\norm{\w}\le D}\brac{\wt^{(\tau)}-\eta(1+K\eta L_{t}) g_{t}^{(\tau)}}$
  for $g_{t}^{(\tau)}\in\partial\ell_{t}(\wt^{(\tau)})$.
  \Cref{alg:dynamic-meta} is constructed as a collection of
  algorithms $\cA_{\tau}$, with an multi-scale experts algorithm
  (\Cref{alg:multi-scale-fixed-share}) to combine their predictions.
  First, observe that the regret decomposes into
  the regret of any expert $\cA_{\tau}$ plus the regret of
  the experts algorithm relative to expert $\cA_{\tau}$:
  \begin{align}
    R_{T}(\vec{\cmp})
    &=
      \sumtT\ell_{t}(\wt)-\ell_{t}(\cmp_{t})\nonumber\\
    &=
      \underbrace{\sumtT\ell_{t}\brac{\wt^{(\tau)}}-\ell_{t}(\cmp_{t})}_{=:R_{T}^{\cA_{\tau}}(\vec{\cmp})} + \sumtT\ell_{t}(\wt)-\ell_{t}\brac{\wt^{(\tau)}}\nonumber\\
    &=
      R_{T}^{\cA_{\tau}}(\vec{\cmp})+\sumtT\ell_{t}\brac{\sum_{\tilde \tau\in\cS}p_{t}(\tilde \tau)\ell_{t}(\wt^{(\tilde \tau)})}-\ell_{t}\brac{\wt^{(\tau)}}\nonumber\\
    \intertext{and by convexity of $\ell_{t}$ and Jensen's inequality:}
    &\le
      R_{T}^{\cA_{\tau}}(\vec{\cmp})+\sumtT\sbrac{\sum_{\tilde\tau\in\cS}p_{t}(\tilde\tau)\ell_{t}\brac{\wt^{(\tilde \tau)}}}-\ell_{t}\brac{\wt^{(\tau)}}\nonumber\\
    &=
      R_{T}^{\cA_{\tau}}(\vec{\cmp})+\sumtT\sum_{\tilde\tau\in\cS}\ell_{t}\brac{\wt^{(\tilde \tau)}}\sbrac{p_{t}(\tilde \tau)-\boldsymbol{1}\Set{\tau=\tilde\tau}}\nonumber\\
    &\overset{(a)}{=}
      R_{T}^{\cA_{\tau}}(\vec{\cmp})\nonumber\\
    &\qquad
      +\sumtT\sum_{\tilde\tau\in\cS}\sbrac{\ell_{t}\brac{\wt^{(\tilde \tau)}}-\ell_{t}(\wtilde_{t})}\sbrac{p_{t}(\tilde \tau)-p^{*}_{\tau}(\tilde \tau)}\nonumber\\
    &\qquad
      +\sumtT\sum_{\tilde\tau\in\cS}\ell_{t}(\wtilde_{t})(p_{t}(\tilde\tau)-p^{*}_{\tau}(\tilde\tau))\nonumber\\
    &\overset{(b)}{=}
      R_{T}^{\cA_{\tau}}(\vec{\cmp})
      +\sumtT\sum_{\tilde\tau\in\cS}\sbrac{\ell_{t}\brac{\wt^{(\tilde \tau)}}-\ell_{t}(\wtilde_{t})}\sbrac{p_{t}(\tilde \tau)-p^{*}_{\tau}(\tilde \tau)}\nonumber\\
    &\overset{(c)}{=}
      R_{T}^{\cA_{\tau}}(\vec{\cmp})+\underbrace{\sumtT\inner{\elltilde_{t},p_{t}-p^{*}_{\tau}}\nonumber}_{=:R_{T}^{\text{Meta}}(p^{*}_{\tau})}\\
    &=
      R_{T}^{\cA_{\tau}}(\vec{\cmp})+R_{T}^{\text{Meta}}(p^{*}_{\tau})\label{eq:dynamic-untuned-1},
  \end{align}
  where $\wtilde_{t}$ is an arbitrary reference point with
  $\norm{\wtilde_{t}}\le D_{\min}$ (and hence is in the domain of all of the
  experts $\cA_{\tau}$), $(a)$ defines $p^{*}_{\tau}(\tilde \tau)=1$ if $\tilde \tau=\tau$ and $0$
  otherwise, $(b)$ observes that
  $\sum_{\tilde\tau\in\cS}\ell_{t}(\wtilde_{t})(p_{t}(\tilde\tau)-p^{*}_{\tau}(\tilde\tau))=\ell_{t}(\wtilde_{t})\sum_{\tilde\tau\in\cS}p_{t}(\tilde\tau)-p^{*}_{\tau}(\tilde\tau)=0$,
  and $(c)$ defines $\elltilde_{t}\in \R^{\abs{\cS}}$ with
  $\tilde \ell_{t,\tau}=\ell_{t}(\wt^{(\tau)})-\ell_{t}(\wtilde_{t})$.

  Now for any $\tau=(\eta,D)\in\cS$ with $D\ge \max_{t}\norm{\cmp_{t}}$, we have
  via \Cref{lemma:ogd-dynamic}
  that
  \begin{align}
    R_{T}^{\cA_{\tau}}(\cmp)
      &\le
      \frac{\norm{\cmp_{T}}^{2}+2DP_{T}}{2\eta}+K\eta\sumtT L_{t}\sbrac{\ell_{t}(\cmp_{t})-\ell_{t}(\wt^{(\tau)})}+2\eta\sumtT\norm{g_{t}^{(\tau)}}^{2}.
      \label{eq:dynamic-untuned-2}
  \end{align}
  To bound $R_{T}^{\text{Meta}}(p_{\tau}^{*})$, observe that for any
  $\tilde \tau=(\tilde \eta,\tilde D)$, we have
  \begin{align*}
    \elltilde_{t,\tilde \tau}&=\ell_{t}(\wt^{(\tilde\tau)})-\ell_{t}(\wtilde_{t})\le \norm{\grad\ell_{t}(\wt^{(\tilde\tau)})}\norm{\wt^{(\tilde\tau)}-\wtilde_{t}}\\
    &\le
      \brac{G_{\max}+L_{\max}\tilde D}2\tilde D,
  \end{align*}
  and so with
  $\mu_{\tilde \tau}=\frac{1}{2\tilde D\brac{G_{\max} + \tilde D/\tilde\eta}}$ and
  $\tilde\eta\le \frac{1}{KL_{\max}}\le \frac{1}{L_{\max}}$
  we have
  \begin{align*}
    \mu_{\tilde\tau}\ell_{t,\tilde \tau}
    &\le
      \frac{1}{2\tilde D\brac{G_{\max} + \tilde D/\tilde\eta}}2\tilde D\brac{G_{\max}+L_{\max}\tilde D}\\
    &\le
      \frac{1}{\brac{G_{\max} + L_{\max}\tilde D}}\brac{G_{\max}+L_{\max}\tilde D}\\
    &= 1,
  \end{align*}
  so these choices meet the
  assumptions of \Cref{thm:multi-scale-fixed-share-template} and
  we have:
  \begin{align*}
    R_{T}^{\text{Meta}}(p^{*}_{\tau})
    &\le
      \sum_{\tilde\tau\in\cS}p^{*}_{\tau}(\tilde\tau)\sbrac{\frac{k\sbrac{\Log{p^{*}_{\tau}(\tilde\tau)/p_{1\tilde\tau}}+1}}{\mu_{\tilde\tau}}+\mu_{\tilde \tau}\sumtT\elltilde_{t\tilde\tau}^{2}}
    +2k\sum_{\tilde\tau\in\cS}\frac{p_{1\tilde \tau}}{\mu_{\tilde\tau}}
  \end{align*}
  for $k\ge 9/2$.
  Recalling that $p_{\tau}^{*}(\tilde\tau)=1$ when $\tilde\tau=\tau$ and $0$
  otherwise and that $\tau=(D,\eta)$, the first sum is bound as
  \begin{align*}
      \frac{k\sbrac{\Log{p^{*}_{\tau}(\tau)/p_{1\tau}}+1}}{\mu_{\tau}}+\mu_{ \tau}\sumtT\elltilde_{t\tau}^{2}
    &=
      2k D\brac{G_{\max}+\frac{ D}{ \eta}}\sbrac{\Log{1/p_{1\tau}}+1}
      +\frac{ \eta}{ 2D\brac{G_{\max}\eta + D}}\sumtT\elltilde_{t,\tau}^{2}\\
    &\le
      2k D\brac{G_{\max}+\frac{ D}{ \eta}}\sbrac{\Log{1/p_{1\tau}}+1}
      +\frac{ \eta}{2 D^{2}}\sumtT\norm{\grad\ell_{t}(\wt^{(\tau)})}^{2} 4D^{2}\\
    &=
      2k D\brac{G_{\max}+\frac{ D}{ \eta}}\sbrac{\Log{1/p_{1\tau}}+1}
      +2\eta\sumtT\norm{g_{t}^{(\tau)}}^{2},
  \end{align*}
  and so with
  $p_{1,\tau}=\frac{\mu_{\tau}^{2}}{\sum_{\tilde\tau\in\cS}\mu_{\tilde\tau}^{2}}$, we
  have
  \begin{align*}
    R_{T}^{\text{Meta}}(p_{\tau}^{*})
    &\le
      2k D\brac{G_{\max}+\frac{ D}{ \eta}}\sbrac{\Log{\frac{\sum_{\tilde\tau\in\cS}\mu_{\tilde \tau}^{2}}{\mu_{\tau}^{2}}}+1}
      +2\eta\sumtT\norm{g_{t}^{(\tau)}}^{2}
    +2k\sum_{\tilde\tau\in\cS}\frac{\mu_{\tilde\tau}}{\sum_{\tilde\tau\in\cS}\mu_{\tilde\tau}^{2}}.
  \end{align*}
  Combining this with \Cref{eq:dynamic-untuned-1,eq:dynamic-untuned-2} yields
  the stated result:
  \begin{align*}
    R_{T}(\vec{\cmp})
    &\le
      \frac{\norm{\cmp_{T}}^{2}+2DP_{T}}{2\eta}
      +K\eta\sumtT L_{t}\sbrac{\ell_{t}(\cmp_{t})-\ell_{t}(\wt^{(\tau)})} +4\eta\sumtT\norm{g_{t}^{(\tau)}}^{2}\\
    &\qquad
      +2k D\brac{G_{\max}+\frac{ D}{ \eta}}\sbrac{\Log{\frac{\sum_{\tilde\tau\in\cS}\mu_{\tilde \tau}^{2}}{\mu_{\tau}}}+1}
    +2k\frac{\sum_{\tilde\tau\in\cS}\mu_{\tilde\tau}}{\sum_{\tilde\tau\in\cS}\mu_{\tilde\tau}^{2}}\\
    &=
      2k C_{\cS}+2kDG_{\max}\Lambda_{T}(\tau)
      +\frac{\norm{\cmp_{T}}^{2}+2DP_{T}+4kD^{2}\Lambda_{T}(\tau)}{2\eta}\\
    &\quad
      +K\eta\sumtT L_{t}\sbrac{\ell_{t}(\cmp_{t})-\ell_{t}(\wt^{(\tau)})} +4\eta\sumtT\norm{g_{t}^{(\tau)}}^{2}\\
  \end{align*}
  where the last line defines the shorthand notations
  \begin{align*}
    C_{\cS}&\defeq\frac{\sum_{\tilde\tau\in\cS}\mu_{\tilde\tau}}{\sum_{\tilde\tau\in\cS}\mu_{\tilde\tau}^{2}}\qquad\text{and}\qquad
    \Lambda_{T}(\tau)\defeq \Log{\frac{\sum_{\tilde\tau\in\cS}\mu_{\tilde \tau}^{2}}{\mu_{\tau}^{2}}}+1.
  \end{align*}

\end{proof}

Next, we provide bounds on the terms
terms $C_{\cS}$ and $\Lambda_{T}$ in terms of the
hyperparameter ranges $[\eta_{\min},\eta_{\max}]$ and $[D_{\min},D_{\max}]$
that the meta-algorithm tunes the hyperparameters over.
\begin{restatable}{lemma}{MuBounds}\label{lemma:mu-bounds}
  Let $0<\eta_{\min}\le \eta_{\max}$, $0<D_{\min}\le D_{\max}$,
  and define the hyperparameter set $\cS=\cS_{\eta}\times\cS_{D}$ for
  $\cS_{\eta}=\Set{\eta_{i}=\sbrac{\eta_{\min}2^{i}\minOp \eta_{\max}}:i\ge 0}$ and
  $\cS_{D}=\Set{D_{j}=\sbrac{D_{\min}2^{j}\minOp D_{\max}}:j\ge 0 }$.
  For each $\tau=(\eta,D)\in\cS$, let
  $\mu_{\tau}=\frac{1}{2D\brac{G_{\max}+D/\eta}}$.
  Then
  \begin{align*}
    C_{\cS}&\defeq\frac{\sum_{\tau\in\cS}\mu_{\tau}}{\sum_{\tau\in\cS}\mu_{\tau}^{2}}
    \le
      2\sqrt{T}D_{\min}\brac{G_{\max}+\frac{D_{\min}}{\eta_{\max}}}
  \end{align*}
  and for any $\tau\in\cS$,
  \begin{align*}
    \Lambda_{T}(\tau)&\defeq \Log{\frac{\sum_{\tilde \tau\in\cS}\mu_{\tilde \tau}^{2}}{\mu_{\tau}^{2}}}+1
    \le
      \Log{\frac{24\eta_{\max}^{2}D^{4}}{\eta_{\min}^{2}D_{\min}^{4}}\minOp \frac{6\abs{\cS_{\eta}}D^{2}}{D_{\min}^{2}}}+1
  \end{align*}
\end{restatable}
\begin{proof}
  For the first statement, we have
  \begin{align*}
    C_{\cS}
    &=\frac{\sum_{\tilde\tau\in \cS}\mu_{\tilde \tau}}{\sum_{\tilde\tau\in\cS}\mu_{\tilde \tau}^{2}}
    \le
      \sqrt{\frac{T}{\sum_{\tilde\tau\in\cS}\mu_{\tilde\tau}^{2}}}
    \le
      \sqrt{\frac{T}{\mu_{(\eta_{\max},D_{\min})}^{2}}}\\
      &=
      \sqrt{T (2D_{\min})^{2}\brac{G_{\max}+D_{\min}/\eta_{\max}}^{2}}\\
    &=
      2\sqrt{T}D_{\min}\brac{G_{\max}+\frac{D_{\min}}{\eta_{\max}}}
  \end{align*}
  where the first inequality applies Cauchy-Schwarz inequality.
  Moreover, for any $\tau=(\eta,D)\in \cS$ we have
  \begin{align*}
    \frac{\sum_{\tilde\tau\in\cS}\mu_{\tilde\tau}^{2}}{\mu_{\tau}^{2}}
    &=
      \frac{1}{\mu_{(\eta,D)}^{2}}\sbrac{\sum_{(\eta_{i},D_{j})\in \cS}\frac{1}{(2D_{j})^{2}\sbrac{G_{\max}+D_{j}/\eta_{i}}^{2}}}\\
    &\le
      \frac{1}{4\mu_{(\eta,D)}^{2}}\sum_{(\eta_{i},D_{j})\in \cS}\frac{\eta_{i}^{2}}{D_{j}^{4}}\\
    &=
      \frac{1}{4\mu_{(\eta,D)}^{2}}\sum_{(\eta_{i},D_{j})\in\cS}\frac{2^{2i}\eta_{\min}^{2}}{D_{\min}^{4}2^{4j}}\\
    &=
      \frac{\eta_{\min}^{2}}{4\mu_{(\eta,D)}^{2}D_{\min}^{4}}\sum_{i=0}^{\lceil\log_{2}(\eta_{\max}/\eta_{\min})\rceil}\sum_{j=0}^{\lceil\log_{2}(D_{\max}/D_{\min})\rceil}\frac{2^{2i}}{2^{4j}}\\
      &\le
      \frac{\eta_{\min}^{2}}{4\mu_{(\eta,D)}^{2}D_{\min}^{4}}\frac{2^{2\lceil\log_{2}(\eta_{\max}/\eta_{\min})\rceil+2}-1}{3}\frac{1}{1-\frac{1}{16}}\\
    &\le
      \frac{\eta_{\min}^{2}}{4\mu_{(\eta,D)}^{2}D_{\min}^{4}}\frac{2^{\log_{2}(\eta_{\max}^{2}/\eta_{\min}^{2})+4}-1}{3}\frac{16}{15}\\
    &\le
      \frac{\eta_{\min}^{2}}{4\mu_{(\eta,D)}^{2}D_{\min}^{4}}16\frac{\eta_{\max}^{2}}{\eta_{\min}^{2}}\frac{16}{45}\\
    &\le\frac{6\eta_{\max}^{2}}{4\mu_{(\eta,D)}^{2}D_{\min}^{4}}
    =\frac{3\eta_{\max}^{2}}{2\mu_{(\eta,D)}^{2}D_{\min}^{4}}\\
  \end{align*}
  At the same time, we can also bound this term as
  \begin{align*}
    \frac{\sum_{\tilde\tau\in\cS}\mu_{\tilde\tau}^{2}}{\mu_{\tau}^{2}}
    &=
      \frac{1}{\mu^{2}_{(\eta,D)}}\sbrac{\sum_{(\eta_{i},D_{j})\in\cS}\frac{1}{(2D_{j})^{2}\sbrac{G_{\max}+D_{j}/\eta_{i}}^{2}}}\\
    &\le
      \frac{1}{4\mu^{2}_{(\eta,D)}}\sum_{(\eta_{i},D_{j})\in\cS}\frac{1}{D_{j}^{2}G_{\max}^{2}}\\
    &\le
      \frac{\abs{\cS_{\eta}}}{4\mu^{2}_{(\eta,D)}G_{\max}^{2}}\sum_{j=0}^{\lceil\log_{2}(D_{\max}/D_{\min})\rceil}\frac{1}{D_{\min}^{2} 2^{2j}}\\
    &\le
      \frac{\abs{\cS_{\eta}}}{4\mu^{2}_{(\eta,D)}G_{\max}^{2}D_{\min}^{2}}\frac{1}{1-\frac{1}{4}}\\
    &\le
      \frac{4\abs{\cS_{\eta}}}{4\cdot 3\mu^{2}_{(\eta,D)}G_{\max}^{2}D_{\min}^{2}}
      =
      \frac{\abs{\cS_{\eta}}}{3\mu^{2}_{(\eta,D)}G_{\max}^{2}D_{\min}^{2}}
  \end{align*}
  Hence,
  \begin{align*}
    \Lambda_{T}(\eta,D)
    &=
      \Log{\frac{\sum_{\tilde\tau}\mu_{\tilde\tau}}{\mu_{\tau}^{2}}}+1\\
    &\le
      \Log{\sbrac{\frac{3\eta_{\max}^{2}}{2D_{\min}^{2}}\minOp \frac{\abs{\cS_{\eta}}}{3G_{\max}^{2}}}\frac{1}{\mu_{(\eta,D)}^{2}D_{\min}^{2}}}+1\\
    &=
      \Log{\sbrac{\frac{3\eta_{\max}^{2}}{2D_{\min}^{2}}\minOp \frac{\abs{\cS_{\eta}}}{3G_{\max}^{2}}}\frac{(2D)^{2}\sbrac{G_{\max}+D/\eta}^{2}}{D_{\min}^{2}}}+1.
  \end{align*}
  Now if $G_{\max}\le D/\eta$, we have
  \begin{align*}
    \Lambda_{T}(\eta,D)
    &\le
      \Log{\frac{3\cdot 4\cdot\eta_{\max}^{2}D^{2}\sbrac{G_{\max}+D/\eta}^{2}}{2D_{\min}^{4}}}+1\\
    &\le
      \Log{\frac{6\eta_{\max}^{2}D^{2}\cdot (2D/\eta)^{2}}{D_{\min}^{4}}}+1\\
    &\le
      \Log{\frac{24\eta_{\max}^{2}D^{4}}{\eta^{2}_{\min}D_{\min}^{4}}}+1
  \end{align*}
  and otherwise
  \begin{align*}
    \Lambda_{T}(\eta,D)
    &\le
      \Log{\frac{4\abs{\cS_{\eta}}D^{2}\sbrac{G_{\max}+D/\eta}^{2}}{3G_{\max}^{2}D_{\min}^{2}}}+1\\
    &\le
      \Log{\frac{6\abs{\cS_{\eta}}D^{2}G_{\max}^{2}}{G_{\max}^{2}D_{\min}^{2}}}+1\\
    &=
      \Log{\frac{6\abs{\cS_{\eta}}D^{2}}{D_{\min}^{2}}}+1.
  \end{align*}
  Thus, we can bound
  \begin{align*}
    \Lambda_{T}(\eta,D)
    &\le
      \Log{\frac{24\eta_{\max}^{2}D^{4}}{\eta_{\min}^{2}D_{\min}^{4}}\minOp \frac{6\abs{\cS_{\eta}}D^{2}}{D_{\min}^{2}}}+1
  \end{align*}
\end{proof}

\Cref{lemma:omega-bound} provides a simple but tedius
calculation which we will use a few times in the proof of \Cref{thm:dynamic}.
\begin{restatable}{lemma}{OmegaBound}\label{lemma:omega-bound}
  Let $\ell_{t}$ be $(G_{t},L_{t})$-quadratically bounded, $c_{1},c_{2}\ge 0$, $\cmp,\w\in W$, and $g_{t}\in\partial\ell_{t}(\w)$.
  Assume $\norm{\w}\le D$ and $\norm{\cmp}\le D$. Then
  \begin{align*}
    c_{1}L_{t}\sbrac{\ell_{t}(\cmp)-\ell_{t}(\w)}+c_{2}\norm{g_{t}}^{2}
    &\le
      3(c_{1}+c_{2})\brac{G_{t}^{2}+L_{t}^{2}D^{2}}
  \end{align*}
\end{restatable}
\begin{proof}
  Since $\ell_{t}$ is $(G_{t},L_{t})$-quadratically bounded, and
  $g_{t}\in\partial\ell_{t}(\w)$ where $\norm{\w}\le D$ we have
  \begin{align*}
    \norm{g_{t}}^{2}
    &\le
      (G_{t}+L_{t}\norm{\w})^{2}
    \le 2G_{t}^{2}+2L_{t}^{2}\norm{\w}^{2}
    \le
      2G_{t}^{2}+2L_{t}^{2}D^{2}.
  \end{align*}
  Moreover, letting $\grad\ell_{t}(\cmp)\in\partial\ell_{t}(\cmp)$ and
  $\norm{\cmp}\le D$ we have
  \begin{align*}
    L_{t}\brac{\ell_{t}(\cmp)-\ell_{t}(\w)}
    &\le
      L_{t}\norm{\grad\ell_{t}(\cmp)}\norm{\cmp-\w}\\
    &\le 2DL_{t}\norm{\grad\ell_{t}(\cmp)}\\
    &\le
      2DL_{t}\brac{G_{t}+L_{t}D}\\
    &=
      2DL_{t}G_{t}+2L_{t}^{2}D^{2}\\
    &\le
      G_{t}^{2}+L_{t}^{2}D^{2}+2L_{t}^{2}D^{2}\\
    &=
      G_{t}^{2}+3L_{t}^{2}D^{2}.
  \end{align*}
  Thus,
  \begin{align*}
    c_{1}L_{t}\brac{\ell_{t}(\cmp)-\ell_{t}(\w)}
    +c_{2}\norm{g_{t}}^{2}
    &\le
      (c_{1}+2c_{2})G_{t}^{2}+(3c_{1}+2c_{2})L_{t}^{2}D^{2}\\
    &\le
      3(c_{1}+c_{2})\brac{G_{t}^{2}+L_{t}^{2}D^{2}}\\
  \end{align*}
\end{proof}

Lastly, we provide two lemmas which let us assume that
there is a $\tau=(\eta,D)\in\cS$ for which $\half D\le M=\max_{t}\norm{\cmp_{t}}\le D$ by showing that
the regret is trivially well-controlled whenever $M$ is ``too big''
(\Cref{lemma:large-M}) or ``too small'' (\Cref{lemma:small-M}).
\begin{restatable}{lemma}{LargeM}\label{lemma:large-M}
  For all $t$ let $\ell_{t}$ be a $(G_{t},L_{t})$-quadratically bounded
  convex function for $G_{t}\in[0,G_{\max}]$ and $L_{t}\in[0,L_{\max}]$. Let
  $\eps>0$,
  $D_{\max}=\eps 2^{T}$, and
  let $\vec{\cmp}=(\cmp_{1},\ldots,\cmp_{T})$  be an
  arbitrary sequence in $W$ such that
  $M:=\max_{t}\norm{\cmp_{t}}\ge D_{\max}$.
  Then for any $\w_{1},\ldots,\w_{T}$ with $\norm{\wt}\le D_{\max}$,
  \begin{align*}
    \sumtT\ell_{t}(\wt)-\ell_{t}(\cmp_{t})
    &\le
      2\brac{G_{\max}M+L_{\max}M^{2}}\log_{2}\brac{\frac{M}{\eps}}.
  \end{align*}
\end{restatable}
\begin{proof}
  Let $g_{t}\in \partial\ell_{t}(\wt)$ and observe that
  \begin{align*}
    \sumtT\ell_{t}(\wt)-\ell_{t}(\cmp_{t})
    &\le
      \sumtT\norm{g_{t}}\norm{\wt-\cmp_{t}}\\
    &\le
      \sumtT\norm{g_{t}}\brac{D_{\max}+\norm{\cmp_{t}}}\\
    &\le
      2M\sumtT\norm{g_{t}}\\
    &\le
      2M\brac{G_{\max}+L_{\max}D_{\max}}T\\
    &\le
      2M\brac{G_{\max}+L_{\max}M}T\\
      &\le
      2\brac{G_{\max}M+L_{\max}M^{2}}\log_{2}\brac{\frac{M}{\eps}},
  \end{align*}
  where the last line uses $M\ge \eps 2^{T}\implies T\le \log_{2}\brac{\frac{M}{\eps}}$.
\end{proof}

\begin{restatable}{lemma}{SmallM}\label{lemma:small-M}
  For all $t$ let $\ell_{t}$ be a $(G_{t},L_{t})$-quadratically bounded
  convex function for $G_{t}\in[0,G_{\max}]$ and $L_{t}\in[0,L_{\max}]$. Let
  $\eps>0$,
  $D_{\min}=\frac{\eps}{T}$,
  $\eta_{\max}=\frac{1}{KL_{\max}}$,
  and
  $\eta_{\min}=\frac{\epsilon}{K(G_{\max}+\epsilon L_{\max})T}$. Let $\wt\in W$ be the
  outputs of
  the algorithm characterized in \Cref{lemma:dynamic-untuned}
  with $\eta=\eta_{\min}$ and $D=D_{\min}$, and
  let $\vec{\cmp}=(\cmp_{1},\ldots,\cmp_{T})$  be an
  arbitrary sequence in $W$ with $M=\max_{t}\norm{\cmp_{t}}\le D_{\min}$.
  Then
  \begin{align*}
    R_{T}(\vec{\cmp})
    &\le
      (G_{\max}+\epsilon L_{\max})\sbrac{K(M+P_{T})+\epsilon \cC_{T}}
  \end{align*}
  where $\cC_{T}\le O\brac{\frac{\Log{\Log{\frac{G_{\max}}{\epsilon L_{\max}}}}}{T}}$.
\end{restatable}
\begin{proof}
  For $M\le D_{\min}$, we can apply \Cref{lemma:dynamic-untuned}
  with $\tau=(\eta_{\min},D_{\min})$ to get
  \begin{align*}
    R_{T}(\vec{\cmp})
    &\le
      2kC_{\cS}+2kD_{\min}G_{\max}\Lambda_{T}(\tau)
      +\frac{\norm{\cmp_{T}}^{2}+2D_{\min}P_{T}+4kD_{\min}^{2}\Lambda_{T}(\tau)}{2\eta_{\min}}\\
    &\quad
      +K\eta_{\min}\sumtT L_{t}\sbrac{\ell_{t}(\cmp_{t})-\ell_{t}(\wt^{(\tau)})} +4\eta_{\min}\sumtT\norm{g_{t}^{(\tau)}}^{2},
  \end{align*}
  where 
  $\mu_{\tilde\tau}=\frac{1}{2\tilde D(G_{\max}+\tilde D/\tilde \eta)}$ for any
  $\tilde\tau=(\tilde \eta,\tilde D)\in \cS$, $k\ge 9/2$, and
  \begin{align*}
    C_{\cS}&\defeq\frac{\sum_{\tilde\tau\in\cS}\mu_{\tilde\tau}}{\sum_{\tilde\tau\in\cS}\mu_{\tilde\tau}^{2}},\qquad
    \Lambda_{T}(\tau)=\Lambda_{T}(\eta_{\min},D_{\min})\defeq \Log{\frac{\sum_{\tilde\tau\in\cS}\mu_{\tilde \tau}^{2}}{\mu_{(\eta_{\min},D_{\min})^{2}}}}+1.
  \end{align*}
  Observe that with $M=\max_{t}\norm{\cmp_{t}}\le D_{\min}$ and
  $\frac{D_{\min}}{\eta_{\min}}=K(G_{\max}+\epsilon L_{\max})$, we have
  \begin{align*}
      \frac{\norm{\cmp_{T}}^{2}+2D_{\min}P_{T}+4kD_{\min}^{2}\Lambda_{T}(\tau)}{2\eta_{\min}}
    &\le
      \frac{D_{\min}}{\eta_{\min}}\half\brac{\norm{\cmp_{T}}+2P_{T}+4kD_{\min}\Lambda_{T}(\tau)}\\
    &=
      \half K(G_{\max}+\epsilon L_{\max})\brac{\norm{\cmp_{T}}+2P_{T}+4k\frac{\epsilon \Lambda_{T}(\tau)}{T}}.
  \end{align*}
  Moreover, by \Cref{lemma:omega-bound} we have
  \begin{align*}
    \sumtT K\eta_{\min}L_{t}\sbrac{\ell_{t}(\cmp_{t})-\ell_{t}(\wt^{(\tau)})}+4\eta_{\min}\norm{g_{t}^{(\tau)}}^{2}
    &\le
      \eta_{\min}\sumtT 3(K+4)\brac{G_{t}^{2}+L_{t}^{2}D_{\min}^{2}}\\
    &\le
      3(K+4)\frac{\epsilon\brac{G_{\max}^{2}+L_{\max}^{2}D_{\min}^{2}}}{K(G_{\max}+\epsilon L_{\max})}\\
    &\le
      \frac{3(K+4)}{K}\brac{\epsilon G_{\max}+\frac{\epsilon^{2}L_{\max}}{T^{2}}}
  \end{align*}

  Plugging in the previous two displays
  back into the full regret bound yields
  \begin{align*}
    R_{T}(\vec{\cmp})
    &\le
      2kC_{\cS}+2k\epsilon G_{\max}\frac{\Lambda_{T}(\tau)}{T}
    +\half K(G_{\max}+\epsilon L_{\max})\brac{\norm{\cmp_{T}}+2P_{T}+4k\frac{\epsilon \Lambda_{T}(\tau)}{T}}\\
    &\qquad+
      \frac{3(K+4)}{K}\sbrac{\epsilon G_{\max}+\frac{L_{\max}\epsilon^{2}}{T^{2}}}\\
    &\le
      2kC_{\cS}
      +\epsilon G_{\max}\sbrac{\frac{3(K+4)}{K}+\frac{(K+1) 2k\Lambda_{T}(\tau)}{T}}
      +\epsilon^{2}L_{\max}\sbrac{\frac{3(K+4)}{KT^{2}}+\frac{2k K\Lambda_{T}(\tau)}{T}}\\
    &\qquad
      +K(G_{\max}+\epsilon L_{\max})\sbrac{M+P_{T}}.
  \end{align*}
  Finally,
  \Cref{lemma:mu-bounds} bounds
  \begin{align*}
    2kC_{\cS}
    &\le
      2k\cdot 2\sqrt{T}D_{\min}\brac{G_{\max}+\frac{D_{\min}}{\eta_{\max}}}\\
    &\le
      4k\sqrt{T}\frac{\epsilon}{T}\sbrac{G_{\max}+\frac{K\epsilon L_{\max}}{T}}\\
    &\le
      \frac{4k\brac{\epsilon G_{\max}+K\epsilon^{2}L_{\max}/T}}{\sqrt{T}}
  \end{align*}
  and
  \begin{align*}
    \Lambda_{T}(\eta_{\min},D_{\min})
    &\le
      \Log{6\abs{\cS_{\eta}}}+1
    \le \Log{\abs{\cS_{\eta}}}+3\\
    &\le
      \Log{\Ceil{\log_{2}\brac{\frac{TG_{\max}}{\epsilon L_{\max}}}}+1}+3\\
    &\le
      \Log{\log_{2}\brac{\frac{TG_{\max}}{\epsilon L_{\max}}}+2}+3\\
  \end{align*}
  Plugging these back in above:
  \begin{align*}
    R_{T}(\vec{\cmp})
    &\le
      \epsilon G_{\max}(K+4)\sbrac{\frac{3}{K}+\frac{k}{\sqrt{T}}+\frac{ 2k\Lambda_{T}(\eta_{\min},D_{\min})}{T}}\\
    &\qquad
      +\epsilon^{2}L_{\max}(K+4)\sbrac{\frac{3}{KT^{2}}+\frac{4k}{T^{3/2}}+\frac{2k \Lambda_{T}(\eta_{\min},D_{\min})}{T}}\\
    &\qquad
      +K(G_{\max}+\epsilon L_{\max})\sbrac{M+P_{T}}\\
    &\le
      \cC_{T}\brac{\epsilon G_{\max}+\epsilon^{2}L_{\max}}
      +K(G_{\max}+\epsilon L_{\max})\sbrac{M+P_{T}}\\
    &=
      (G_{\max}+\epsilon L_{\max})\sbrac{K(M+P_{T})+\epsilon \cC_{T}}
  \end{align*}
  where
  \begin{align*}
    \cC_{T}
    &\le
      (K+4)\brac{\frac{3}{K}+\frac{4k}{\sqrt{T}}+\frac{2k\brac{\Log{\log_{2}\brac{\frac{TG_{\max}}{\epsilon L_{\max}}}+2}+3}}{T}}\\
    &\le O\brac{\frac{\Log{\Log{\frac{G_{\max}}{\epsilon L_{\max}}}}}{T}}
  \end{align*}

\end{proof}

\subsection{Proof of Theorem~\ref{thm:dynamic}}
\label{app:dynamic-thm}
\begin{minipage}{\columnwidth}
\begin{manualtheorem}{\ref{thm:dynamic}}
  For all $t$ let $\ell_{t}:W\to \R$ be a $(G_{t},L_{t})$-quadratically bounded
  convex function with $G_{t}\in[0,G_{\max}]$ and $L_{t}\in[0,L_{\max}]$.
  Let $\epsilon>0$, $K\ge 8$, $\beta_{t}=1-\Exp{-1/T}$ for all $t$, and for any $i,j\ge 0$ let
  $D_{j}=\frac{\epsilon}{T} \sbrac{2^{j}\minOp 2^{T}}$ and
  $\eta_{i}=\sbrac{\frac{\epsilon 2^{i}}{K\brac{G_{\max}+\epsilon L_{\max}}T}\minOp\frac{1}{KL_{\max}}}$,
  and let $\cS=\Set{(\eta_{i},D_{j}):i,j\ge 0}$.
  For each $\tau=(\eta,D)\in\cS$ let
  $\mu_{\tau}=\frac{1}{2D\brac{G_{\max}+D/\eta}}$,
  and set
  $p_{1}(\tau)=\frac{\mu_{\tau}^{2}}{\sum_{\tilde\tau\in\cS}\mu_{\tilde\tau}^{2}}$.
  Then for any
  $\vec{\cmp}=(\cmp_{1},\ldots,\cmp_{T})$ in $W$,
  \Cref{alg:dynamic-meta} guarantees
  \begin{align*}
    R_{T}(\vec{\cmp})
    &\le
      O\Bigg(
      \bigg[G_{\max}+(M+\epsilon)L_{\max}\bigg]\bigg[(M+\epsilon)\Lambda_{T}^{*}+P_{T}\bigg]
      +\sqrt{(M^{2}\Lambda_{T}^{*}+MP_{T})\sumtT G_{t}^{2}+L_{t}^{2}M^{2}},
      \Bigg).
  \end{align*}
  where
  $P_{T}=\sum_{t=2}^{T}\norm{\cmp_{t}-\cmp_{\tmm}}$,
  $M=\max_{t}\norm{\cmp_{t}}$, and
  $\Lambda_{T}^{*}\le O\brac{\Log{\frac{MT\Log{T}}{\epsilon}}+\Log{\Log{\frac{G_{\max}}{\epsilon L_{\max}}}}}$.
  Moreover,
  when the losses are $L_{t}$-smooth, the bound automatically improves to
  \begin{align*}
    R_{T}(\vec{\cmp})
    &\le
      O\Bigg(
      \bigg[G_{\max}+(M+\epsilon)L_{\max}\bigg]\bigg[(M+\epsilon)\Lambda_{T}^{*}+P_{T}\bigg]\\
  &\qquad\qquad
      +\sqrt{(M^{2}\Lambda_{T}^{*}+MP_{T})\sbrac{\sumtT L_{t}\sbrac{\ell_{t}(\cmp_{t})-\ell_{t}^{*}}\minOp \sumtT G_{t}^{2}+L_{t}^{2}M^{2}}}
      \Bigg).
  \end{align*}
\end{manualtheorem}
\end{minipage}
\begin{proof}
  First observe that we can assume that there is
  a $\tau=(\eta,D)\in\cS$ for which $D\ge \max_{t}\norm{\cmp_{t}}=M$,
  since otherwise using \Cref{lemma:large-M} with $\eps=\frac{\epsilon}{T}$
  the regret is bounded as
  \begin{align}
    R_{T}(\vec{\cmp})\le 2M\brac{G_{\max}+ML_{\max}}\Log{\frac{MT}{\epsilon}}.\label{eq:dynamic:large-M}
  \end{align}
  Likewise, if $M\le D_{\min}$ then by \Cref{lemma:small-M} we have
  \begin{align}
    R_{T}(\vec{\cmp})\le (G_{\max}+L_{\max}\epsilon)\sbrac{K(M+P_{T})+\epsilon \cC_{T}},\label{eq:dynamic:small-M}
  \end{align}
  where $\cC_{T}\le O\brac{\frac{\Log{\Log{\frac{G_{\max}}{\epsilon L_{\max}}}}}{T}}$. Otherwise, we have $M\in[D_{\min},D_{\max}]$,
  in which case there is a $D_{j}=\frac{\epsilon 2^{j}}{T}$ for which
  $D_{j}\ge M\ge D_{j-1}=\half D_{j}$, so for any
  $\tau=(\eta,D_{j})\in\cS$ we can apply
  \Cref{lemma:dynamic-untuned} to get
  \begin{align*}
    R_{T}(\vec{\cmp})
    &\le
      2kC_{\cS}+2kD_{j}G_{\max}\Lambda_{T}(\tau)
      +\frac{\norm{\cmp_{T}}^{2}+2D_{j}P_{T}+4kD_{j}^{2}\Lambda_{T}(\tau)}{2\eta}\\
    &\quad
      +K\eta\sumtT L_{t}\sbrac{\ell_{t}(\cmp_{t})-\ell_{t}(\wt^{(\tau)})} +4\eta\sumtT\norm{g_{t}^{(\tau)}}^{2},
  \end{align*}
  where $g_{t}^{(\tau)}\in\partial\ell_{t}(\wt^{(\tau)})$,
  $P_{T}=\sum_{t=2}^{T}\norm{\cmp_{t}-\cmp_{\tmm}}$, and
  \begin{align*}
    C_{\cS}
    &=
      \frac{\sum_{\tilde\tau\in\cS}\mu_{\tilde\tau}}{\sum_{\tilde\tau\in\cS}\mu_{\tilde\tau}}\\
    \Lambda_{T}(\eta,D_{j})
    &=
      \Log{\frac{\sum_{\tilde\tau\in\cS}\mu_{\tilde\tau}^{2}}{\mu_{(\eta,D_{j})}^{2}}}+1\\
    &=
      \Log{D_{j}^{2}\sbrac{G_{\max}+\frac{D_{j}}{\eta}}^{2}\sum_{\tilde\tau\in\cS}\mu_{\tilde\tau}^{2}}+1\\
    &\le
      \Log{(2M)^{2}\sbrac{G_{\max}+\frac{2M}{\eta_{\min}}}^{2}\sum_{\tilde\tau\in\cS}\mu_{\tilde\tau}^{2}}+1\\
    &=
      \Lambda_{T}(\eta_{\min},2M).
  \end{align*}
  Thus, bounding $D_{j}\le 2M$ and denoting
  $\Omega_{T}:=\sumtT KL_{t}\sbrac{\ell_{t}(\cmp_{t})-\ell_{t}(\wt^{(\tau)})}+4\norm{g_{t}^{(\tau)}}^{2}$,
  we have:
  \begin{align}
    R_{T}(\vec{\cmp})
    &\le
      2kC_{\cS}+4kMG_{\max}\Lambda_{T}(\eta_{\min},2M)
      +\frac{M^{2}\brac{1+16k\Lambda_{T}(\eta_{\min},2M)}+4MP_{T}}{2\eta}\nonumber\\
    &\qquad
      \eta\underbrace{\sumtT \sbrac{KL_{t}\sbrac{\ell_{t}(\cmp_{t})-\ell_{t}(\wt^{(\tau)})}+4\norm{g_{t}^{(\tau)}}^{2}}}_{=:\Omega_{T}}
      .\label{eq:dynamic:untuned-eta}
  \end{align}
  Next, we show that there is an $\eta$ for which the above expression is well-controlled.

  Observe that choosing $\eta$ optimally in \Cref{eq:dynamic:untuned-eta} would yield
  \begin{align*}
    \eta^{*}=\sqrt{\frac{M^{2}(1+16k\Lambda_{T}(\eta_{\min},2M))+4MP_{T}}{2\Omega_{T}}}.
  \end{align*}
  If $\eta^{*}\ge\eta_{\max}$, then choosing $\eta=\eta_{\max}$ yields
  \begin{align}
    R_{T}(\vec{\cmp})
    &\le
      2kC_{\cS}+4kMG_{\max}\Lambda_{T}(\eta_{\min},2M)\nonumber
      +\frac{M^{2}\brac{1+16k\Lambda_{T}(\eta_{\min},2M)}+4MP_{T}}{2\eta_{\max}}
      +\eta^{*}\Omega_{T}\nonumber\\
    &=
      2kC_{\cS}+4kMG_{\max}\Lambda_{T}(\eta_{\min},2M)
      +\frac{KL_{\max}}{2}\sbrac{M^{2}(1+16k\Lambda_{T}(\eta_{\min},2M))+4MP_{T}}\nonumber\\
    &\qquad
      +\sqrt{\half\sbrac{M^{2}\brac{1+16k\Lambda_{T}(\eta_{\min},2M)}+4MP_{T}}\Omega_{T}}.\label{eq:dynamic:large-eta}
  \end{align}
  Similarly,
  if $\eta^{*}\le \eta_{\min}$, then choosing $\eta=\eta_{\min}$ yields
  \begin{align*}
    R_{T}(\vec{\cmp})
    &\le
      2kC_{\cS}+4kMG_{\max}\Lambda_{T}(\eta_{\min},2M)
      +\frac{M^{2}\brac{1+16k\Lambda_{T}(\eta_{\min},2M)}+4MP_{T}}{2\eta^{*}}
      +\eta_{\min} \Omega_{T}\nonumber\\
    &=
      2kC_{\cS}+4kMG_{\max}\Lambda_{T}(\eta_{\min},2M)
      +\sqrt{\half\sbrac{M^{2}\brac{1+16k\Lambda_{T}(\eta_{\min},2M)}+4MP_{T}}\Omega_{T}}\nonumber\\
    &\quad
      +\frac{\epsilon\Omega_{T}}{K\brac{G_{\max}+\epsilon L_{\max}}T}.
  \end{align*}
  Observe that by \Cref{lemma:omega-bound}, we have
  \begin{align*}
    \Omega_{T}
    &=
       \sumtT KL_{t}\sbrac{\ell_{t}(\cmp_{t})-\ell_{t}(\wt^{(\tau)})}+4\norm{g_{t}^{(\tau)}}^{2}\\
      &\le
        \sumtT3(K+4)\brac{G_{\max}^{2}+L_{\max}^{2}D_{j}^{2}}\\
    &\le
      3(K+4)\brac{G_{\max}^{2}+4M^{2}L_{\max}^{2}}T.
  \end{align*}
  Thus
  \begin{align*}
    \frac{\epsilon\Omega_{T}}{K\brac{G_{\max}+\epsilon L_{\max}}T}
    &\le
      \frac{\epsilon\cdot3(K+4)\brac{G_{\max}^{2}+4M^{2}L_{\max}^{2}}T}{K\brac{G_{\max}+\epsilon L_{\max}}T}\\
    &\le
        \frac{3(K+4)}{K}\brac{\epsilon G_{\max}+4M^{2}L_{\max}}\\
    &\le
        (K+4)\brac{\epsilon G_{\max}+4M^{2}L_{\max}}
  \end{align*}
  for $K\ge 3$.
  so overall when $\eta^{*}\le\eta_{\min}$ the regret can be bounded as
  \begin{align}
    R_{T}(\vec{\cmp})
    &\le
      2kC_{\cS}+4kMG_{\max}\Lambda_{T}(\eta_{\min},2M)
      +\sqrt{\half\sbrac{M^{2}\brac{1+16k\Lambda_{T}(\eta_{\min},2M)}+4MP_{T}}\Omega_{T}}\nonumber\\
    &\qquad
      +(K+4)\epsilon G_{\max}+4(K+4)M^{2}L_{\max}.\label{eq:dynamic:small-eta}
  \end{align}
  Finally, if $\eta^{*}\in[\eta_{\min},\eta_{\max}]$, then there is an
  $\eta_{i}=\frac{2^{i}\epsilon }{K\brac{G_{\max}+\epsilon L_{\max}}T}$ such
  that $\eta_{i}\le \eta^{*}\le \eta_{i+1}=2\eta_{i}$, so choosing
  $\eta=\eta_{i}$ \Cref{eq:dynamic:untuned-eta} is bounded by
  \begin{align}
    R_{T}(\vec{\cmp})
    &\le
      2kC_{\cS}+4kMG_{\max}\Lambda_{T}(\eta_{\min},2M)
      +\frac{M^{2}\brac{1+16k\Lambda_{T}(\eta_{\min},2M)}+4MP_{T}}{\eta^{*}}
      +\eta^{*}\Omega_{T}\nonumber\\
      &\le
      2kC_{\cS}+4kMG_{\max}\Lambda_{T}(\eta_{\min},2M)
      +3\sqrt{\half\sbrac{M^{2}\brac{1+16k\Lambda_{T}(\eta_{\min},2M)}+4MP_{T}}\Omega_{T}}\label{eq:dynamic:tuned-eta}.
  \end{align}
  Now combining \Cref{eq:dynamic:large-M,eq:dynamic:small-M,eq:dynamic:large-eta,eq:dynamic:small-eta,eq:dynamic:tuned-eta}, we have
  \begin{align*}
    R_{T}(\vec{\cmp})
    &\le
      2M\brac{G_{\max}+ML_{\max}}\Log{\frac{MT}{\epsilon}}\\
    &\quad
      +(G_{\max}+L_{\max}\epsilon)\sbrac{K(M+P_{T})+\epsilon \cC_{T}}\\
    &\quad
      +2kC_{\cS}+4kMG_{\max}\Lambda_{T}(\eta_{\min},2M)\nonumber\\
    &\quad
      +3\sqrt{\half\sbrac{M^{2}\brac{1+16k\Lambda_{T}(\eta_{\min},2M)}+4MP_{T}}\Omega_{T}}\\
    &\quad
      +(K+4)\epsilon G_{\max}+4(K+4)M^{2}L_{\max}\\
    &\quad
      +\frac{KL_{\max}}{2}\sbrac{M^{2}(1+16k\Lambda_{T}(\eta_{\min},2M))+4MP_{T}}\nonumber.
    \end{align*}
    From \Cref{lemma:mu-bounds} we have
    \begin{align*}
      &C_{\cS}
      \le
        2\sqrt{T}D_{\min}\brac{G_{\max}+\frac{D_{\min}}{\eta_{\max}}}\\
      &\quad\le
        \frac{2K\brac{\epsilon G_{\max}+\epsilon^{2}L_{\max}}}{\sqrt{T}}\\
      &\Lambda_{T}(\eta_{\min},2M)\\
      &\quad\le
          \Log{\frac{6\abs{\cS}(2M)^{2}}{D_{\min}^{2}}}+1\\
      &\quad\le
        \Log{\frac{24M^{2}T^{2}\brac{\Ceil{\log_{2}\brac{\frac{TG_{\max}}{\epsilon L_{\max}}}}+1}}{\epsilon^{2}}}+1\\
      &\quad\le
        2\Log{\frac{5MT}{\epsilon}} + \Log{\log_{2}\brac{\frac{TG_{\max}}{\epsilon L_{\max}}}+2}+1
    \end{align*}
    Hence, hiding constants we may write
    \begin{align*}
      R_{T}(\vec{\cmp})
      &\le
        O\Bigg(
        G_{\max}((M+\epsilon)\Lambda_{T}^{*}+P_{T})
        +L_{\max}\sbrac{(M+\epsilon)^{2}\Lambda_{T}^{*} +(M+\epsilon)P_{T}}
        +\sqrt{(M^{2}\Lambda_{T}^{*}+MP_{T})\Omega_{T}},
        \Bigg).
    \end{align*}
    where
    $\Lambda_{T}^{*}\le O\brac{\Log{\frac{MT}{\epsilon}}+\Log{\Log{\frac{TG_{\max}}{\epsilon L_{\max}}}}}\le O\brac{\Log{\frac{MT\Log{T}}{\epsilon}}+\Log{\Log{\frac{G_{\max}}{\epsilon L_{\max}}}}}$.
    Finally, the proof is completed by observing that
    if the $\ell_{t}$ are $L_{t}$-smooth, then using the
    self-bounding property we have
    $\norm{g_{t}^{(\tau)}}^{2}\le 2L_{t}\brac{\ell_{t}(\wt^{(\tau)})-\ell_{t}^{*}}$
    for $\ell_{t}^{*}=\min_{\w\in W}\ell_{t}(\w)$, and thus
    \begin{align*}
      \Omega_{T}
      &=
        \sumtT KL_{t}\sbrac{\ell_{t}(\cmp_{t})-\ell_{t}(\wt^{(\tau)})}+4\sumtT\norm{g_{t}^{(\tau)}}^{2}\\
      &\le
        \sumtT KL_{t}\sbrac{\ell_{t}(\cmp_{t})-\ell_{t}(\wt^{(\tau)})}+8\sumtT L_{t}\sbrac{\ell_{t}(\wt^{(\tau)})-\ell_{t}^{*}}\\
        &\le
          \sumtT KL_{t}\sbrac{\ell_{t}(\cmp_{t})-\ell_{t}^{*}}
    \end{align*}
    where the second-to-last line chooses $K\ge 8$, and simultaneously we have using
    \Cref{lemma:omega-bound} that
    \begin{align*}
      \Omega_{T}
      &\le 3(K+4)\sumtT\sbrac{G_{t}^{2} + L_{t}^{2}D_{j}^{2}}\\
      &\le
        3(K+4)\sumtT\sbrac{G_{t}^{2}+4L_{t}^{2}M^{2}},
    \end{align*}
    and so we have
    $\Omega_{T}\le O\brac{\sumtT L_{t}\sbrac{\ell_{t}(\cmp_{t})-\ell_{t}^{*}}\minOp \sumtT G_{t}^{2}+L_{t}^{2}M^{2}}$.
\end{proof}

\subsection{Proof of Theorem~\ref{thm:smooth-dynamic}}%
\label{app:smooth-dynamic}
\begin{manualtheorem}{\ref{thm:smooth-dynamic}}
  For all $t$ let $\ell_{t}:W\to \R$ be  $(G_{t},L_{t})$-quadratically bounded
  and $L_{t}$-smooth convex function with $G_{t}\in[0,G_{\max}]$ and $L_{t}\in[0,L_{\max}]$.
  Let $\epsilon>0$, $K\ge 8$, and for any $i,j\ge 0$ let
  $D_{j}=\frac{\epsilon }{\sqrt{T}}\sbrac{2^{j}\minOp 2^{T}}$ and
  $\eta_{i}=\frac{1}{KL_{\max}\sqrt{T}}\sbrac{2^{i}\minOp \sqrt{T}}$,
  and let $\cS=\Set{(\eta_{i},D_{j}):i,j\ge 0}$.
  Then for any
  $\vec{\cmp}=(\cmp_{1},\ldots,\cmp_{T})$ in $W$,
  \Cref{alg:dynamic-meta} guarantees
  \begin{align*}
    R_{T}(\vec{\cmp})
    &\le
      O\Bigg(
      G_{\max}(M+\epsilon)\Lambda_{T}^{*}+L_{\max}(M+\epsilon)^{2}\Lambda_{T}^{*}+L_{\max}(M+\epsilon)P_{T}\\
    &\qquad
      +\sqrt{\sumtT \sbrac{\ell_{t}(\cmp_{t})-\ell_{t}^{*}}^{2}}
      +\sqrt{(M^{2}\Lambda_{T}^{*}+MP_{T})\sumtT L_{t}\sbrac{\ell_{t}(\cmp_{t})-\ell_{t}^{*}}}\Bigg),
  \end{align*}
  where $M=\max_{t}\norm{\cmp_{t}}$,
  $P_{T}=\sum_{t=2}^{T}\norm{\cmp_{t}-\cmp_{\tmm}}$,
  and $\Lambda_{T}^{*}\le O\brac{\Log{\frac{M\sqrt{T}\Log{\sqrt{T}}}{\epsilon}}}$.
\end{manualtheorem}
\begin{proof}
  By \Cref{lemma:large-M}, we can assume that there is
  a $\tau=(\eta,D)\in\cS$ for which $D\ge \max_{t}\norm{\cmp_{t}}=M$,
  since otherwise the regret is bounded as
  \begin{align}
    R_{T}(\vec{\cmp})\le 2M\brac{G_{\max}+ML_{\max}}\Log{\frac{M\sqrt{T}}{\epsilon}}.\label{eq:smooth-dynamic:large-M}
  \end{align}
  Hence, we can assume there is a $(\eta,D)\in\cS$ which has $M\le D$.
  For any such $(\eta,D)\in\cS$,
  we can apply
  \Cref{lemma:dynamic-untuned} to get
  \begin{align*}
    R_{T}(\vec{\cmp})
      &\le
      2k C_{\cS}+2kDG_{\max}\Lambda_{T}(\tau)
      +\frac{\norm{\cmp_{T}}^{2}+2DP_{T}+4kD^{2}\Lambda_{T}(\tau)}{2\eta}\\
    &\quad
      +K\eta \sumtT L_{t}\sbrac{\ell_{t}(\cmp_{t})-\ell_{t}(\wt^{(\tau)})} +4\eta\sumtT\norm{g_{t}^{(\tau)}}^{2},
  \end{align*}
  where $g_{t}^{(\tau)}\in\partial\ell_{t}(\wt^{(\tau)})$, $P_{T}=\sum_{t=2}^{T}\norm{\cmp_{t}-\cmp_{\tmm}}$
  and
  \begin{align*}
    C_{\cS}&\defeq\frac{\sum_{\tilde\tau\in\cS}\mu_{\tilde\tau}}{\sum_{\tilde\tau\in\cS}\mu_{\tilde\tau}^{2}},\qquad
    \Lambda_{T}(\tau)\defeq \Log{\frac{\sum_{\tilde\tau\in\cS}\mu_{\tilde \tau}^{2}}{\mu_{\tau}^{2}}}+1,
  \end{align*}
  where for any $\tilde\tau=(\tilde D,\tilde \eta)\in\cS$ we define
  $\mu_{(\tilde\eta,\tilde D)}=\frac{1}{2\tilde D\brac{G_{\max}+\tilde D/\tilde\eta}}$.
  Using the self-bounding property of smooth functions,
  for any $g_{t}^{(\tau)}\in\partial\ell_{t}(\wt^{(\tau)})$ we have
  $\norm{g_{t}^{(\tau)}}^{2}\le 2L_{t}\sbrac{\ell_{t}(\wt^{(\tau)})-\ell_{t}^{*}}$
  for $\ell_{t}^{*}=\argmin_{\w\in W}\ell_{t}(\w)$, so
  the last line is bound as
  \begin{align*}
      K\eta \sumtT L_{t}\sbrac{\ell_{t}(\cmp_{t})-\ell_{t}(\wt^{(\tau)})} + 8\eta\sumtT L_{t}\sbrac{\ell_{t}(\wt^{(\tau)})-\ell_{t}^{*}}
     &\le
      K\eta \sumtT L_{t}\sbrac{\ell_{t}(\cmp_{t})-\ell_{t}^{*}}
  \end{align*}
  for $K\ge 8$. Hence,
  \begin{align}
    R_{T}(\vec{\cmp})
    &\le
      2k C_{\cS}+2kDG_{\max}\Lambda_{T}(\tau)
      +\frac{\norm{\cmp_{T}}^{2}+2DP_{T}+4kD^{2}\Lambda_{T}(\tau)}{2\eta}\nonumber\\
    &\quad
      +K\eta\sumtT L_{t}\sbrac{\ell_{t}(\cmp_{t})-\ell_{t}^{*}}\label{eq:smooth-dynamic:untuned}
  \end{align}
  Now suppose that $M\le D_{\min}$, then choosing
  $\tau=\tau_{\min}=(\eta_{\min},D_{\min})$ we would have
  \begin{align}
    R_{T}(\vec{\cmp})
    &\le
      2k C_{\cS}+2kD_{\min}G_{\max}\Lambda_{T}(\tau_{\min})
      +\frac{\norm{\cmp_{T}}^{2}+2D_{\min}P_{T}+4kD^{2}_{\min}\Lambda_{T}(\tau_{\min})}{2\eta_{\min}}\nonumber\\
    &\quad
      +K\eta_{\min}\sumtT L_{t}\sbrac{\ell_{t}(\cmp_{t})-\ell_{t}^{*}}\nonumber\\
    &\le
      2k C_{\cS}+2kD_{\min}G_{\max}\Lambda_{T}(\tau_{\min})\nonumber
      +\frac{D_{\min}}{\eta_{\min}}\half\brac{M+2P_{T}+4kD_{\min}\Lambda_{T}(\tau_{\min})}\nonumber\\
    &\quad
      +\frac{1}{\sqrt{T}L_{\max}}\sumtT L_{t}\sbrac{\ell_{t}(\cmp_{t})-\ell_{t}^{*}}\nonumber\\
    &\le
      2k C_{\cS}+2kG_{\max}\frac{\epsilon\Lambda_{T}(\tau_{\min})}{\sqrt{T}}
      +K \epsilon L_{\max}\brac{M+P_{T}+2k\frac{\epsilon \Lambda_{T}(\tau_{\min})}{\sqrt{T}}}\nonumber\\
    &\quad
      +\sqrt{\sumtT \sbrac{\ell_{t}(\cmp_{t})-\ell_{t}^{*}}^{2}}\label{eq:smooth-dynamic:small-M}
  \end{align}
  where the last line applies Cauchy-Schwarz inequality,
  observes that $D_{\min}/\eta_{\min}=K\epsilon L_{\max}$, and recalls $D_{\min}=\frac{\epsilon}{\sqrt{T}}$.
  Finally, assume that $M\in[D_{\min},D_{\max}]$, then there is a
  $D_{j}=\frac{\epsilon 2^{j}}{\sqrt{T}}$ for which
  $D_{j}\ge M\ge D_{j-1}=\half D_{j}$. Then, choosing
  $\tau=(\eta,D_{j})$, \Cref{eq:smooth-dynamic:untuned} yields
  \begin{align}
    R_{T}(\vec{\cmp})
    &\le
      2k C_{\cS}+4kMG_{\max}\Lambda_{T}(\eta_{\min},2M)
      +\frac{M^{2}+4MP_{T}+16kM^{2}\Lambda_{T}(\eta_{\min},2M)}{2\eta}\nonumber\\
    &\quad
      +K\eta\sumtT L_{t}\sbrac{\ell_{t}(\cmp_{t})-\ell_{t}^{*}}\label{eq:smooth-dynamic:tuned-M}
  \end{align}
  where we've observed that
  \begin{align*}
    \Lambda(\eta,D_{j})
    &=
      \Log{\frac{\sum_{\tilde\tau\in\cS}\mu_{\tilde\tau}^{2}}{\mu_{(\eta,D_{j})}}}+1\\
    &=
      \Log{\sum_{\tilde\tau\in\cS}\mu_{\tilde\tau}^{2}D_{j}^{2}\sbrac{G_{\max}+D_{j}/\eta}^{2}}+1\\
    &\le
      \Log{\sum_{\tilde\tau\in\cS}\mu_{\tilde\tau}^{2}(2M)^{2}\sbrac{G_{\max}+2M/\eta}^{2}}+1\\
    &=\Lambda_{T}(\eta,2M)
  \end{align*}
  so it remains to show that there is
  an $\eta$ that favorably balances the last two terms of \Cref{eq:smooth-dynamic:tuned-M}.

  Observe that the optimal choice for $\eta$ would be
  \begin{align*}
    \eta^{*}=\sqrt{\frac{M^{2}(1+16k\Lambda_{T}(\eta_{\min},2M))+4MP_{T}}{2K\sumtT L_{t}\sbrac{\ell_{t}(\cmp_{t})-\ell_{t}^{*}}}}.
  \end{align*}
  If $\eta^{*}\le \eta_{\min}$ then choosing $\eta=\eta_{\min}$ we have
  \begin{align}
    R_{T}(\vec{\cmp})
    &\le
      2k C_{\cS}+4kMG_{\max}\Lambda_{T}(\eta_{\min},2M)
      +\frac{M^{2}(1+16k\Lambda_{T}(\eta_{\min},2M)+4MP_{T}}{2\eta^{*}}
      +K\eta_{\min}\sumtT L_{t}\sbrac{\ell_{t}(\cmp_{t})-\ell_{t}^{*}}\nonumber\\
    &\le
      2k C_{\cS}+4kMG_{\max}\Lambda_{T}(\eta_{\min},2M)
      +\sqrt{\frac{K}{2}\brac{M^{2}(1+16k\Lambda_{T}(\eta_{\min},2M))+4MP_{T}}\Omega_{T}}\nonumber\\
    &\qquad
      +\sqrt{\sumtT\sbrac{\ell_{t}(\cmp_{t})-\ell_{t}^{*}}^{2}},\label{eq:smooth-dynamic:small-eta}
  \end{align}
  where the last line defines the short-hand notation
  $\Omega_{T}=\sumtT L_{t}\sbrac{\ell_{t}(\cmp_{t})-\ell_{t}^{*}}$ and
  uses Cauchy-Schwarz inequality to bound
  $K\eta_{\min}\sumtT L_{t}\sbrac{\ell_{t}(\cmp_{t})-\ell_{t}^{*}}\le \sqrt{\sumtT \sbrac{\ell_{t}(\cmp_{t})-\ell_{t}^{*}}^{2}}$.
  Likewise, if $\eta^{*}\ge \eta_{\max}$ then by choosing $\eta=\eta_{\max}$ we
  have via \Cref{eq:smooth-dynamic:tuned-M} that
  \begin{align}
    R_{T}(\vec{\cmp})
    &\le
      2k C_{\cS}+4kMG_{\max}\Lambda_{T}(\eta_{\min},2M)
      +\frac{M^{2}+4MP_{T}+16kM^{2}\Lambda_{T}(\eta_{\min},2M)}{2\eta_{\max}}\nonumber\\
    &\quad
      +K\eta^{*}\sumtT L_{t}\sbrac{\ell_{t}(\cmp_{t})-\ell_{t}^{*}}\nonumber\\
    &=
      2k C_{\cS}+4kMG_{\max}\Lambda_{T}(\eta_{\min},2M)
      +K L_{\max}\brac{M^{2}(1+8k\Lambda_{T}(\eta_{\min},2M))+2MP_{T}}\nonumber\\
    &\quad
      +\sqrt{\frac{K}{2}\brac{M^{2}(1+16k\Lambda_{T}(\eta_{\min},2M))+4MP_{T}}\Omega_{T}}\label{eq:smooth-dynamic:large-eta}
  \end{align}
  Finally, if $\eta^{*}\in[\eta_{\min},\eta_{\max}]$ then there is an
  $\eta_{i}=\frac{2^{i}}{\epsilon L_{\max}\sqrt{T}}$ for which
  $\eta_{i}\le \eta^{*}\le \eta_{i+1}=2\eta_{i}$,
  so \Cref{eq:smooth-dynamic:tuned-M} is gives us
  \begin{align}
    R_{T}(\vec{\cmp})
    &\le
      2k C_{\cS}+4kMG_{\max}\Lambda_{T}(\eta_{\min},2M)
      +3\sqrt{\frac{K}{2}\brac{M^{2}(1+16k\Lambda_{T}(\eta_{\min},2M))+4MP_{T}}\Omega_{T}}\label{eq:smooth-dynamic:tuned}
  \end{align}
  Finally, combining
  \Cref{eq:smooth-dynamic:small-M,eq:smooth-dynamic:large-M,eq:smooth-dynamic:large-eta,eq:smooth-dynamic:small-eta,eq:smooth-dynamic:tuned},
  we have
  \begin{align*}
    R_{T}(\vec{\cmp})
    &\le
      2k C_{\cS}\nonumber
      +4kG_{\max}\sbrac{M\Lambda_{T}(\eta_{\min},2M)+\frac{\epsilon\Lambda_{T}(\tau_{\min})}{\sqrt{T}}}\nonumber\\
    &\quad
    +2M\brac{G_{\max}+ML_{\max}}\Log{\frac{M\sqrt{T}}{\epsilon}}\\
    &\quad
      +K \epsilon L_{\max}\brac{M+P_{T}+2k\frac{\epsilon \Lambda_{T}(\tau_{\min})}{\sqrt{T}}}\nonumber\\
    &\quad
      +K L_{\max}\brac{M^{2}(1+8k\Lambda_{T}(\eta_{\min},2M))+2MP_{T}}\nonumber\\
    &\quad
      +\sqrt{\sumtT\sbrac{\ell_{t}(\cmp_{t})-\ell_{t}^{*}}^{2}}\\
    &\quad
      +3\sqrt{\frac{K}{2}\brac{M^{2}(1+16k\Lambda_{T}(\eta_{\min},2M))+4MP_{T}}\Omega_{T}}\\
  \end{align*}
  Lastly, note that by \Cref{lemma:mu-bounds}, we have
  \begin{align*}
    C_{\cS}
    &\le
      2\sqrt{T}D_{\min}\brac{G_{\max}+\frac{D_{\min}}{\eta_{\max}}}\\
    &=
      2\epsilon G_{\max}+\frac{2K\epsilon^{2} L_{\max}}{\sqrt{T}}\\
    \Lambda_{T}(\tau)
    &\le
      \Log{\frac{24\eta_{\max}^{2}D^{4}}{\eta_{\min}^{2}D_{\min}^{4}}\minOp \frac{6\abs{\cS_{\eta}}D^{2}}{D_{\min}^{2}}}+1\\
    &\le
      \Log{\frac{6\abs{\cS_{\eta}}D^{2}}{D_{\min}^{2}}}+1\\
  \end{align*}
  so
  $\Lambda_{T}(\eta_{\min},D_{\min})\le \Log{6\log_{2}\brac{\lceil\log_{2}(\sqrt{T})\rceil+1}}\le O\brac{\log(\log(\sqrt{T}))}$
  and
  $\Lambda_{T}(\eta_{\min},2M)\le \Log{\frac{24TM^{2}\log_{2}\brac{\lceil\log_{2}(\sqrt{T})\rceil+1}}{\epsilon^{2}}}\le  O\brac{\Log{M\sqrt{T}\Log{\sqrt{T}}/\epsilon}}$.
  Overall, we have
  \begin{align*}
    R_{T}(\vec{\cmp})
    &\le
      O\Bigg(
      G_{\max}(M+\epsilon)\Lambda_{T}^{*}\\
    &\qquad
    +L_{\max}(M+\epsilon)^{2}\Lambda_{T}^{*}+L_{\max}(M+\epsilon)P_{T}\\
    &\qquad
      +\sqrt{\sumtT\sbrac{\ell_{t}(\cmp_{t})-\ell_{t}^{*}}^{2}}\\
    &\qquad
      +\sqrt{\brac{M^{2}\Lambda_{T}^{*}+MP_{T}}\Omega_{T}}\bigg)
  \end{align*}
  where
  $\Lambda_{T}^{*}\le O\brac{\Log{\frac{M\sqrt{T}\Log{\sqrt{T}}}{\epsilon}}+\Log{\Log{\sqrt{T}}}}\le O\brac{\Log{\frac{M\sqrt{T}\Log{\sqrt{T}}}{\epsilon}}}$.
\end{proof}

\subsection{Proof of Theorem~\ref{thm:dynamic-lb}}%
\label{app:dynamic-lb}

We focus on the case where $G/L\le M$, since otherwise when $G/L\ge M$
the loss function $\ell_{t}(\w)=(\half G+\half LM)\xi_{t}\w$ for $\xi_{t}\in\Set{-1,1}$
satisfies $\abs{\ell_{t}'(\w)}=\half(G+LM)\le G$ for any $\w\in W$, so $\ell_{t}$ is $G$-Lipschitz.
Hence, existing lower bounds tell us that there exists a sequence
$\xi_{t}\in[-1,1]$ such that
$R_{T}(\vec{\cmp})\ge \Omega\big(G\sqrt{MP_{T}T}\big)\ge\Omega\big(\half(G+LM)\sqrt{MP_{T}T}\big)=\Omega\big(\half G\sqrt{MP_{T}T}+\half LM^{3/2}\sqrt{P_{T}T}\big)$
where $M=\max_{t}\norm{\cmp_{t}}$ and $P_{T}=\sum_{t=2}^{T}\norm{\cmp_{t}-\cmp_{\tmm}}$
\cite{zhang2018adaptive}.
\begin{manualtheorem}{\ref{thm:dynamic-lb}}
  For any $M>0$ there is a sequence
  of $(G,L)$-quadratically bounded functions with $\frac{G}{L}\le M$
  such that for any $\gamma\in[0,\half]$,
  \begin{align*}
      R_{T}(\vec{\cmp})\ge \frac{G}{4}M^{1-\gamma}\sbrac{P_{T}T}^{\gamma}+\frac{L}{8}M^{2-\gamma}\sbrac{P_{T}T}^{\gamma}.
  \end{align*}
  where $P_{T}=\sum_{t=2}^{T}\norm{\cmp_{t}-\cmp_{\tmm}}$
  and $M\ge\max_{t}\norm{\cmp_{t}}$.
\end{manualtheorem}
\begin{proof}
  On each round $t$, we can always find a $\cmp_{t}$ such that $\cmp_{t}\perp\wt$.
  Let $\norm{\cmp_{t}}:= \sigma\le M$ for some $\sigma$ to be decided.
  Let $G>0$, $L\ge 0$ such that $G/L\le \sigma$, let $\xi_{t}=\frac{\cmp_{t}}{\norm{\cmp_{t}}}$, and on each round set
  \begin{align*}
    \ell_{t}(\w)=-\half G\inner{\xi_{t},w}+ \frac{L}{4}(\sigma - \inner{\xi_{t},\w})^{2}.
  \end{align*}
  Observe that these losses are $(\tilde G,\tilde L)$ quadratically bounded with
  $\tilde G=\half G+\half \sigma L$ and $\tilde L= L$, and
  $\tilde G/\tilde L \le \sigma\le M$ as required.
  Since $\wt\perp\xi_{t}$ and $\inner{\xi_{t},\cmp_{t}}=\norm{\cmp_{t}}=\sigma$,
  we have
  \begin{align*}
    R_{T}(\vec{\cmp})=\sumtT\ell_{t}(\wt)-\ell_{t}(\cmp_{t})\ge \half G\sigma T+ \frac{L}{4}T\sigma^{2}.
  \end{align*}
  Note also that the path-length of this comparator sequence is bounded as
  \begin{align*}
    P_{T}=\sum_{t=2}^{T}\norm{\cmp_{t}-\cmp_{\tmm}}\le 2\sigma T.
  \end{align*}
  Now for $\mu\in[0,1/2]$ set $\sigma = MT^{-\mu}$, then
  the path-length is bounded as
  \begin{align*}
    P_{T}\le 2MT^{1-\mu}
  \end{align*}
  and the regret is bounded below by
  \begin{align*}
    \half G M T^{1-\mu}+\frac{L}{4}T^{1-2\mu}M^{2}.
  \end{align*}
  Now set $\gamma=\frac{1-2\mu}{2-\mu}\in[0,\half]$
  and consider the second term:
  \begin{align*}
    \frac{L}{4}T^{1-2\mu}M^{2}
    &=
      \frac{L}{4}(MT^{1-\mu})^{\gamma}(MT^{1-\mu})^{1-\gamma}T^{-\mu}M\\
    &\ge
      \frac{L}{4\cdot 2^{\gamma}}(P_{T})^{\gamma}(MT^{1-\mu})^{1-\gamma}T^{-\mu}M\\
    &=
      \frac{L}{8}M^{2-\gamma}P_{T}^{\gamma}T^{(1-\mu)(1-\gamma)-\mu}\\
    &=
      \frac{L}{8}M^{2-\gamma}\sbrac{P_{T}T}^{\gamma}
  \end{align*}
  where the last line observes $\gamma = \frac{1-2\mu}{2-\mu}\in[0,\half]$, so that $(1-\mu)(1-\gamma)-\mu=\gamma$.
  Similarly,
  \begin{align*}
    \half GMT^{1-\mu}&=\half G(MT^{1-\mu})^{\gamma}(MT^{1-\mu})^{1-\gamma}\ge \frac{1}{2 \cdot 2^{\gamma}} G M^{1-\gamma}(P_{T})^{\gamma}T^{(1-\mu)(1-\gamma)}\\\
    &\ge \frac{1}{4}GM^{1-\gamma}(P_{T}T)^{\gamma}.
  \end{align*}
  so
  \begin{align*}
    R_{T}(\vec{\cmp})\ge \frac{G}{4}M^{1-\gamma}\sbrac{P_{T}T}^{\gamma}+\frac{L}{8}M^{2-\gamma}\sbrac{P_{T}T}^{\gamma}.
  \end{align*}
\end{proof}

\section{Supporting Lemmas}

We collect here various miscellaneous supporting lemmas
that we use throughout paper. The following lemma is standard
but shown here for completeness.

\begin{restatable}{lemma}{SqrtBounds}\label{lemma:sqrt-bounds}
  Let $a_{1},\ldots, a_{T}$ be arbitrary non-negative numbers in $\R$.
  Then
  \begin{align*}
    \sqrt{\sumtT a_{t}}\le \sumtT\frac{a_{t}}{\sqrt{\sum_{s=1}^{t}a_{s}}}\le 2\sqrt{\sumtT a_{t}}
  \end{align*}
\end{restatable}
\begin{proof}
  By concavity of $x\mapsto\sqrt{x}$, we have
  \begin{align*}
    \sqrt{a_{1:t}}-\sqrt{a_{1:\tmm}}\ge \frac{a_{t}}{2\sqrt{a_{1:t}}},
  \end{align*}
  so summing over $t$ and observing the resulting telescoping sum yields
  \begin{align*}
    \sumtT\frac{a_{t}}{\sqrt{a_{1:t}}} \le 2 \sumtT\sqrt{a_{1:t}}-\sqrt{a_{1:\tmm}} = 2\sqrt{a_{1:T}}.
  \end{align*}
  For the lower bound, observe that
  \begin{align*}
    \sumtT\frac{a_{t}}{\sqrt{a_{1:t}}}\ge \sumtT\frac{a_{t}}{\sqrt{a_{1:T}}} = \frac{a_{1:T}}{\sqrt{a_{1:T}}}=\sqrt{a_{1:T}}
  \end{align*}
\end{proof}

We borrow the following lemma from \citet{jacobsen2022parameter}.
\begin{restatable}{lemma}{SimpleRadiallySymmetricBound}\label{lemma:simple-radially-symmetric-bound}
  \citep[Lemma 7]{jacobsen2022parameter}
  Let $f:\R\to\R$ and let $g:W\to\R$ be defined as $g(x)=f(\norm{x})$.
  Suppose that $f'(x)$ is concave and non-negative.
  If $f$ is twice-differentiable at $\norm{x}$ and $\norm{x}>0$, then
  \begin{align*}
    \grad^{2}g(x)\succeq f''(\norm{x})I
  \end{align*}
\end{restatable}

The following is a simple modification of the stability lemma used in
\citet{jacobsen2022parameter}, reported here with slight modification to handle a leading constant.
\begin{restatable}{lemma}{LipschitzStability}\label{lemma:lipschitz-stability}
  For all $t$, set $\psi_{\tpp}(\w)=\Psi_{\tpp}(\norm{\w})$
  where
  $\Psi_{t}:\R_{\ge 0}\to\R_{\ge 0}$ is a convex function satisfying
  $\Psi_{t}'(x)\ge 0$, $\Psi_{t}''(x)\ge 0$, and $\Psi_{t}'''(x)\le 0$ for all
  $x\ge 0$. Let $c>0$ and
  assume that there exists an $\mathring{x}_{t}>0$ and $G_{t}$-Lipschitz convex function
  $\eta_{t}:\R_{\ge 0}\to\R_{\ge0}$ such that
  $\abs{\Psi_{t}'''(x)}\le\frac{2 \eta_{t}'(x)}{(c+1)^{2}}\Psi_{t}''(x)^{2}$ for all
  $x\ge \mathring{x}_{t}$.
  Then for any $\wtpp,\wt\in W$,
  \begin{align*}
    \hat\delta_{t}&\defeq c G_{t}\norm{\wt-\wtpp}-D_{\psi_{t}}(\wtpp|\wt) - \eta_{t}(\norm{\wtpp})G_{t}^{2}
    \le
      \frac{(c+1)^{2} G_{t}^{2}}{2 \Psi_{t}''(\mathring{x}_{t})}
  \end{align*}
\end{restatable}
\begin{proof}
  The proof follows using similar arguments to \citet{jacobsen2022parameter} with
  a few minor adjustments to correct for the leading term $c$.

  First, consider the case that the origin is contained in the line segment
  connecting $w_t$ and $w_{t+1}$. Then, there exists sequences
  $\hat w^1_t,\hat w^2_t\dots$ and $\hat w^1_{t+1},\hat w^2_{t+1}\dots$ such that
  $\lim_{n\to\infty} \hat w^n_t=w_t$, $\lim_{n\to\infty} \hat w^n_{t+1}=w_{t+1}$ and $0$
  is not contained in the line segment connecting $\hat w^n_t$ and $\hat w^n_{t+1}$
  for all $n$. Since $\psi$ is twice differentiable everywhere except the
  origin, if we define
  $\hat \delta_t^n = G_{t}\norm{\hat w_t^n - \hat w_{t+1}^n} - D_{\psi_t}(\hat w_{t+1}^{n}|\hat w_t^{n}) - \eta_t(\|\hat w_{t+1}^{n}\|)G_{t}^{2}$,
  then $\lim_{n\to\infty}\hat \delta_t^n=\hat \delta_t$. Thus, it suffices to prove the result for the case that the origin is \emph{not} contained in the line segment connecting $w_t$ and $w_{t+1}$. The rest of the proof considers exclusively this case.

  For brevity denote $\hat\delta_{t}\defeq G_{t}\norm{\wt-\wtpp}-D_{\psi_{t}}(\wtpp|\wt)-\eta_{t}(\norm{\wtpp})\norm{g_{t}}^{2}$.
  Since the origin is not in the line segment connecting $w_t$ and $w_{t+1}$, $\psi_t$ is twice differentiable on this line segment. Thus, By Taylor's theorem, there is a $\wtilde$ on the line connecting
  $\wt$ and $\wtpp$ such that
  \begin{align*}
    D_{\psi_{t}}(\wtpp|\wt)
    &=
      \half\norm{\wt-\wtpp}^{2}_{\grad^{2}\psi_{t}(\wtilde)}
    \ge
      \half\norm{\wt-\wtpp}^{2}\Psi_{t}''(\norm{\wtilde})
\end{align*}
where the last line observes $\psi_{t}(\w)=\Psi_{t}(\norm{\w})$ and uses the regularity assumptions $\Psi_{t}'''(x)\le0$, and
$\Psi_{t}'(x)\ge 0$ for $x\ge 0$ to apply \Cref{lemma:simple-radially-symmetric-bound}.
Hence,
\begin{align*}
  \hat\delta_{t}&=c G_{t}\norm{\wt-\wtpp}-D_{\psi_{t}}(\wtpp|\wt)-\eta_{t}(\norm{\wtpp})G_{t}^{2}\\
  &\le
    cG_{t}\norm{\wt-\wtpp}-\half\norm{\wt-\wtpp}^{2}\Psi_{t}''(\norm{\wtilde})
            -\eta_{t}(\norm{\wtpp})G_{t}^{2}\\
  &\overset{(a)}{\le}
    cG_{t}\norm{\wt-\wtpp}-\half\norm{\wt-\wtpp}^{2}\Psi_{t}''(\norm{\wtilde})
            -\eta_{t}(\norm{\wtilde})G_{t}^{2}+\eta_{t}'(\norm{\wtilde})G_{t}^{2}\norm{\wtpp-\wtilde}\\
  &\overset{(b)}{\le}
    (c+1)G_{t}\norm{\wt-\wtpp}-\half\norm{\wt-\wtpp}^{2}\Psi_{t}''(\norm{\wtilde})
    -\eta_{t}(\norm{\wtilde})G_{t}^{2}\\
  &\overset{(c)}{\le}
    \frac{(c+1)^{2}G_{t}^{2}}{2\Psi_{t}''(\norm{\wtilde})}-\eta_{t}(\norm{\wtilde})G_{t}^{2}\\
\end{align*}
where $(a)$ uses convexity of $\eta_{t}(x)$, $(b)$ uses the Lipschitz assumption
$\eta_{t}'(\norm{\wtilde})\le 1/G_{t}$ and the fact that
$\norm{\wtilde-\wt}\le\norm{\wtpp-\wt}$ for any $\wtilde$ on the line connecting
$\wt$ and $\wtpp$,
and $(c)$ uses Fenchel-Young inequality.
If $\norm{\wtilde}\le\mathring{x}_{t}$, then we have
\begin{align*}
    \frac{(c+1)^{2}G_{t}^{2}}{2\Psi_{t}''(\norm{\wtilde})}-\eta_{t}(\norm{\wtilde})G_{t}^{2}\le \frac{(c+1)^{2}G_{t}^{2}}{2\Psi_{t}''(\mathring{x}_{t})},
\end{align*}
which follows from the fact that $\Psi_{t}'''(x)\le 0$ implies $\Psi_{t}''(x)$ is
non-increasing in $x$, and hence $\Psi_{t}''(\norm{\wtilde})\ge \Psi_{t}''(\mathring{x}_t)$. Otherwise,
if
$\norm{\wtilde}\ge \mathring{x}_t$,
we have by assumption that
$\frac{\abs{\Psi_{t}'''(x)}}{\Psi_{t}''(x)^{2}}=\frac{-\Psi_{t}'''(x)}{\Psi_{t}''(x)^{2}}=\frac{d}{dx}\frac{1}{\Psi_{t}''(x)}\le\frac{2 \eta_{t}'(x)}{(c+1)^{2}}$
for any $x\ge \mathring{x}_t$, so
integrating from $\mathring{x}_t$ to $\norm{\wtilde}$ we have
\begin{align*}
  \frac{1}{\Psi_{t}''(\norm{\wtilde})}-\frac{1}{\Psi_{t}''(\mathring{x}_t)}&\le \frac{2}{(c+1)^{2}}\int_{\mathring{x}_t}^{\norm{\wtilde}}\eta_{t}'(x)dx,
\end{align*}
so:
\begin{align*}
  \frac{1}{\Psi_{t}''(\norm{\wtilde})}&\le \frac{1}{\Psi_{t}''(\mathring{x}_t)}+\frac{2}{(c+1)^{2}}\int_{\mathring{x}_t}^{\norm{\wtilde}}\eta_{t}'(x)dx\\
                                    &\le \frac{1}{\Psi_{t}''(\mathring{x}_t)}+\frac{2}{(c+1)^{2}}\int_{0}^{\norm{\wtilde}}\eta_{t}'(x)dx\\
                                  &=\frac{1}{\Psi_{t}''(\mathring{x}_t)}+\frac{2\eta_{t}(\norm{\wtilde})}{(c+1)^{2}},
\end{align*}
and hence,
\begin{align*}
  \frac{(c+1)^{2}G_{t}^{2}}{2\Psi_{t}''(\norm{\wtilde})}-\eta_{t}(\norm{\wtilde})G_{t}^{2}
  &\le
    \frac{(c+1)^{2}G_{t}^{2}}{2\Psi_{t}''(\mathring{x}_t)}+\frac{(c+1)^{2}G_{t}^{2}}{2}\frac{2}{(c+1)^{2}}\eta_{t}(\norm{\wtilde})
    -\eta_{t}(\norm{\wtilde})G_{t}^{2}\\
  &=
      \frac{(c+1)^{2}G_{t}^{2}}{2\Psi_{t}''(\mathring{x}_t)}+\eta_{t}(\norm{\wtilde})G_{t}^{2}-\eta_{t}(\norm{\wtilde})G_{t}^{2}\\
  &=
    \frac{(c+1)^{2}G_{t}^{2}}{2\Psi_{t}''(\mathring{x}_{t})},
\end{align*}
so in either case we have
\begin{align*}
  \hat\delta_{t}&=
  G_{t}\norm{\wt-\wtpp}-D_{\psi_{t}}(\wtpp|\wt)-\eta_{t}(\norm{\wtpp})G_{t}^{2}\\
  &\le
  \frac{(c+1)^{2}G_{t}^{2}}{2\Psi_{t}''(\mathring{x}_{t})}.
\end{align*}

\end{proof}

\end{document}